\documentclass[10pt,twocolumn,letterpaper]{article}

\usepackage{cvpr}

\usepackage{booktabs}
\usepackage{multirow} 
\usepackage[most]{tcolorbox}

\definecolor{cvprblue}{rgb}{0.21,0.49,0.74}
\usepackage[pagebackref,breaklinks,colorlinks,allcolors=cvprblue]{hyperref}

\def\confName{CVPR}
\def\confYear{2027}

\title{NEGATE: Constrained Semantic Guidance for Linguistic Negation in Text-to-Video Diffusion}

\author{
   Taewon Kang$^{1}$, Ming C. Lin$^{1}$\thanks{Corresponding author} \\
   $^{1}$University of Maryland at College Park, United States \\
   {\large \texttt{taewon@umd.edu, lin@umd.edu}} \\
}

\definecolor{rxgreen}{RGB}{0,143,76}

\begin{document}
\maketitle

\begin{abstract}
    Negation is a fundamental linguistic operator, yet it remains inadequately modeled in vision-language generation systems. In this work, we present a formal treatment of linguistic negation in vision-language models by modeling it as a structured feasibility constraint on semantic guidance within diffusion dynamics. Rather than introducing heuristics or retraining model parameters, we reinterpret classifier-free guidance as defining a semantic update direction and enforce negation by projecting the update onto a convex constraint set derived from linguistic structure. This formulation provides a unified framework for handling diverse negation phenomena, including object absence, graded non-inversion semantics, multi-negation composition, and scope-sensitive disambiguation. Our approach is training-free, compatible with pretrained diffusion backbones, and defined on the classifier-free guidance update, which lets it operate over temporally evolving video trajectories. In addition, we introduce a structured negation-centric benchmark suite that isolates distinct linguistic failure modes in generative systems, to further research in this area. Experiments demonstrate that our method achieves robust negation compliance while preserving visual fidelity and structural coherence, moving the treatment of linguistic negation in generative vision-language models beyond representation-level evaluation.
\end{abstract}

\section{Introduction}
\begin{figure}[t]
    \centering
    \includegraphics[width=\linewidth]{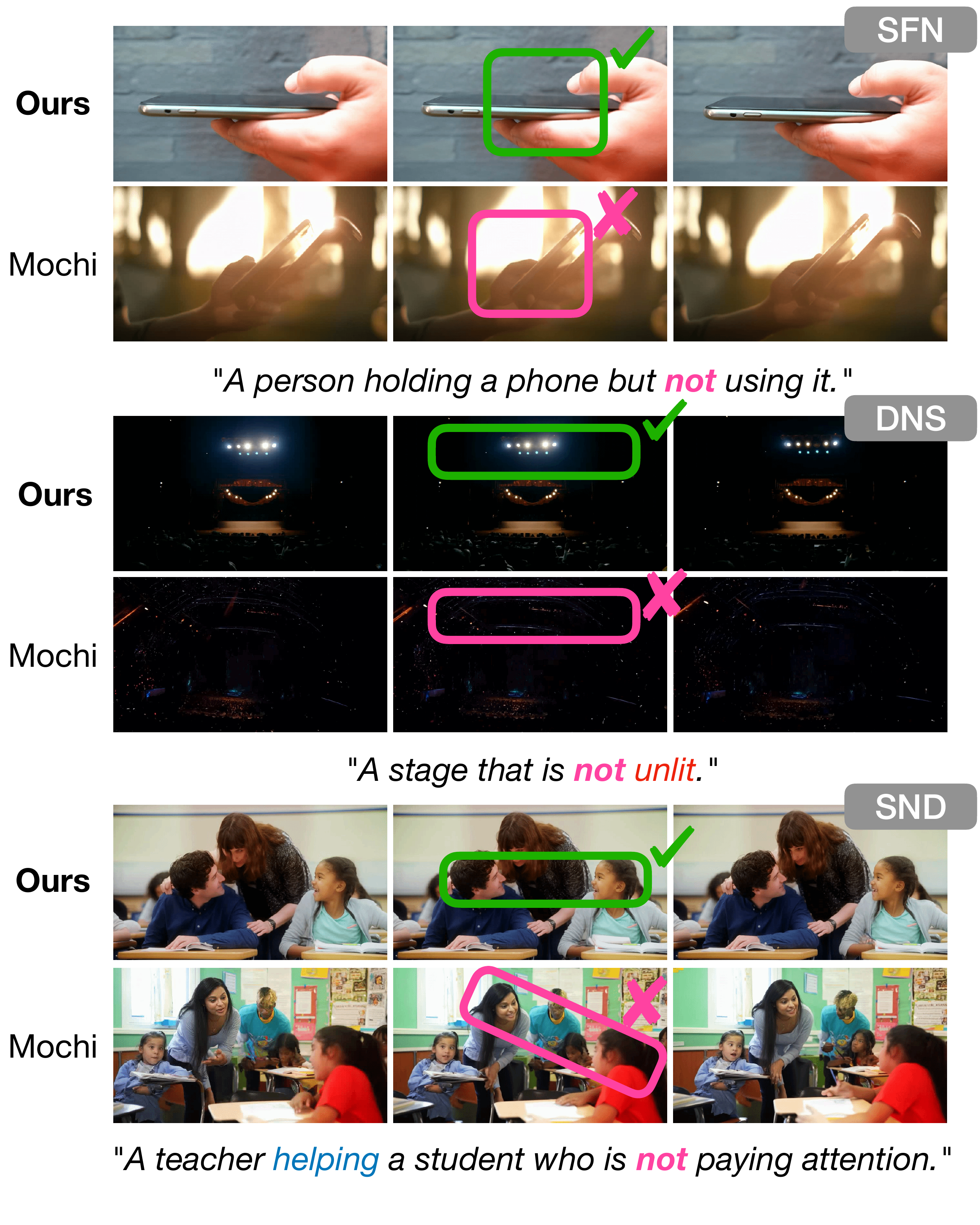}
    \caption{\textbf{Negation beyond object removal.} 
    Structural Functional Negation (SFN), Double Negation Sensitivity (DNS), and Scoped Negation Disambiguation (SND). Baseline diffusion models collapse negation into semantic inversion or scope errors, whereas our model enforces negation as a constraint on diffusion dynamics, yielding correct logical interpretation and stable scene composition. Qualitative results and comparisons with more video diffusion models are provided in Section~\ref{sec:qualitative}.}
    \label{fig:teaser}
\end{figure}

Vision-language models (VLMs) have achieved remarkable success in text-conditioned image and video generation. Modern diffusion-based systems can synthesize complex visual scenes from natural language descriptions with high fidelity and compositional richness~\cite{chefer2023attend,xie2025dymo}. However, despite these advances, a fundamental limitation persists: current generative models struggle to correctly interpret and enforce \emph{linguistic negation}. Negation is not merely the absence of a concept, nor is it equivalent to supplying an externally specified exclusion. Prompts such as ``a highway at sunset, with no vehicles,'' ``a person holding a phone but not using it,'' or ``a room that is not unlit'' encode structured linguistic constraints involving scope, composition, graded moderation, and logical interaction. Psycholinguistic studies suggest that negation is processed through activation followed by inhibition of affirmative representations~\cite{kaup2006processing,kaup2007experiential}, highlighting its structured and dynamic nature. Existing generative models frequently violate these constraints, producing outputs that include forbidden objects, misapply scope, or over-correct toward unintended opposites. This failure reveals a deeper gap: while diffusion models align with affirmative semantics, they lack a principled formulation for negation as a structured linguistic operator. Recent work has highlighted this limitation from a representational perspective. In particular, recent work~\cite{alhamoud2025vision,singh2024learn,xiao2026notjustwhatsthere} demonstrates that vision-language models often fail to distinguish between positive and negated captions in retrieval settings. These studies establish that negation sensitivity is weak at the embedding level. However, they primarily focus on \emph{benchmarking} representation separability in static image–text matching. They do not address how negation should influence the generative process itself, nor how structured linguistic negation can be enforced over temporal trajectories in video generation.

In this work, we shift the perspective from representation evaluation to \emph{generative formulation}. We ask a different question: how should linguistic negation be modeled as part of the semantic content of a prompt, and how should that modeling constrain the evolution of generative trajectories? Rather than treating negation as a data deficiency to be corrected through retraining, we formalize negation as a structured feasibility condition on semantic guidance within diffusion dynamics, drawing inspiration from constrained dynamical perspectives in diffusion-based control~\cite{zhang2025constrained,zampini2025training}. More specifically, we reinterpret classifier-free guidance as defining a semantic update direction in latent space and treat negation as a constraint restricting projection along directions associated with negated concepts. This leads to a convex feasibility formulation in which each diffusion step computes the minimal modification required to restore compliance. Most importantly, this formulation does not introduce a new generative backbone, nor does it require retraining. Instead, it provides a {\em principled mapping from linguistic structure to constraint geometry in semantic guidance space}.

This modeling perspective enables us to systematically handle diverse linguistic phenomena, including: (i) object-level absence, (ii) temporal suppression in evolving scenes, (iii) implicit category exclusivity, (iv) multi-negation composition, (v) structural functional negation, (vi) graded non-inversion semantics, (vii) double negation sensitivity, and (viii) scope-sensitive disambiguation. Although linguistically distinct, we show that all of these cases admit a unified convex constraint parameterization, transforming heterogeneous negation forms into a consistent feasibility problem over generative trajectories. Because constraints are enforced at the level of trajectory evolution, the framework applies directly to video generation, where negation violations may emerge temporally after initial structure formation; our experiments target this video setting. We further introduce a structured negation-centric benchmark suite designed to isolate and evaluate distinct linguistic failure modes in generative systems. Unlike prior negation datasets focused on representation separability~\cite{alhamoud2025vision,singh2024learn}, our benchmark measures distributional compliance under structured constraints, explicitly targeting trajectory-level violations.

The key contributions of this work are threefold:
\begin{enumerate}
\item \textbf{Formal Modeling of Linguistic Negation in VLMs.}
We model diverse linguistic negation phenomena as {\em structured convex feasibility constraints in semantic guidance space}, within a single formulation.
\item \textbf{Constraint-Based Generative Enforcement.}
We introduce a training-free mechanism that enforces negation during diffusion through minimal-energy projection, ensuring stability and compliance without architectural modification.
\item \textbf{Structured Benchmarking Beyond Representation.}
We construct a {\em negation-centric evaluation suite spanning eight linguistically distinct categories}, enabling systematic assessment of generative compliance in the video setting.
\end{enumerate}

By reframing negation as a structured semantic constraint rather than a prompt heuristic or an embedding deficiency, this work connects linguistic structure to constraint geometry in generative modeling.

\section{Related Work}

\subsection{Negation and Compositionality in Vision-Language Models}
Negation is a fundamental yet challenging phenomenon in language understanding. Psychological studies suggest that negation is processed through a two-stage mechanism in which the affirmative representation is first activated and subsequently inhibited~\cite{kaup2006processing,kaup2007experiential}. This cognitive perspective highlights that negation is not merely the absence of affirmation, but an active semantic operation. Compositional weaknesses in modern vision-language models (VLMs) were exposed by Winoground~\cite{thrush2022winoground}, which revealed near-chance performance when matching captions containing identical words in different orders. The ARO benchmark~\cite{yuksekgonul2022and} showed that many VLMs behave like bag-of-words models, largely ignoring relational and order structure, while SugarCrepe~\cite{hsieh2023sugarcrepe} identified exploitable biases in compositional benchmarks. Building on these findings, Learn ``No'' to Say ``Yes'' Better~\cite{singh2024learn}, NegBench~\cite{alhamoud2025vision}, and NegVQA~\cite{zhang2025negvqa} demonstrated significant degradation under negated queries across retrieval and VQA settings, and CLIPGLASSES~\cite{xiao2026notjustwhatsthere} improved negation sensitivity via inference-time embedding adjustments. These efforts operate at the representation level and primarily address alignment deficiencies. In contrast, our approach does not alter representation learning. Instead, we {\em formulate linguistic negation as a dynamical feasibility constraint that directly governs diffusion guidance updates} during sampling.

\subsection{Inference-Time Control and Constrained Diffusion}
Inference-time intervention has become central to semantic control in diffusion models~\cite{bar2024lumiere,chen2023control,fei2024dysen,girdhar2311emu,khachatryan2023text2video,qing2024hierarchical,singer2022make,wang2023modelscope,weng2024art,zhang2024show,zhang2023controlvideo,henschel2024streamingt2v,qiu2023freenoise,openai,blattmann2023align,ge2023preserve,wang2024lavie,yin2023nuwa,he2022latent,blattmann2023stable,wang2023gen,chen2023seine,oh2024mevg,villegas2022phenaki,polyak2025moviegencastmedia,veo2,kong2024hunyuanvideo,wu2025hunyuanvideo}, including text-to-video systems such as Imagen Video~\cite{ho2022imagen} and Make-A-Video~\cite{singer2022make}. Training-free compositional methods manipulate cross-attention~\cite{feng2022training}, combine energy components~\cite{liu2022compositional}, or align syntax-aware attention maps~\cite{rassin2023linguistic}. Concept editing approaches, including Erasing Concepts from Diffusion Models~\cite{gandikota2023erasing} and Unified Concept Editing~\cite{gandikota2024unified}, modify model weights, often via closed-form projections, to permanently remove or alter concepts. Constrained sampling has been explored via projection or proximal updates~\cite{zhang2025constrained,zampini2025training}, and broader probabilistic frameworks such as Posterior Regularization~\cite{ganchev2010posterior} impose structured constraints during model learning. LLM-grounded Video Diffusion~\cite{lian2023llm} further injects structured language planning into video generation. While these approaches manipulate attention, combine energies, edit weights, or enforce spatial and latent constraints, negation is treated implicitly as a conditioning or suppression signal rather than as a logically defined constraint over generative dynamics. Our formulation differs in that we {\em derive a closed-form minimal correction satisfying a half-space feasibility condition in semantic guidance space}. Rather than heuristically amplifying or suppressing tokens, we {\em compute the smallest admissible update required to enforce negation consistency} at each timestep. Furthermore, unlike classical control formulations defined over physical states~\cite{beaver2024lq}, we {\em operate in semantic guidance space and derive analytic projections tailored to linguistic negation}. This enables a principled treatment of negation across both image and video diffusion backbones without retraining or architectural modification.

\section{Method}
\label{sec:method}

\begin{figure}[t]
    \centering
    \includegraphics[width=\linewidth]{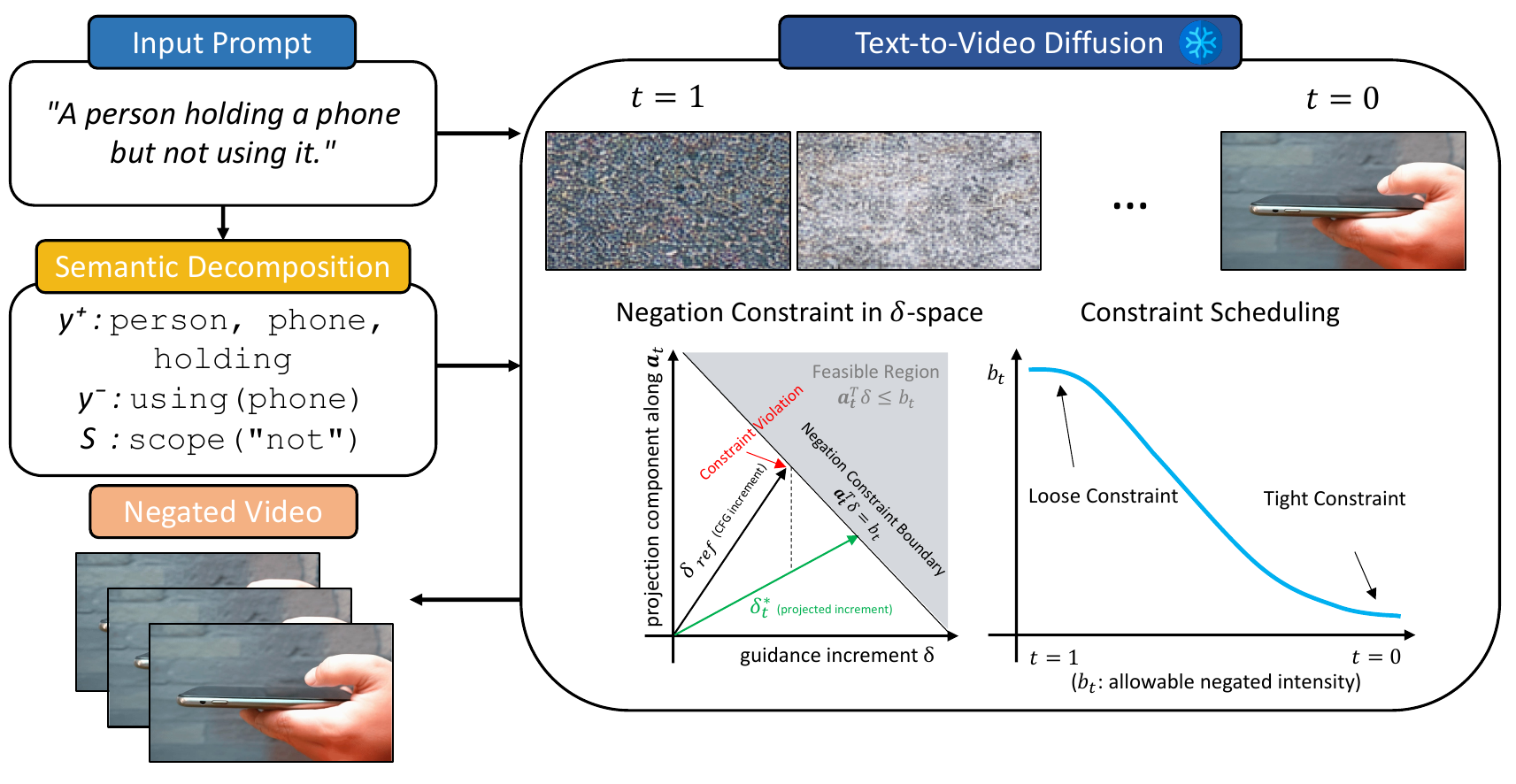}
    \caption{
    \textbf{Overview of our negation-aware framework via convex feasibility projection.}
    Given a natural-language prompt containing linguistic negation (e.g., ``a person holding a phone but not using it''), we first perform \emph{semantic decomposition} into affirmed concepts $y^{+}$, negated concepts $y^{-}$, and scope structure $\mathcal{S}$. In classifier-free guidance (CFG), the reference guidance increment is defined as  $\delta_{\mathrm{ref}} = \gamma(\epsilon_{\mathrm{text}} - \epsilon_{\mathrm{uncond}})$, which attracts the diffusion trajectory toward affirmed semantics but does not constrain negated variables. We construct a negation direction 
    $a_t = \epsilon_{\mathrm{neg}} - \epsilon_{\mathrm{uncond}}$, representing the semantic increment that increases alignment with the negated concept. At each reverse-time step ($t=1 \rightarrow 0$), we enforce a half-space constraint $a_t^\top \delta \le b_t$ in guidance space and project the reference increment onto the feasible region, producing the corrected update $\delta_t^*$. A temporal scheduling strategy progressively tightens the constraint threshold $b_t$, allowing early structural formation while enforcing strict negation at later stages. The resulting denoising trajectory yields negation-compliant video generation without retraining the text-to-video diffusion model.
    }
    \label{fig:method_figure}
\end{figure}

We formulate linguistic negation in vision-language generation as a feasibility constraint imposed on semantic guidance within pretrained diffusion dynamics. Rather than introducing explicit repulsive potentials or heuristic gradient subtraction, we compute at each timestep the minimal correction to a reference semantic guidance update required to satisfy a negation-consistent convex constraint. Our formulation separates three components: (i) pretrained diffusion dynamics, (ii) semantic guidance directions induced by language, and (iii) a feasibility operator derived from linguistic negation structure.

\subsection{Controlled Diffusion Dynamics}

Let $x_t \in \mathbb{R}^{N \times d}$ denote the latent variable at diffusion time $t \in [0,1]$, where $t=1$ corresponds to pure Gaussian noise and $t=0$ corresponds to the data sample. Sampling proceeds by integrating the reverse-time probability flow ODE
\begin{equation}
\frac{d x_t}{dt} = f_\theta(x_t,t),
\end{equation}

\noindent
from $t=1$ to $t=0$, where $f_\theta$ is a fixed pretrained diffusion backbone. The diffusion network is trained to predict the additive Gaussian noise injected during forward diffusion. We denote this prediction by

\[
\epsilon_\theta(x_t,t,c),
\]
\noindent
where $c$ is a conditioning embedding.

Let $\Phi(\cdot)$ denote the pretrained text encoder. 
Given a prompt $y$, its conditioning embedding is defined as

\[
c = \Phi(y).
\]

Classifier-free guidance (CFG) modifies the predicted noise as

\begin{equation}
\epsilon_{\text{CFG}}
=
\epsilon_{\text{uncond}}
+
\gamma(\epsilon_{\text{text}} - \epsilon_{\text{uncond}}),
\end{equation}
\noindent
where

\[
\epsilon_{\text{text}} = \epsilon_\theta(x_t,t,c),
\quad
\epsilon_{\text{uncond}} = \epsilon_\theta(x_t,t,\varnothing),
\]
\noindent
and $\gamma > 0$ is the guidance scale.

We define the reference semantic increment

\begin{equation}
\delta_{\text{ref}}
=
\gamma(\epsilon_{\text{text}} - \epsilon_{\text{uncond}}),
\end{equation}
\noindent
which represents the semantic guidance direction that increases alignment with affirmed semantics. However, this reference increment does not explicitly constrain negated concepts.

\subsection{Semantic Decomposition of Linguistic Negation}

Given a prompt $y$, we decompose it into

\[
y = (y^{+}, y^{-}, \mathcal{S}),
\]

\noindent
where $y^{+}$ encodes affirmed semantic components, $y^{-}$ encodes the linguistically grounded semantic span subject to negation or restriction, $\mathcal{S}$ encodes syntactic scope and logical composition structure.

This decomposition is obtained through a deterministic linguistic preprocessing stage operating at the span and predicate level.  
We identify negation operators or restrictive modifiers (e.g., explicit negation cues, \emph{only}-constraints, contrastive conjunctions, and logical modifiers) and extract the minimal syntactic unit governed by each operator using dependency structure.  Depending on linguistic form, the extracted unit may correspond to: (i) an object-level noun phrase (e.g., absent object), (ii) a predicate or action phrase (functional negation), (iii) a category-level restriction (implicit allow-set constraints), (iv) a temporally modulated clause, or (v) a logically composed span involving multiple operators.  Rather than enumerating object instances or rewriting prompts, we canonicalize the extracted span into a semantically coherent phrase suitable for encoding. For restrictive forms such as ``only X,'' the allow-set is interpreted as inducing a complementary semantic direction at the category level. For functional negation, predicate structure is preserved. For nested or multiple negations, $\mathcal{S}$ retains compositional information so that logical sign or sequential constraints can be consistently parameterized downstream. Let $c_{\text{neg}} = \Phi(y^{-})$ denote the conditioning embedding obtained by applying the same pretrained text encoder to the isolated negated component.

A negation branch is constructed as

\begin{equation}
a_t
=
\epsilon_{\text{neg}} - \epsilon_{\text{uncond}},
\end{equation}
\noindent
where

\[
\epsilon_{\text{neg}} = \epsilon_\theta(x_t,t,c_{\text{neg}}).
\]

All noise predictions are obtained from the same pretrained diffusion model $\epsilon_\theta(\cdot)$ without additional training. The resulting vector $a_t$ represents the local semantic direction that increases alignment with the extracted negated span under the current latent state. Importantly, $a_t$ does not correspond to an antonymic or manually inverted direction. Instead, it captures the intrinsic diffusion direction associated with the linguistic unit identified as restricted. Linguistic negation is therefore enforced by constraining motion along $a_t$, rather than replacing the concept with its opposite or introducing external heuristic penalties. This formulation ensures that diverse negation phenomena (object absence, functional suppression, category restriction, logical composition, temporal modulation, and scope disambiguation) are uniformly reduced to structured span extraction followed by convex feasibility enforcement in semantic guidance space. \textbf{Implementation details of decomposition are provided in Appendix~\ref{supp:implementation_decomposition}.}

\subsection{Negation as Convex Feasibility}

We interpret linguistic negation as restricting semantic projection along $a_t$. Define

\begin{equation}
\phi_t(\delta)
=
a_t^\top \delta.
\end{equation}
\noindent
Negation is enforced by requiring
\begin{equation}
\phi_t(\delta_t) \le b_t.
\label{eq:halfspace_extended}
\end{equation}
\noindent

This defines a closed half-space in guidance space. The bound $b_t$ determines the admissible semantic intensity of the negated concept. Different linguistic forms correspond to different constraint parameterizations: \noindent Absent Object Consistency (AOC): $b_t \le 0$, Non-Inversion Mitigation (NMI): $b_t = \tau_t$, Double Negation Sensitivity (DNS): $a_t \leftarrow \sigma_t a_t$, Scoped Negation Disambiguation (SND): $a_t = a_{t,i^*}$. Where $\sigma_t \in \{-1,1\}$ encodes logical composition and $i^*$ is determined by syntactic structure. Thus diverse linguistic phenomena reduce to a unified convex constraint on semantic projection. Standard negative prompting (NP) corresponds to a 2-branch CFG variant in which the negated concept embedding $c_{\mathrm{neg}}$ is placed in the unconditional slot, yielding the update
\begin{equation}
\epsilon_{\mathrm{NP}} = \epsilon_{\mathrm{neg}} + \gamma\,(\epsilon_{\mathrm{text}} - \epsilon_{\mathrm{neg}}).
\end{equation}
This is equivalent to a fixed, unscheduled, unconstrained subtraction along the negation direction $a_t$, applied uniformly across the entire trajectory. In contrast, our formulation (i) preserves the unconditional anchor $\epsilon_{\mathrm{uncond}}$ to maintain global semantic stability, (ii) applies a correction along $a_t$ \emph{only when} the feasibility constraint $a_t^\top \delta \le b_t$ is violated, and (iii) tightens $b_t$ adaptively over time via the schedule introduced in Section~\ref{subsec:temporal_scheduling}. Thus NP can be viewed as a degenerate parameterization of our framework with $b_t \equiv -\infty$ and no anchor preservation, which removes both the temporal modulation and the structural awareness required for compositional, scoped, or doubly negated constructions. The empirical consequences of this distinction are reported in Section~\ref{sec:quantitative_evaluation}.

\subsection{Minimal-Energy Projection}

We compute the corrected update via
\begin{equation}
\delta_t^{*}
=
\arg\min_{\delta}
\frac{1}{2}
\|\delta - \delta_{\text{ref}}\|_2^2
\quad
\text{s.t.}
\quad
a_t^\top \delta \le b_t.
\end{equation}

The Karush–Kuhn–Tucker (KKT) conditions yield:
\begin{equation}
\delta_t^{*}
=
\delta_{\text{ref}} - \lambda_t a_t,
\end{equation}
\noindent
with

\begin{equation}
\lambda_t
=
\frac{
\max\{0,\, a_t^\top \delta_{\text{ref}} - b_t\}
}{
\|a_t\|_2^2
}.
\end{equation}
\noindent
Thus projection occurs only when feasibility is violated.
The final prediction becomes
\begin{equation}
\epsilon_t^{*}
=
\epsilon_{\text{uncond}} + \delta_t^{*}.
\end{equation}

\subsection{Stability and Convergence Properties}

Because the feasible set is convex and the objective is strictly convex, the solution exists and is unique. Furthermore, projection onto a closed half-space is non-expansive:
\begin{equation}
\|\delta_t^{*} - \delta_t^{\prime *}\|
\le
\|\delta_{\text{ref}} - \delta_{\text{ref}}'\|.
\end{equation}
\noindent
This ensures Lipschitz continuity of the corrected guidance. Unlike steep repulsive potentials, the projection does not introduce stiffness into the ODE. The correction magnitude is bounded by
\begin{equation}
\|\delta_t^{*} - \delta_{\text{ref}}\|
=
\lambda_t \|a_t\|
\le
\frac{|a_t^\top \delta_{\text{ref}} - b_t|}{\|a_t\|}.
\end{equation}
\noindent
Thus no oscillatory amplification occurs.

\subsection{Temporal Scheduling}
\label{subsec:temporal_scheduling}

We define
\begin{equation}
b_t
=
(1 - \alpha_t) b_{\text{init}}
+
\alpha_t b_{\text{final}},
\end{equation}
\noindent
with

\begin{equation}
\alpha_t = \left(\frac{t}{T}\right)^p.
\end{equation}
\noindent
Early diffusion emphasizes structural formation ($b_t$ loose), while later stages enforce strict compliance.

\subsection{Unified Treatment of Linguistic Cases}

The category-specific parameterizations below operate on $(a_t,b_t)$ constructed from the parsed negated span $y^{-}$ and its scope $\mathcal{S}$. Importantly, $a_t$ is always computed from the same pretrained diffusion backbone via $c_{\mathrm{neg}}=\Phi(y^{-})$; therefore, diversity across linguistic phenomena arises from structured span extraction rather than model modification. We explicitly parameterize linguistic negation across eight structured categories, which will later form our benchmarking suite: (1) Absent Object Consistency (AOC), (2) Late Emergence Negation (LEN), (3) Implicit Natural-Only Attribute (INA), (4) Multi-Negation Composition (MNC), (5) Structural Functional Negation (SFN), (6) Non-Inversion Mitigation (NMI), (7) Double Negation Sensitivity (DNS), (8) Scoped Negation Disambiguation (SND). Each category corresponds to a distinct linguistic phenomenon but can be represented through the same convex feasibility structure by appropriate parameterization of $(a_t, b_t)$: \textbf{AOC:} $b_t \le 0$ (strict absence), \textbf{LEN:} $b_t = b_t(t)$ (time-varying bound), \textbf{INA:} $a_t \in \text{complement category space}$, \textbf{MNC:} sequential half-space projections, \textbf{SFN:} $a_t = \nabla \text{(functional attribute)}$, \textbf{NMI:} $b_t = \tau_t > 0$ (bounded moderation), \textbf{DNS:} $a_t \leftarrow \sigma a_t$ (logical sign composition), \textbf{SND:} $a_t = a_{t,i^*}$ (scope-resolved direction). Thus, although linguistically diverse, all eight negation categories reduce to convex feasibility enforcement in semantic guidance space. This unified treatment clarifies that the contribution lies not in inventing a new diffusion controller, but in formally modeling linguistic negation structure and demonstrating that diverse corner cases admit a principled projection-based solution.

\section{Experiments and Results}
\subsection{Implementation Details} 
Our method is implemented as a constraint-aware guidance formulation within diffusion sampling, without modifying or retraining the backbone model. During CFG, we compute three prediction branches: unconditional, affirmative text-conditioned, and a negation-conditioned branch derived from the input prompt via the deterministic span-extraction stage described in Section~\ref{sec:method} (full implementation in Appendix~\ref{supp:implementation_decomposition}). Instead of directly subtracting the negation prediction, we interpret negation as a soft inequality constraint on the guidance direction and perform a minimal-energy projection of the attractive CFG update onto a time-dependent half-space defined by the negation branch. The constraint strength follows a polynomial schedule $\alpha_t = (t/T)^p$ with $p=2.0$, gradually tightening semantic suppression toward later timesteps. The final bound on the negation-direction component is set to $0.0$, enforcing asymptotic removal of the forbidden concept, and numerical stabilization uses $\varepsilon = 10^{-8}$. The initial constraint bound is determined from the first-step inner product per sample and annealed thereafter. All backbone-specific parameters (e.g., inference steps, scheduler, guidance scale) follow default configurations~\cite{mochi}, and no additional learnable parameters are introduced. We instantiate the framework on Mochi as a controlled testbed in which the feasibility projection and the temporal schedule can be ablated in isolation; the formulation is defined entirely in CFG space and is invariant to backbone architecture, and we verify in Section~\ref{sec:qualitative} that the targeted failure mode is shared across HunyuanVideo and CogVideoX.

\subsection{Benchmarking Datasets}
Standard vision-language benchmarks are \textbf{insufficient for evaluating generative negation compliance.} Existing datasets such as MS-COCO, WebVid, or negation-focused retrieval benchmarks are designed to measure representation separability or caption alignment, rather than distributional behavior of generated samples under negation constraints. Negation prompts are rare and not systematically structured, and these datasets do not capture temporal emergence of forbidden concepts in video diffusion. To address this gap, we construct a controlled negation-centric evaluation suite specifically designed to probe constraint enforcement during generation. The benchmark consists of eight complementary categories targeting distinct linguistic and dynamical failure modes. All quantitative results in the main paper are averaged over this benchmark. Rather than maximizing prompt volume, our benchmark prioritizes linguistic precision and categorical coverage, with each prompt carefully designed to isolate the target negation phenomenon and prevent confounding across adjacent linguistic categories. The dataset is used exclusively for evaluation; no training or fine-tuning is performed. \textbf{Detailed dataset construction, category definitions, and examples are provided in Appendix~\ref{sec:detailed_dataset}}.

\begin{figure}[t]
    \centering
    \includegraphics[width=\linewidth]{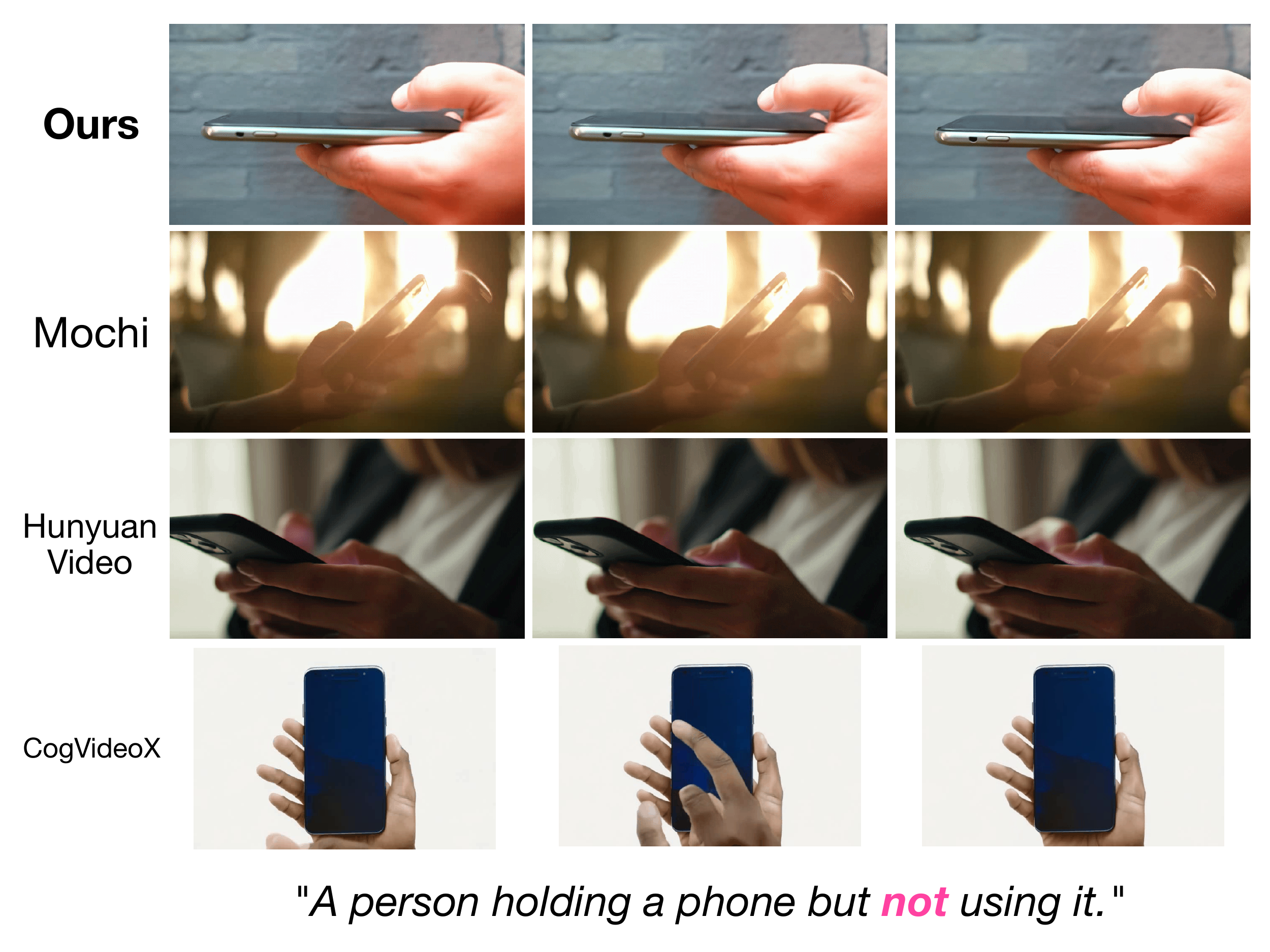}
    \caption{\textbf{Representative qualitative comparison (SFN).}
    We compare our model with state-of-the-art diffusion baselines (Mochi, HunyuanVideo, CogVideoX) on Structural Functional Negation (SFN): ``A person holding a phone but not using it.'' While baselines often collapse the negation into unintended interaction (e.g., phone-using gestures) or fail to maintain the intended constraint, our model preserves the object presence and suppresses the prohibited behavior, demonstrating negation control beyond object removal.}
    \label{fig:qualitative_main}
\end{figure}

\subsection{Qualitative Results}
\label{sec:qualitative}
Figure~\ref{fig:qualitative_main} shows a representative comparison on Structural Functional Negation (SFN), where the target object must remain present while a prohibited interaction is suppressed (``a person holding a phone but not using it''). This setting is particularly challenging because negation cannot be satisfied by simply removing the object; instead, it requires behavior-level control while preserving the overall scene structure. State-of-the-art diffusion baselines frequently produce unintended phone-using gestures or fail to maintain the negated constraint consistently. In contrast, our model enforces negation as a constraint on diffusion dynamics, preserving object presence and scene composition while suppressing the forbidden semantics. \textbf{More qualitative results across diverse negation categories and additional model comparisons are provided in Appendix~\ref{sec:detailed_qualitative}}.

\begin{table}[t]
\begin{small}
    \centering
    \resizebox{0.5\textwidth}{!}{%
    \begin{tabular}{lcccccc}
        \toprule
        \textbf{Method} & \textbf{CLIPScore} $\uparrow$ & \textbf{CLIP-neg} $\downarrow$ & \textbf{BLIP} $\uparrow$ & \textbf{DINO-conf} $\downarrow$ & \textbf{NCS} $\uparrow$ & \textbf{NVR} $\downarrow$ \\
        \midrule
        Mochi & 0.2796 & 0.2411 & 0.8183 & 0.0175 & 3.5488 & 0.3625 \\
        Negative Prompt & 0.2551 & 0.2405 & 0.8059 & \textbf{0.0128} & 3.3725 & 0.4175 \\
        HunyuanVideo & 0.2743 & 0.2620 & 0.7776 & 0.0184 & 3.0725 & 0.4400 \\
        CogVideoX & 0.2836 & 0.2442 & 0.7869 & 0.0181 & 3.3900 & 0.3800 \\
        \midrule
        \textbf{Ours (Full Model)} & \textbf{0.2924} & \textbf{0.2240} & \textbf{0.8201} & 0.0167 & \textbf{4.0700} & \textbf{0.2300} \\
        \midrule
        Ours w/o Repulsive Energy$^\dagger$ & 0.2796 & 0.2411 & 0.8183 & 0.0175 & 3.5488 & 0.3625 \\
        Ours w/o Constraint Scheduling & 0.2800 & 0.2405 & 0.7869 & 0.0174 & 3.5789 & 0.3025 \\
        \midrule
    \end{tabular}}
    \caption{\textbf{Quantitative evaluation of negation-aware video generation.} CLIPScore measures overall alignment between generated frames and the full textual prompt. CLIP-neg reports similarity between generated frames and the negated concept (lower is better). BLIP evaluates caption-level semantic alignment without modifying the prompt. DINO-conf denotes the average detection confidence of negated objects (lower indicates stronger negation compliance). The \emph{Negative Prompt} row applies standard 2-branch CFG with our extracted negated concept placed in the unconditional slot, sharing the same backbone (Mochi) and the same negation extraction as our method; it isolates the contribution of the proposed feasibility projection. $^\dagger$By construction the w/o Repulsive Energy variant does not form the negation branch, so its denoising update is identical to standard CFG and its row coincides with the Mochi baseline; it is listed to make the reduction explicit. \textbf{Refer Appendix~\ref{sec:detailed_quantitative_evaluation} for detailed analysis.}}
    \label{tab:negation_quantitative}
\end{small}
\end{table}

\subsection{Quantitative Evaluation}
\label{sec:quantitative_evaluation}
We evaluate negation-aware generation using standard, widely adopted metrics without modifying prompts or introducing threshold-based rules. Our evaluation measures (1) global prompt alignment, (2) negated concept suppression, and (3) object-level presence. Global alignment is measured using CLIPScore between generated frames and the full textual prompt, including negation phrases. Negation compliance is quantified via CLIP similarity to the explicitly negated concept (CLIP-neg), where lower values indicate stronger suppression. We further compute BLIP-based caption similarity for text-level semantic fidelity and open-vocabulary detection confidence (DINO-conf) for object-level presence of the forbidden concept. To complement embedding-based evaluation, we additionally introduce two direct vision-language metrics: the \emph{Negation Compliance Score (NCS)} and the \emph{Negation Violation Rate (NVR)}. NCS is obtained from a multimodal language model that directly reasons over sampled video frames and assigns a 1--5 compliance score, while NVR measures the empirical frequency of explicit forbidden-concept violations under the same judge. These metrics evaluate negation satisfaction at the semantic reasoning level rather than within embedding space. Detailed definitions and implementation details are provided in the Supplementary. All results are averaged over our structured benchmark. As shown in Table~\ref{tab:negation_quantitative}, our method achieves the highest CLIPScore while simultaneously reducing CLIP-neg compared to all baselines and DINO-conf relative to the Mochi backbone. Importantly, it also attains the highest NCS and the lowest NVR, confirming improved negation compliance under direct multimodal reasoning. Notably, negation-aware control improves both global alignment and constraint compliance, demonstrating that trajectory-level constraint enforcement does not degrade positive semantic fidelity. \textbf{More quantitative analysis are provided in Appendix~\ref{sec:detailed_quantitative_evaluation}}.

\begin{table}[t]
\begin{small}
    \centering
    \resizebox{0.5\textwidth}{!}{%
    \begin{tabular}{lcccccc}
        \toprule
        \textbf{Method} & \textbf{SC} $\uparrow$ & \textbf{BC} $\uparrow$ & \textbf{TF} $\uparrow$ & \textbf{DD} & \textbf{AQ} $\uparrow$ & \textbf{IQ} $\uparrow$ \\
        \midrule
        Mochi            & 0.9772 & 0.9580 & 0.9843 & 0.2600 & 0.5511 & 0.6217 \\
        Negative Prompt  & 0.9704 & 0.9524 & 0.9832 & 0.2600 & 0.5280 & 0.6156 \\
        HunyuanVideo     & 0.9846 & 0.9780 & 0.9917 & 0.2275 & 0.5534 & 0.6438 \\
        CogVideoX        & 0.9583 & 0.9612 & 0.9680 & 0.6725 & 0.4860 & 0.5712 \\
        \midrule
        \textbf{Ours}    & 0.9761 & 0.9583 & 0.9851 & 0.2725 & 0.5405 & 0.6172 \\
        \midrule
        Ours w/o Repulsive Energy$^\dagger$ & 0.9744 & 0.9583 & 0.9842 & 0.2625 & 0.5481 & 0.6220 \\
        Ours w/o Constraint Scheduling      & 0.9677 & 0.9521 & 0.9830 & 0.3000 & 0.5093 & 0.5963 \\
        \bottomrule
    \end{tabular}}
    \caption{\textbf{Reference-free perceptual quality (VBench).} Six VBench quality dimensions evaluated on the same generated videos in custom-input mode. SC, subject consistency; BC, background consistency; TF, temporal flickering; DD, dynamic degree; AQ, aesthetic quality; IQ, imaging quality (normalized to $[0,1]$). DD measures the share of videos with large motion and carries no preferred direction. $^\dagger$This variant reduces analytically to standard CFG, so its deviation from the Mochi row reflects run-to-run sampling variance.}
    \label{tab:negation_vbench}
\end{small}
\end{table}

Negation compliance would mean little if it came at the cost of perceptual quality, so Table~\ref{tab:negation_vbench} additionally reports reference-free quality scores from VBench~\cite{huang2024vbench} on the same generated videos, covering six quality dimensions evaluated in custom-input mode. The w/o Repulsive Energy variant reduces analytically to standard CFG, so its deviation from the Mochi row estimates run-to-run sampling variance, and the gaps between Ours and Mochi on subject consistency, background consistency, and temporal flickering all lie within that variance. Aesthetic quality is $0.011$ lower and imaging quality $0.005$ lower, which follows from compliance itself. A clip that correctly omits the forbidden object depicts a sparser scene, and frame-level quality predictors reward the content-rich completions that a violating clip restores. The two degraded rows confirm that this near-parity is a property of the constrained projection rather than of any intervention. Negative prompting lowers subject consistency, background consistency, and aesthetic quality together, and removing the constraint schedule produces the largest quality drop in the table, mirroring at the perceptual level the compliance losses both variants show in Table~\ref{tab:negation_quantitative}. These scores are consistent with the Artifact Avoidance ratings of the user study (Figure~\ref{fig:user_study}), where our method also ranked highest.

\begin{figure}[htb!]
\centering
\includegraphics[width=\linewidth]{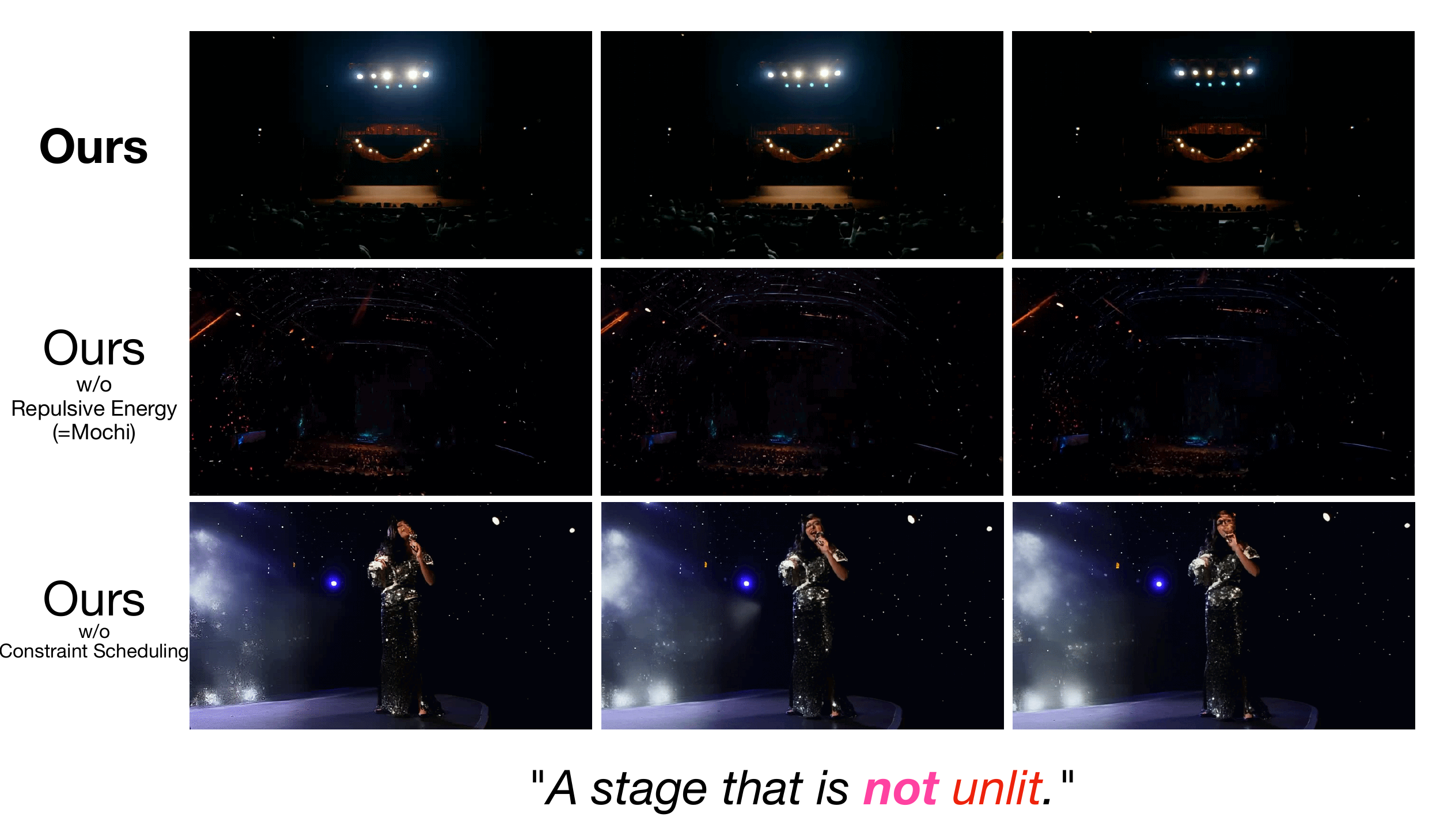}
\caption{\textbf{Ablation study on negation control.} We ablate both components on a double-negation prompt (DNS), ``A stage that is not unlit,'' whose correct reading is a lit stage. The full model produces a stable, correctly lit stage. \textit{w/o Repulsive Energy} disables the negation branch and is analytically equivalent to standard CFG, i.e., to the Mochi baseline; the stage remains dark, violating the double negation. \textit{w/o Constraint Scheduling} keeps the projection active but applies it uniformly across timesteps, disrupting early structure formation and causing scale drift and unnatural light-source placement. Baseline comparisons against all backbones are provided in Section~\ref{sec:qualitative}; quantitative counterparts are reported in Table~\ref{tab:negation_quantitative}.}
\label{results:ablation}
\end{figure}

\subsection{Ablation Study} 
We ablate our two core components: (i) the repulsive constraint term and (ii) time-dependent constraint scheduling. The repulsive term determines \emph{what} semantic direction is suppressed, while scheduling controls \emph{when} suppression is enforced along the diffusion trajectory. The w/o Repulsive Energy variant does not construct the negation branch and therefore reduces analytically to standard CFG; its row in Table~\ref{tab:negation_quantitative} coincides with the Mochi baseline by construction, and the gap between this row and the full model (NCS $3.5488 \to 4.0700$, NVR $0.3625 \to 0.2300$) measures the net contribution of the feasibility projection. Removing constraint scheduling, a non-degenerate variant in which the projection remains active with a fixed bound, degrades both global alignment and negation stability, reducing CLIPScore and increasing CLIP-neg, with the same trend under direct vision-language evaluation (NCS falls to $3.5789$, NVR rises to $0.3025$). This indicates that temporally modulated constraint strength is essential to prevent interference with early structural formation while ensuring late-stage suppression. Figure~\ref{results:ablation} illustrates both configurations, with the w/o Repulsive Energy row doubling as the Mochi baseline. \textbf{More detailed quantitative and qualitative analysis are provided in Appendix~\ref{sec:detailed_ablation}}.

\subsection{User Study}

\begin{figure}[t]
    \centering
    \includegraphics[width=\linewidth]{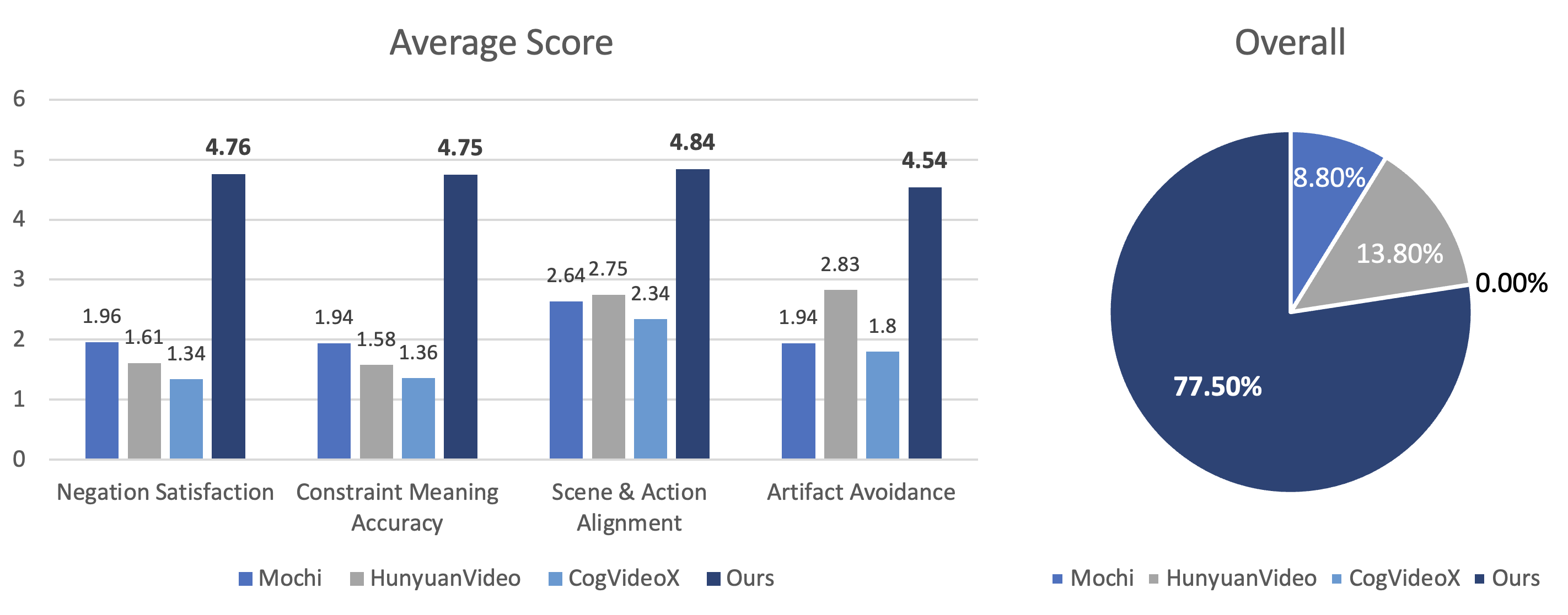}
    \caption{\textbf{User study results for negation-aware video generation.} 50 participants evaluated four anonymized methods across four criteria: \emph{Negation Satisfaction}, \emph{Constraint Meaning Accuracy}, \emph{Scene \& Action Alignment}, and \emph{Artifact Avoidance} (left). Our method consistently achieved the highest average ratings across all dimensions (4.54–4.84), substantially outperforming Mochi, HunyuanVideo, and CogVideoX. In the overall preference comparison (right), \textbf{77.5\%} of votes favored our method, compared to 13.8\% for HunyuanVideo, 8.8\% for Mochi, and 0.0\% for CogVideoX.}
    \label{fig:user_study}
\end{figure}

We conducted a user study with 50 participants to assess perceptual quality and negation compliance in video generation. For each of eight prompts, participants were shown four anonymized results generated by Mochi, HunyuanVideo, CogVideoX, and \emph{Ours}, with presentation order randomized per question to prevent positional bias. Participants rated each output on four 5-point Likert criteria: (1) \emph{Negation Satisfaction}, (2) \emph{Constraint Meaning Accuracy}, (3) \emph{Scene \& Action Alignment}, and (4) \emph{Artifact Avoidance}, and additionally selected an overall preferred result. As illustrated in Figure~\ref{fig:user_study}, our method achieved the highest mean scores across all axes, reaching 4.76 (negation satisfaction), 4.75 (constraint meaning accuracy), 4.84 (scene alignment), and 4.54 (artifact avoidance), while competing methods remained near the disagreement range (1.34–2.83). In overall preference, \textbf{77.5\%} of votes favored \emph{Ours}, indicating a clear perceptual advantage over existing methods. These findings suggest that modeling linguistic negation as a constrained diffusion process substantially improves semantic compliance and visual fidelity. \textbf{See Appendix~\ref{supplementary:user_study} for full user study details.}

\section{Conclusions}
We formulated linguistic negation as a structured feasibility condition on the classifier-free guidance update: a deterministic decomposition extracts the negated span, its direction in guidance space defines a half-space constraint, and a scheduled minimal-energy projection enforces the constraint during sampling. The formulation covers diverse negation phenomena (object absence, structural functional negation, graded non-inversion semantics, double negation sensitivity, and scoped negation disambiguation) within a single constraint template, requires no retraining or architectural change, and improves negation compliance on our eight-category benchmark while preserving global prompt fidelity, as confirmed by embedding metrics, a vision-language judge, and a 50-participant user study. Because the correction is defined on the guidance update alone, it carries over to other backbones, although our controlled quantitative study is confined to Mochi. Extending the constraint family to other linguistic operators, such as quantification and modality, is a natural next step.

\section{Limitations and Future Work}
\label{sec:limitation}
Our current formulation approximates linguistic constraints within a tractable geometric structure over semantic update directions. Certain forms of pragmatically inferred or world-knowledge-dependent negation may require richer semantic parsing and non-linear constraint modeling. Additionally, constraint effectiveness depends on the fidelity of the underlying vision-language representation space. We evaluate the proposed projection on Mochi, which serves as a controlled testbed where the feasibility correction and the temporal schedule can be ablated in isolation; the formulation itself is defined exclusively in classifier-free guidance space and is invariant to backbone architecture, and Appendix~\ref{supp:backbone} elaborates on this scope together with the cross-backbone evidence, drawn from our comparisons against HunyuanVideo and CogVideoX, that establishes negation insensitivity as a property of the broader T2V model class. The NCS/NVR judge is queried with the negated concept produced by the same extraction module the method uses, so extraction errors shared by the method and the judge are not penalized; the user study provides a human check, but judge-independent validation of the scope-sensitive categories remains future work. Preliminary exploration on other backbones further indicates that per-backbone calibration of the guidance and schedule parameters is more involved than matching guidance scales alone, so we confine quantitative claims to Mochi. Future work may extend this formulation beyond negation to broader classes of linguistic operators, including quantification, modality, and compositional logical structure. Integrating formal semantic representations with neural trajectory control could further bridge symbolic linguistic theory and large-scale generative models.

{
    \small
    \bibliographystyle{ieeenat_fullname}
    \bibliography{main}
}

\newpage
\onecolumn
\appendix

\section{Ethical Statement}
\label{supplementary:ethics}

\begin{tcolorbox}[breakable, title=Ethical Statement]
This work proposes a training-free, constraint-based formulation of linguistic negation for vision-language generation by projecting diffusion guidance updates onto a feasible set derived from prompt structure, with the goal of improving logical faithfulness and safety-relevant exclusion behavior in text-to-image and text-to-video systems; as with all generative models, the approach may be misused to create deceptive or harmful media, and improved controllability should not be conflated with policy compliance, since harmful content can still be produced if prompted, so the method is not a universal safety solution but a mechanism for enforcing structured constraints when user intent and downstream policy are aligned; because our framework operates entirely on frozen pretrained backbones, it inherits representational biases and uneven concept coverage from underlying models and datasets, and constraint effectiveness may vary across concepts or contexts, so improvements in negation compliance do not imply fairness guarantees; we introduce a structured negation-centric benchmark used exclusively for evaluation (not training), designed to probe linguistic structure rather than elicit sensitive or targeted content, and we recommend that future extensions avoid prompts involving protected groups, disallowed activities, or identity-based inference; the reported user study involved adult participants (18+) recruited through open online channels, participation was voluntary and anonymous, no personally identifying information was collected, results are reported only in aggregate, and the study protocol was reviewed by University of Maryland's Institutional Review Board (IRB) (Kuali-IRB Number \#1179-1) and determined to qualify for exemption (45 CFR Part 46 -- Protection of Human Subjects) under applicable regulations; from an environmental and transparency perspective, we report computation time and peak memory consumption for all configurations, and because our approach introduces no additional training and operates at inference time without architectural modification, it does not incur retraining overhead beyond standard diffusion inference; overall, we believe that trajectory-level modeling of linguistic operators can improve reliability and constraint consistency in generative systems, while emphasizing that responsible deployment requires appropriate content policies, safeguards, and auditing procedures. Regarding licensing of scientific artifacts, all video diffusion backbones used in this work are publicly released for research use: Mochi 1~\cite{mochi} is distributed under the Apache 2.0 license, CogVideoX~\cite{yang2025cogvideox} is distributed under the Apache 2.0 license, and HunyuanVideo~\cite{kong2024hunyuanvideo} is distributed under the Tencent Hunyuan Community License, which permits research use within its specified territorial scope; all auxiliary evaluation tools (CLIP and BLIP) are released under permissive open-source licenses (MIT and BSD-3-Clause, respectively) and are used here solely for non-commercial academic evaluation consistent with their intended research purpose; we do not redistribute any pretrained model weights, and our use is restricted to inference-time evaluation without retraining or fine-tuning; the negation-centric benchmark introduced in this work consists exclusively of authored English-language text prompts containing no personal data, identifying information, or sensitive content, and we release it under the CC BY-NC 4.0 license for non-commercial research use, consistent with the upstream licensing of the diffusion backbones on which it is evaluated;
\end{tcolorbox}

\section{Detailed Benchmarking Datasets}
\label{sec:detailed_dataset}

Unlike prior works that evaluate negation understanding in vision-language representations, our objective is to assess \emph{negation compliance in generative diffusion trajectories}. This distinction fundamentally changes the role of datasets in our evaluation protocol.

\paragraph{Why Standard Datasets Are Insufficient.}
Conventional vision-language benchmarks such as MS-COCO, CC, or WebVid are designed for descriptive captioning and semantic alignment. Negation-specific prompts are \textbf{rare, incidental, and not systematically structured.} More importantly, these datasets are intended to evaluate \textbf{representation separability or retrieval consistency}, rather than distributional behavior of generated samples under negation constraints. 

In representation learning settings, the evaluation question typically takes the form:
\[
S(I, T_{\text{pos}}) > S(I, T_{\text{neg}})
\]
which measures embedding separability. In contrast, our problem is generative and distributional:
\[
\mathbb{P}_{x_0 \sim p_\theta(\cdot \mid y)}\big[ \text{negated concept appears} \big].
\]
Standard datasets such as NegCLIP, CC-Neg, and ARO \textbf{do not provide structured prompts to measure this probability, nor do they capture temporal emergence of forbidden content in video diffusion.} Therefore, using them directly would neither test nor reveal the core property studied in this work: constraint enforcement during generation.

For this reason, we construct a controlled negation-centric evaluation suite designed explicitly to probe generative behavior under negation constraints. Importantly, this dataset is used solely for evaluation; no training or fine-tuning is performed.

\paragraph{Dataset Scale.}
For each of the eight categories described below, we construct 50 prompt pairs, resulting in a total of 400 evaluation sets. Each set consists of structured prompts designed to probe negation compliance under controlled conditions. Rather than maximizing prompt volume, our benchmark prioritizes linguistic precision and categorical coverage. Each of the 50 prompts per category is carefully designed to isolate the target negation phenomenon, preventing confounding factors from adjacent linguistic categories. All quantitative results reported in the main paper are computed over this benchmark. The dataset is used exclusively for evaluation; no training or fine-tuning is performed.

\paragraph{Design Principles.}
Our benchmarking suite is built around eight complementary categories, each targeting a distinct dynamical or linguistic failure mode observed in text-to-video diffusion models. The categories are constructed to (1) isolate negation as a primary semantic constraint, (2) prevent trivial prompt-engineering solutions, (3) expose trajectory-level violations that may emerge temporally, and (4) disentangle linguistic phenomena such as graded negation, compositional structure, and scope ambiguity from purely object-level suppression.

\subsubsection*{(1) AOC: Absent Object Consistency}

The Absent Object Consistency (AOC) set evaluates whether explicitly negated objects remain absent throughout the entire video. Prompts follow the structure:
\begin{quote}
``A [scene] with no [object].''
\end{quote}
Representative example:
\begin{quote}
``A highway at sunset, with \textbf{no} vehicles.'' \\
``A city skyline at night, with \textbf{no} lights on.'' \\
``A snowfield, with \textbf{no} footprints.''

\end{quote}

In video diffusion models, forbidden objects frequently emerge in later frames even if absent initially. This set directly measures temporal consistency of negation compliance. It isolates the simplest but most critical form of negation: object absence.

\subsubsection*{(2) LEN: Late Emergence Negation}

The Late Emergence Negation (LEN) set targets a common trajectory-level failure mode: negated content appearing only after structural formation. Prompts introduce gradual scene evolution:
\begin{quote}
``A concert stage lighting up, with \textbf{no} performers.'' \\
``A ski slope after snowfall, with \textbf{no} ski tracks appearing.'' \\
``A dock at dusk, with \textbf{no} ropes appearing.''
\end{quote}

Such prompts test whether time-dependent constraint scheduling successfully prevents semantic drift during later diffusion stages. This category is particularly important for video models, where temporal hallucination is prevalent.

\subsubsection*{(3) INA: Implicit Natural-only Attribute}

The Implicit Natural-only Attribute (INA) set evaluates negation under \emph{semantic exclusivity} rather than explicit lexical negation. Instead of prompts of the form ``with no [object],'' INA specifies a closed category whose elements are exhaustive. Representative example:

\begin{quote}
``A scene featuring \textbf{only} natural elements.'' \\
``A meadow under the night sky with \textbf{only} grass and stars.'' \\
``A natural cave interior featuring \textbf{only} stone and shadows.''
\end{quote}

In this formulation, exclusion must be inferred from the semantic scope of ``only.'' The prompt implicitly prohibits humans, vehicles, buildings, and other artificial structures without enumerating them. This design tests whether the model suppresses objects that are statistically likely under scene priors but logically disallowed by category closure.

Formally, let $\mathcal{C}_{\text{allowed}}$ denote the semantically closed set implied by the prompt and $\mathcal{C}_{\text{prior}}$ denote high-probability co-occurring objects under the generative prior. In many natural scenes,
\[
\mathcal{C}_{\text{prior}} \not\subset \mathcal{C}_{\text{allowed}},
\]
creating a conflict between prior likelihood and semantic constraint. INA evaluates whether the model suppresses objects in 
\[
\mathcal{C}_{\text{prior}} \setminus \mathcal{C}_{\text{allowed}},
\]
without relying on explicit negation markers.

Unlike AOC, which measures explicit object absence, INA probes the model's ability to enforce logically implied exclusions under strong co-occurrence bias.

\subsubsection*{(4) MNC: Multi-Negation Composition}

The Multi-Negation Composition (MNC) set evaluates compositional robustness under multiple simultaneous constraints:
\begin{quote}
``A classroom scene with \textbf{no} students, \textbf{no} teacher, and \textbf{no} books.'' \\
``A mountain cabin with \textbf{no} doors, \textbf{no} windows, and \textbf{no} chimney.'' \\
``A snow-covered road with \textbf{no} tire tracks, \textbf{no} signs, and \textbf{no} vehicles.''
\end{quote}

This setting probes whether repulsive energies combine stably without degrading generation quality or causing constraint collapse. Representation-based negation learning often struggles under multiple exclusions due to similarity conflicts; MNC tests the stability of trajectory-level control under compounded constraints.

\subsubsection*{(5) SFN: Structural Functional Negation}

Structural Functional Negation (SFN) evaluates negation over actions or states rather than object existence:
\begin{quote}
``A person holding a phone but \textbf{not} using it.'' \\
``A clock ticking but \textbf{not} moving its hands.'' \\
``A factory assembly line active but \textbf{not} producing items.''
\end{quote}

Here, the object must remain present while a specific action is suppressed. This category differentiates between simple object removal and higher-order semantic control. It tests whether negation-aware diffusion can modulate behavior-level attributes without eliminating the underlying entity.

\subsubsection*{(6) NMI: Non-Inversion Mitigation}

The Non-Inversion Mitigation (NMI) set evaluates cases where negation does not correspond to strict semantic inversion. In natural language, expressions such as ``not bad'' or ``not uncommon'' do not necessarily imply the opposite extreme (e.g., ``excellent'' or ``rare''), but instead express moderated or context-dependent meanings.

\begin{quote}
``A dog that is \textbf{not} aggressive.'' \\
``A harbor that is \textbf{not} fully occupied.'' \\
``A mountain peak that is \textbf{not} completely snow-covered.''
\end{quote}

These prompts require the model to attenuate specific attributes rather than completely eliminate or invert them. A naive repulsive formulation may over-correct, producing semantically exaggerated opposites (e.g., empty beach, extremely casual meal). 

Formally, this category probes whether negation enforcement behaves as a \emph{bounded constraint} rather than a binary complement operation. Let $A$ denote a continuous attribute (e.g., crowdedness). Instead of enforcing $A = 0$, NMI requires:
\[
A(x_t) \le \tau,
\]
where $\tau$ represents a moderation threshold rather than strict absence. This category evaluates whether the generative trajectory respects graded semantic constraints without collapsing to degenerate extremes.

\subsubsection*{(7) DNS: Double Negation Sensitivity}

The Double Negation Sensitivity (DNS) set evaluates prompts containing multiple nested or compositional negations. In natural language, double negation may either cancel (logical double negation) or intensify negation depending on linguistic context.

\begin{quote}
``A stage that is \textbf{not} \textit{unlit}.'' \\
``A beach that is \textbf{not} shell-\textit{less}.'' \\
``A rooftop that is \textbf{not} vent-\textit{less}.''
\end{quote}

Primary negation cues are highlighted in \textbf{bold}, while secondary negation operators (e.g., \textit{unlit}, \textit{-less}) are shown in \textit{italics} to illustrate compositional negation structure.

In these cases, naive repulsive composition may lead to attraction–repulsion conflicts, over-suppression, or unstable semantic oscillation. DNS tests whether the control mechanism correctly resolves nested negation structure without generating contradictory or degenerate outputs.

From a constraint perspective, this set examines whether sequential negations yield consistent feasible regions rather than competing gradient directions. It specifically probes robustness of the minimal-energy correction mechanism under sign-sensitive semantic composition.

\subsubsection*{(8) SND: Scoped Negation Disambiguation}

The Scoped Negation Disambiguation (SND) set evaluates the model’s ability to correctly identify the semantic scope of negation within structurally ambiguous sentences. Unlike SFN, which suppresses a specific action while preserving object presence, SND focuses on determining \emph{which component} of the sentence is negated.

\begin{quote}
``A teacher \textit{helping} a student who is \textbf{not} paying attention.'' \\
``A boat anchored near an island that is \textbf{not} \textit{inhabited}.'' \\
``A robot next to a screen that is \textbf{not} \textit{displaying content.}''
\end{quote}
Negation cues are highlighted in \textbf{bold}, while the negated semantic targets are shown in \textit{italics}.

In these prompts, negation applies to a specific subordinate clause or attribute. Incorrect scope resolution may lead to suppression of the wrong entity (e.g., removing the dog instead of stopping its motion).

Formally, let $g_i(x_t)$ denote candidate semantic components. Scope detection determines which $g_i$ participates in the constraint:
\[
g_{i^*}(x_t) \le \varepsilon,
\]
while leaving other components unconstrained. This dataset isolates linguistic disambiguation from functional negation (SFN) by requiring correct identification of the constrained semantic variable rather than merely suppressing an action.

\paragraph{Summary.}
Together, AOC, LEN, INA, MNC, SFN, NMI, DNS, and SND form a structured evaluation suite that probes multiple dimensions of negation-aware generation, including object absence, temporal stability, implicit semantic exclusion, compositional robustness, functional suppression, graded non-inversion semantics, nested negation consistency, and scope-sensitive constraint identification. These categories are deliberately designed to reflect distinct dynamical and linguistic failure modes in generative diffusion models. By isolating these complementary dimensions, the benchmark provides a targeted and interpretable evaluation of constraint-consistent diffusion trajectory control under both explicit and structurally complex negation.

\begin{figure*}[t]
    \centering
    \includegraphics[width=0.8\linewidth]
    {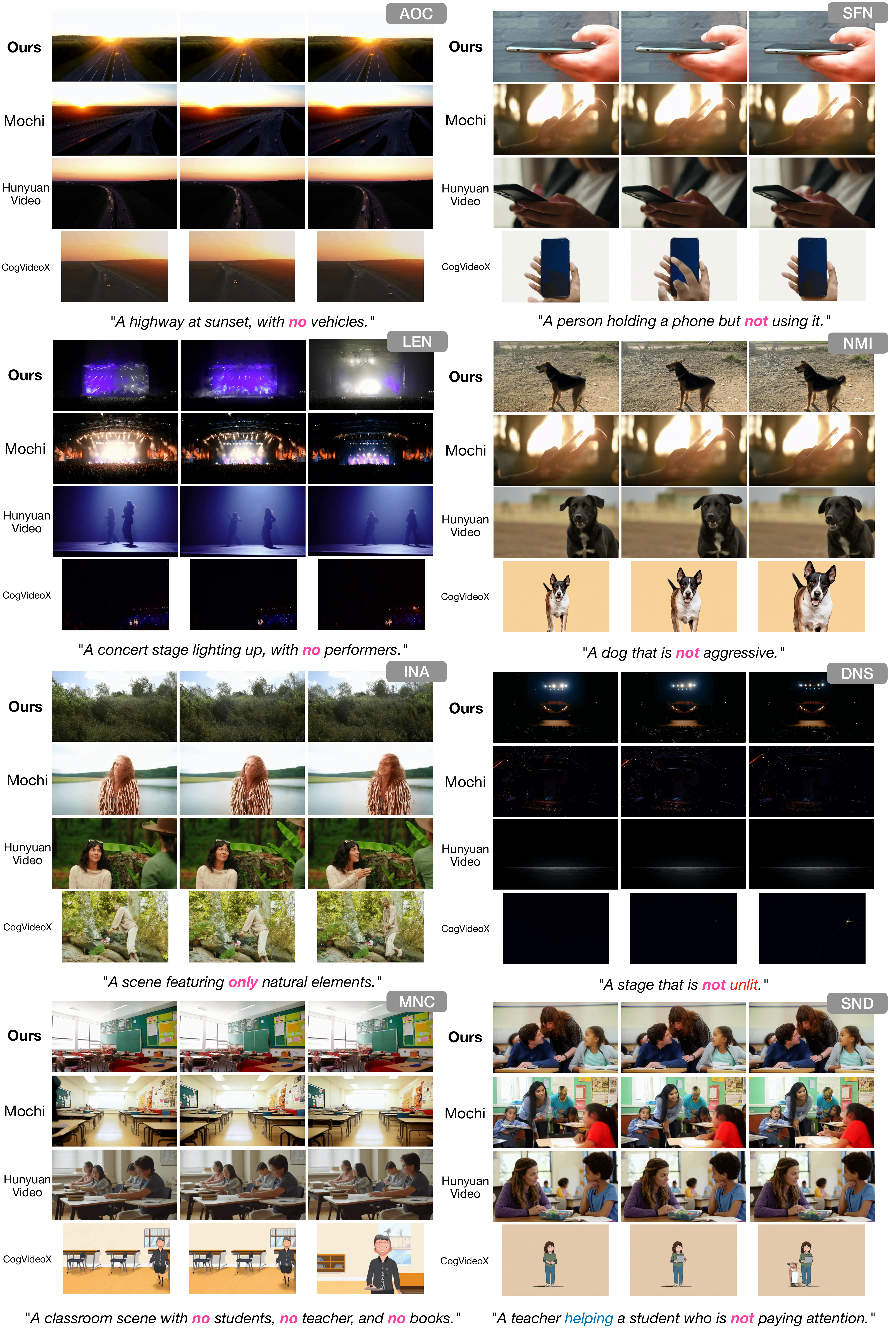}
    \caption{\textbf{Qualitative comparison across diverse negation scenarios.} We evaluate our model against state-of-the-art diffusion models (Mochi, HunyuanVideo, CogVideoX) on eight negation settings: AOC (Absent Object Consistency), LEN (Late Emergence Negation), INA (Implicit Natural-Only Attribute), MNC (Multi-Negation Composition), SFN (Structural Functional Negation), NMI (Non-Inversion Mitigation), DNS (Double Negation Sensitivity), and SND (Scoped Negation Disambiguation). Across simple object removal, functional suppression, logical inversion, and scope resolution, our model consistently preserves scene structure while correctly enforcing negation constraints.}
    \label{fig:detailed_qualitative}
\end{figure*}

\begin{figure*}[t]
    \centering
    \includegraphics[width=0.8\linewidth]
    {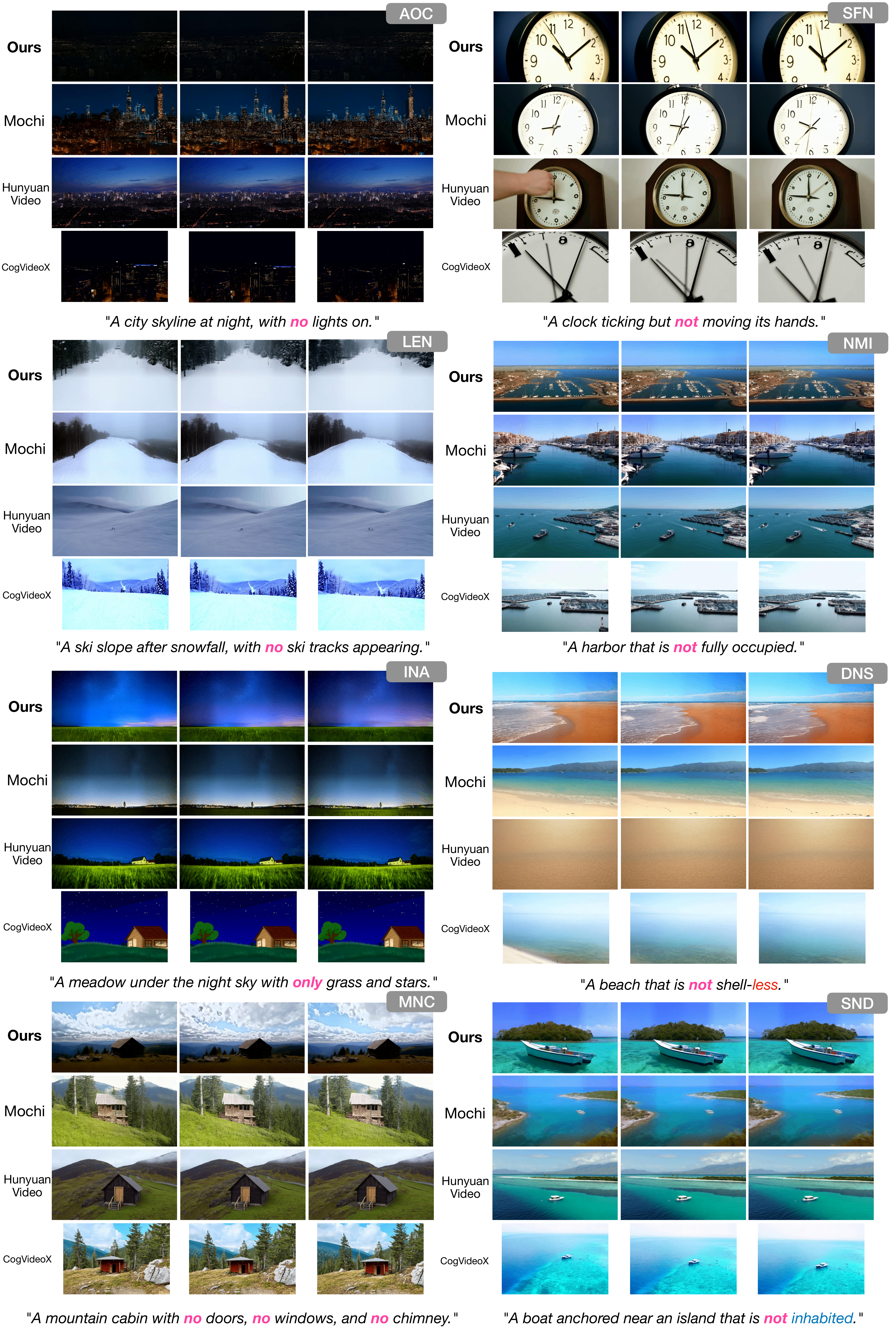}
    \caption{\textbf{Additional qualitative comparisons across diverse negation scenarios (Set 1).} We compare our method with diffusion-based video generation models (Mochi, HunyuanVideo, CogVideoX) on eight negation categories: AOC (Absent Object Consistency), LEN (Late Emergence Negation), INA (Implicit Natural-Only Attribute), MNC (Multi-Negation Composition), SFN (Structural Functional Negation), NMI (Non-Inversion Mitigation), DNS (Double Negation Sensitivity), and SND (Scoped Negation Disambiguation). Each row shows a negation prompt that requires suppressing objects, attributes, or functional behaviors while preserving the scene structure. Our method consistently enforces the negation constraints while maintaining semantic coherence across diverse scenes.}
    \label{fig:detailed_qualitative_supp1}
\end{figure*}

\begin{figure*}[t]
    \centering
    \includegraphics[width=0.8\linewidth]
    {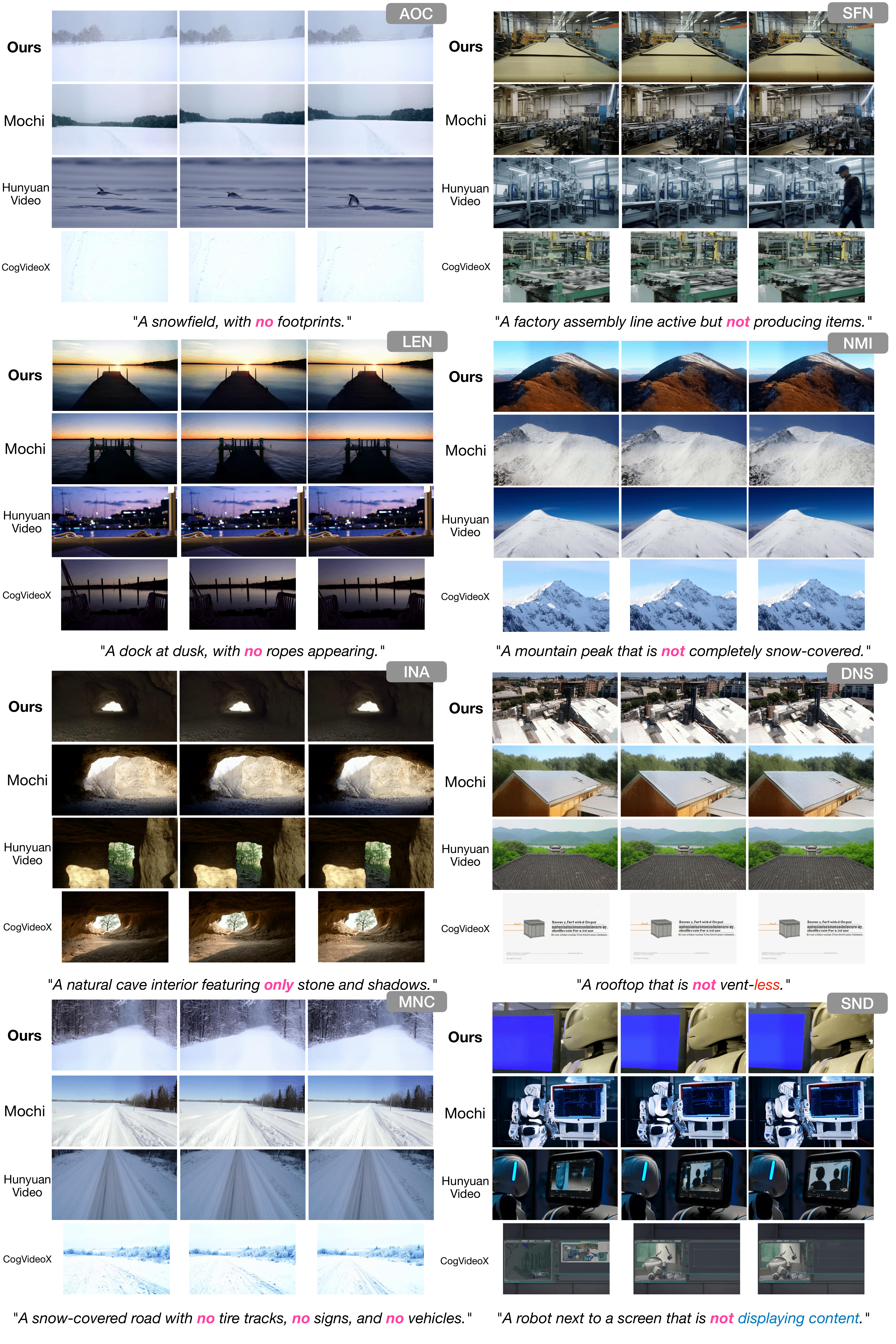}
    \caption{\textbf{Additional qualitative comparisons across diverse negation scenarios (Set 2).} This figure presents a second set of prompts covering the same eight negation categories as in Figure~\ref{fig:detailed_qualitative}, including AOC, LEN, INA, MNC, SFN, NMI, DNS, and SND. These examples further highlight the robustness of our method across different scene contexts, ranging from natural landscapes and urban environments to structured indoor scenes and human-object interactions. While baseline models frequently violate negation constraints by introducing forbidden objects or attributes, our method reliably suppresses such elements while preserving scene realism and structural consistency.}
    \label{fig:detailed_qualitative_supp2}
\end{figure*}

\section{Detailed Qualitative Results}
\label{sec:detailed_qualitative}

\begin{table*}[t]
\begin{small}
    \centering
    \resizebox{\textwidth}{!}{%
    \begin{tabular}{lcccccc}
        \toprule
        \textbf{Method} & \textbf{CLIPScore} $\uparrow$ & \textbf{CLIP-neg} $\downarrow$ & \textbf{BLIP} $\uparrow$ & \textbf{DINO-conf} $\downarrow$ & \textbf{NCS} $\uparrow$ & \textbf{NVR} $\downarrow$ \\
        \midrule
        Mochi & 0.2796 $\pm$ 0.0252 & 0.2411 $\pm$ 0.0288 & 0.8183 $\pm$ 0.0774 & 0.0175 $\pm$ 0.0314 & 3.5488 $\pm$ 1.8639 & 0.3625 $\pm$ 0.4813 \\
        Negative Prompt & 0.2551 $\pm$ 0.0370 & 0.2405 $\pm$ 0.0225 & 0.8059 $\pm$ 0.0786 & \textbf{0.0128 $\pm$ 0.0169} & 3.3725 $\pm$ 1.6780 & 0.4175 $\pm$ 0.4937 \\
        HunyuanVideo & 0.2743 $\pm$ 0.0360 & 0.2620 $\pm$ 0.0388 & 0.7776 $\pm$ 0.1018 & 0.0184 $\pm$ 0.0268 & 3.0725 $\pm$ 1.8778 & 0.4400 $\pm$ 0.4970 \\
        CogVideoX & 0.2836 $\pm$ 0.0320 & 0.2442 $\pm$ 0.0414 & 0.7869 $\pm$ 0.0984 & 0.0181 $\pm$ 0.0269 & 3.3900 $\pm$ 1.8873 & 0.3800 $\pm$ 0.4859 \\
        \midrule
        \textbf{Ours (Full Model)} & \textbf{0.2924 $\pm$ 0.0225} & \textbf{0.2240 $\pm$ 0.0314} & \textbf{0.8201 $\pm$ 0.0729} & 0.0167 $\pm$ 0.0278 & \textbf{4.0700 $\pm$ 1.6406} & \textbf{0.2300 $\pm$ 0.4213} \\
        \midrule
        Ours w/o Repulsive Energy$^\dagger$ & 0.2796 $\pm$ 0.0252 & 0.2411 $\pm$ 0.0288 & 0.8183 $\pm$ 0.0774 & 0.0175 $\pm$ 0.0314 & 3.5488 $\pm$ 1.8639 & 0.3625 $\pm$ 0.4813 \\
        Ours w/o Constraint Scheduling & 0.2800 $\pm$ 0.0305 & 0.2405 $\pm$ 0.0477 & 0.7869 $\pm$ 0.0881 & 0.0174 $\pm$ 0.0245 & 3.5789 $\pm$ 1.8439 & 0.3025 $\pm$ 0.4599 \\
        \midrule
    \end{tabular}}
    \caption{\textbf{Detailed quantitative evaluation with mean and standard deviation.} 
    All metrics are reported as mean $\pm$ standard deviation over the full evaluation set. 
    NCS denotes the Negation Compliance Score obtained from a direct vision-language judge, and NVR denotes the Negation Violation Rate computed from the judge's binary forbidden-presence decision. $^\dagger$By construction the w/o Repulsive Energy variant does not form the negation branch, so its denoising update is identical to standard CFG and its row coincides with the Mochi baseline; it is listed to make the reduction explicit.}
    \label{tab:negation_quantitative_supp}
\end{small}
\end{table*}

\begin{figure*}[t]
    \centering
    \includegraphics[width=\linewidth]
    {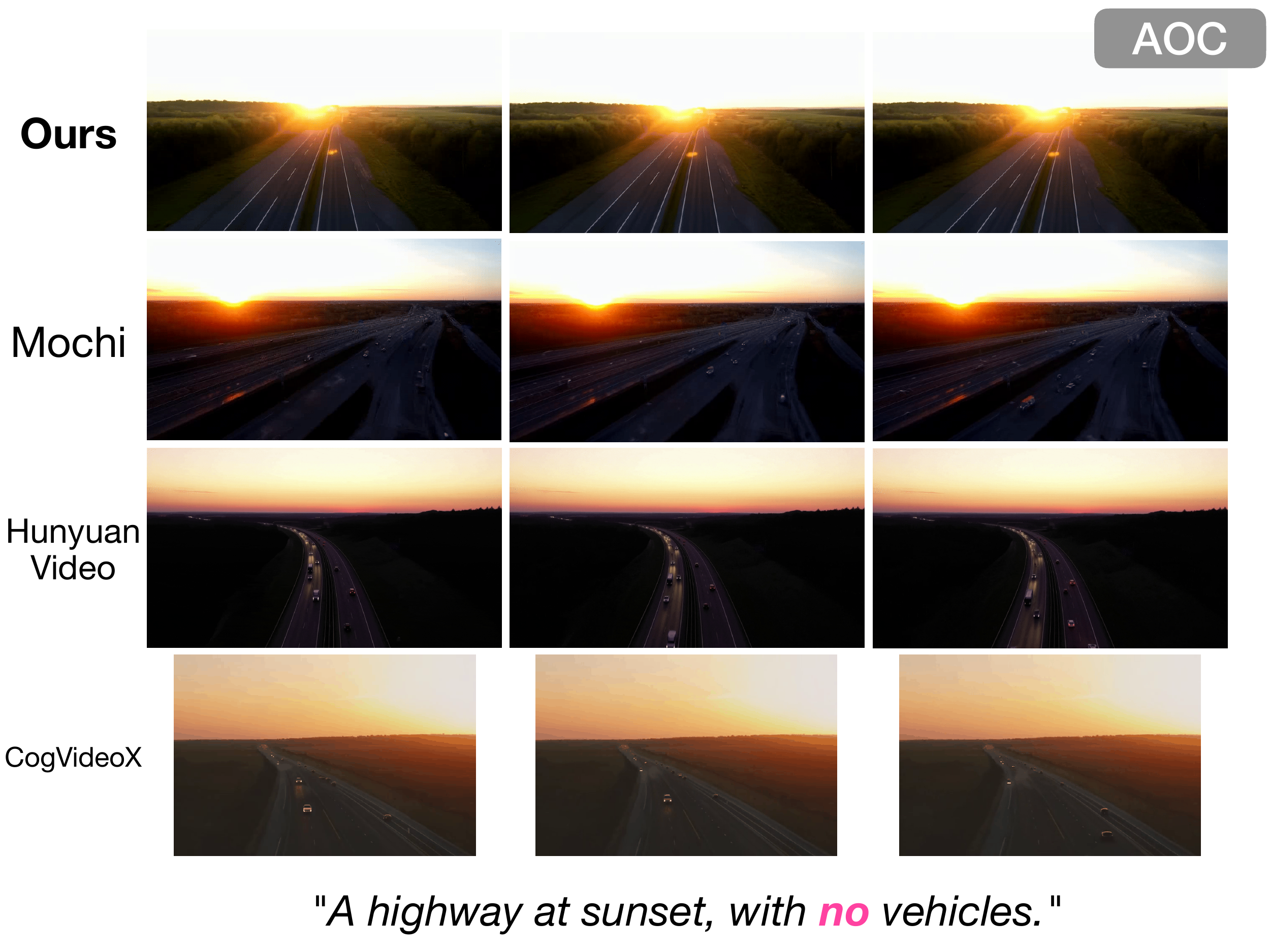}
    \caption{\textbf{Comparison of negation compliance under the AOC (Absence of Object) scenario.} 
    Prompt: ``A highway at sunset, with no vehicles.'' 
    Our model removes the forbidden object while preserving global scene structure and lighting consistency, whereas competing diffusion models introduce unintended vehicles.}
    \label{fig:qual_aoc}
\end{figure*}

\begin{figure*}[t]
    \centering
    \includegraphics[width=\linewidth]
    {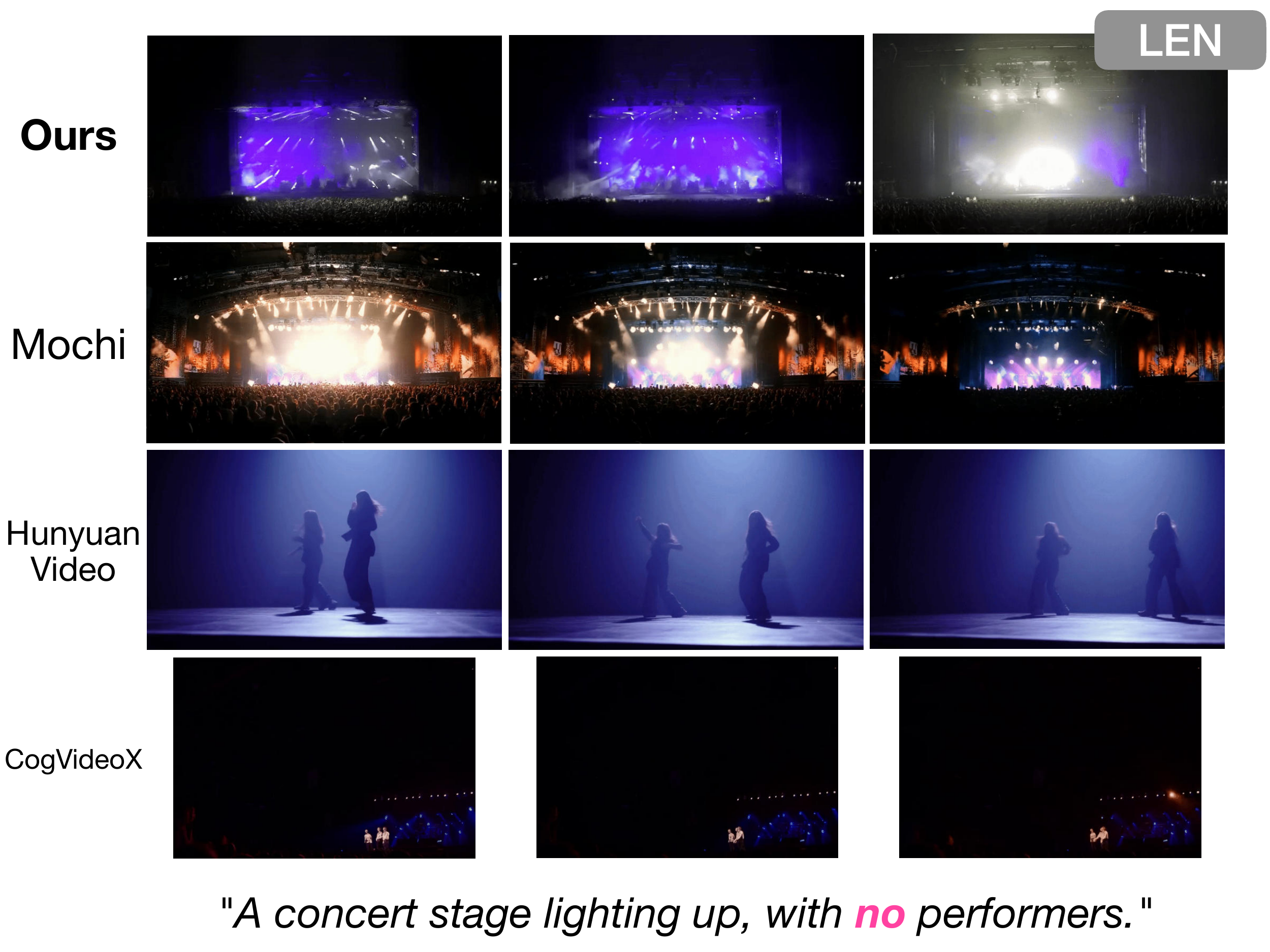}
    \caption{\textbf{Comparison of negation compliance under the LEN (Late Emergence Negation) scenario.} 
    Prompt: ``A concert stage lighting up, with no performers.'' 
    Our model enforces entity-level exclusion while maintaining realistic stage illumination, whereas baselines often hallucinate performers.}
    \label{fig:qual_len}
\end{figure*}

\begin{figure*}[t]
    \centering
    \includegraphics[width=\linewidth]
    {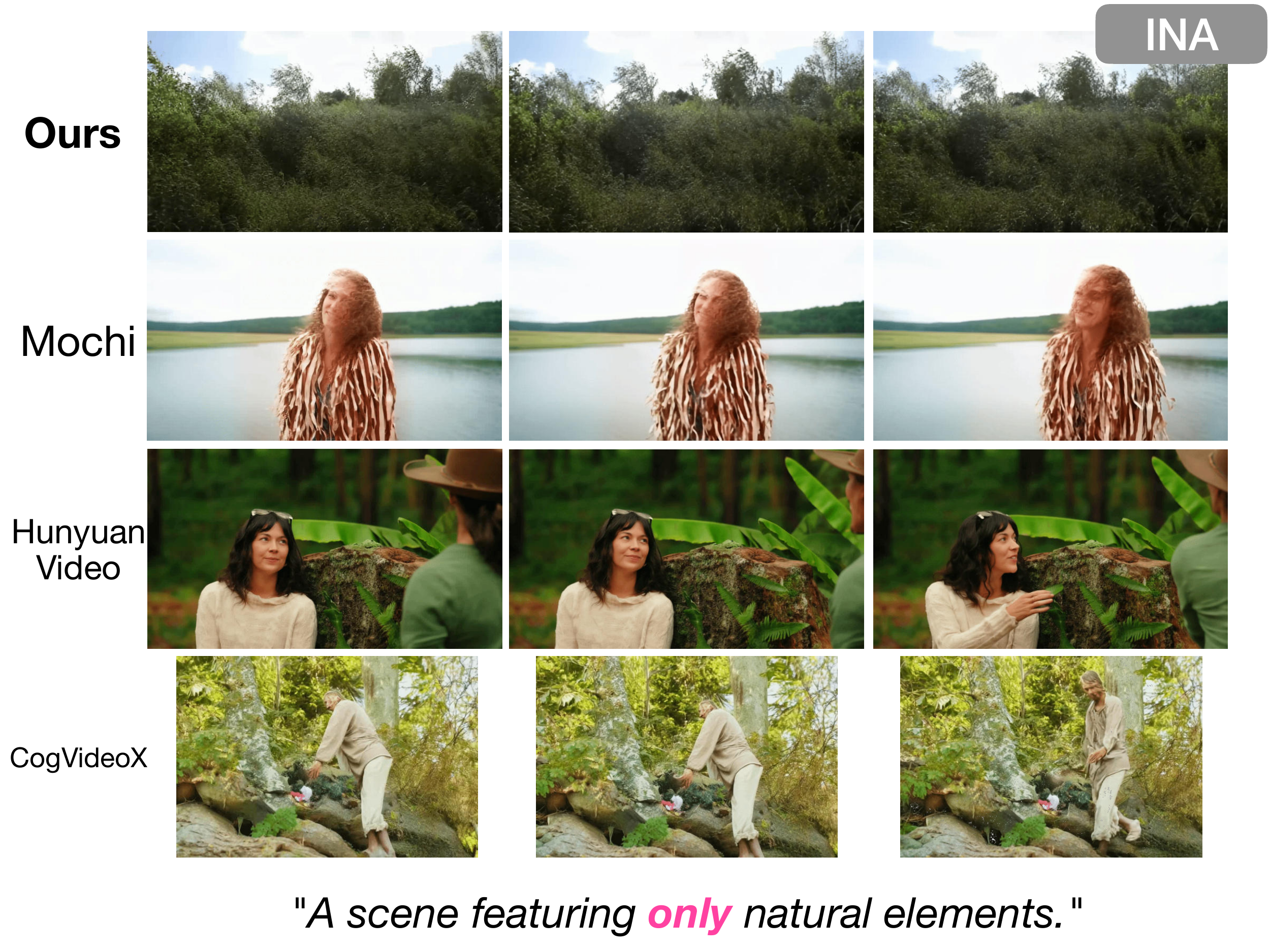}
    \caption{\textbf{Comparison of negation compliance under the INA (Implicit Natural-Only Attribute) scenario.} 
    Prompt: ``A scene featuring only natural elements.'' 
    Our model suppresses human presence while preserving ecological coherence, unlike competing methods that introduce artificial objects.}
    \label{fig:qual_ina}
\end{figure*}

\begin{figure*}[t]
    \centering
    \includegraphics[width=\linewidth]
    {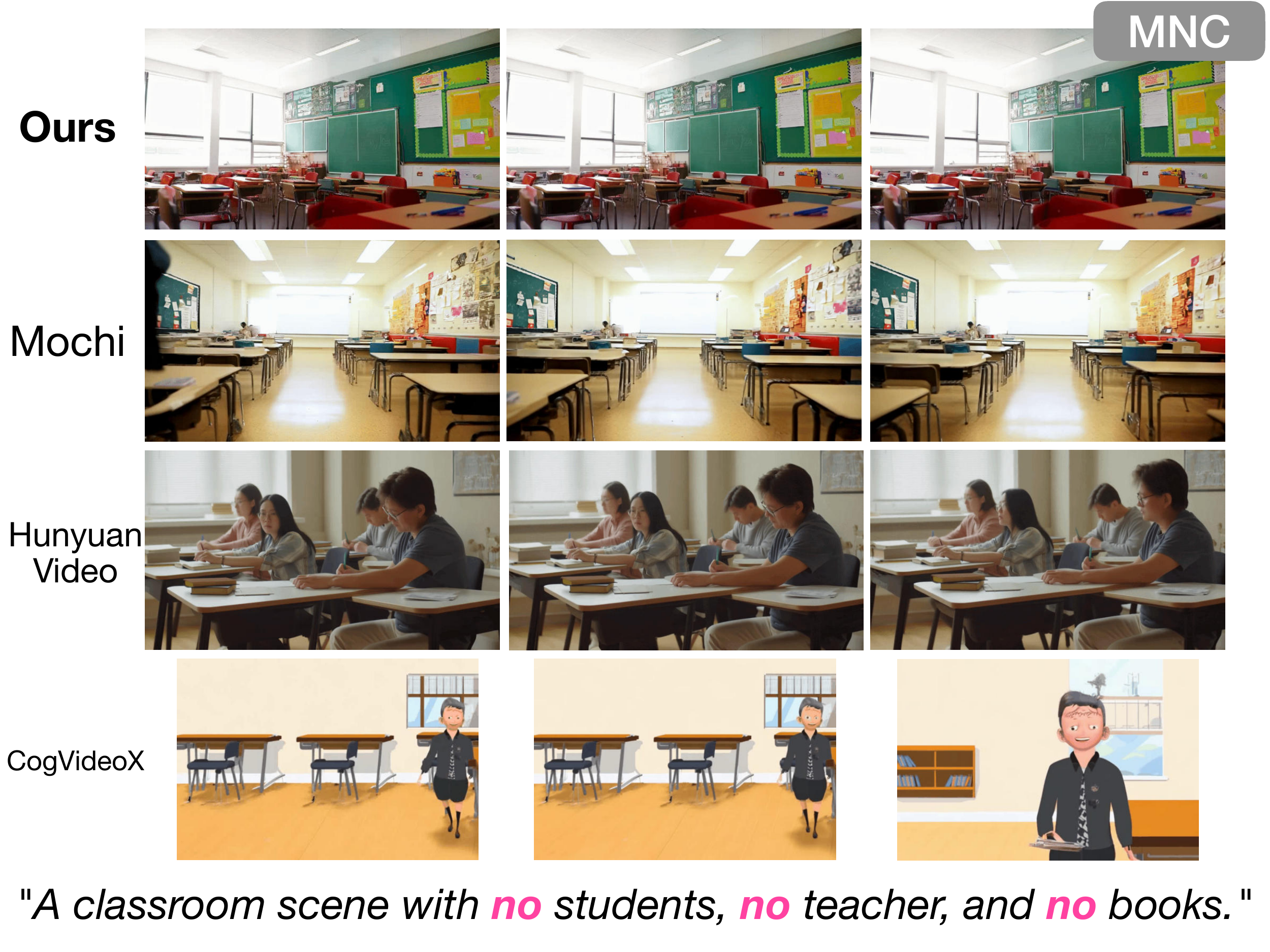}
    \caption{\textbf{Comparison of negation compliance under the MNC (Multi-Negation Composition) scenario.}  Prompt: ``A classroom scene with no students, no teacher, and no books.'' Our model simultaneously enforces multiple exclusion constraints while preserving structural layout, whereas baselines fail to suppress all forbidden entities.}
    \label{fig:qual_mnc}
\end{figure*}

\begin{figure*}[t]
    \centering
    \includegraphics[width=\linewidth]
    {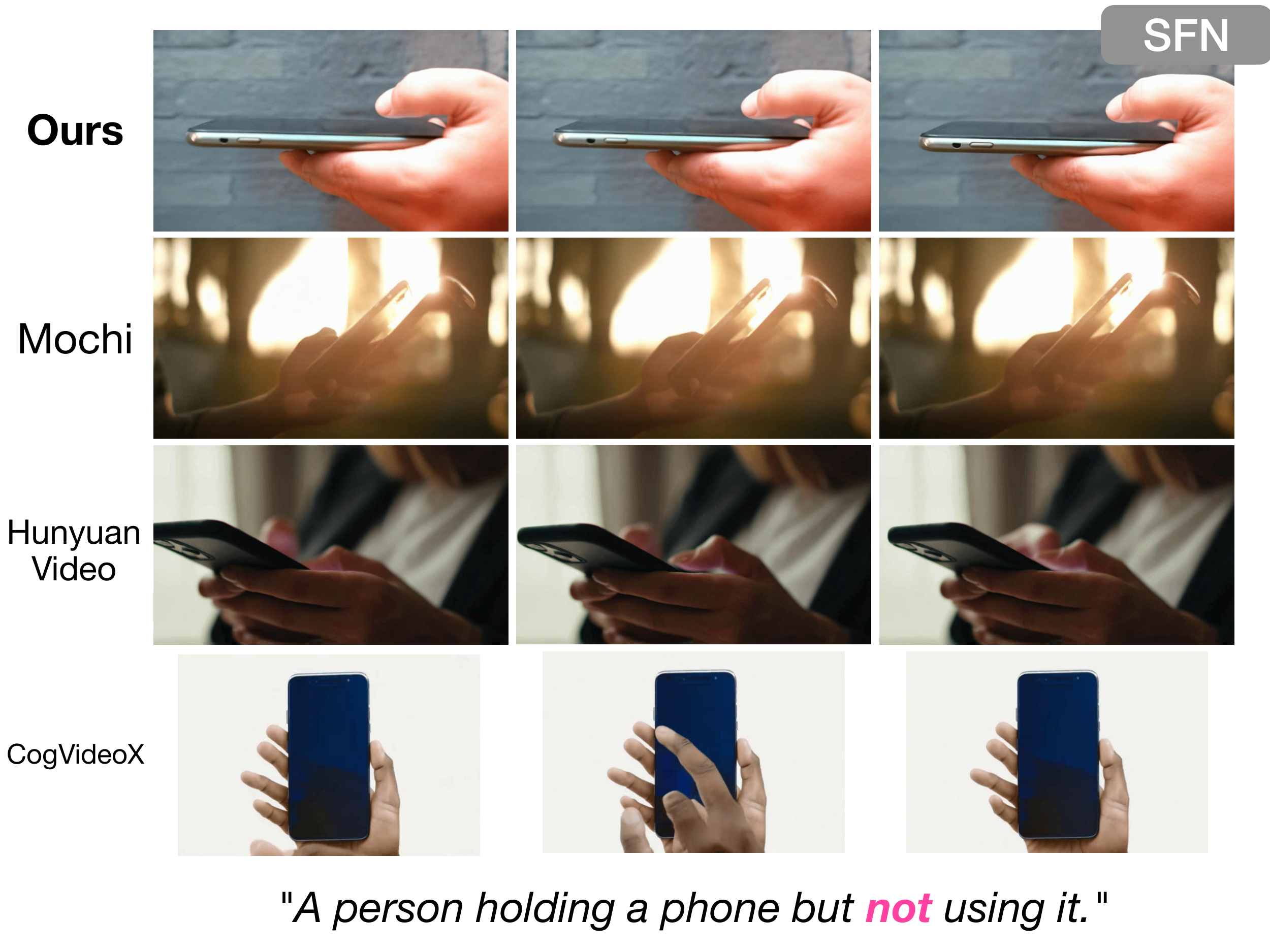}
    \caption{\textbf{Comparison of negation compliance under the SFN (Structural Functional Negation) scenario.} Prompt: ``A person holding a phone but not using it.'' Our model suppresses the functional interaction while preserving object presence, whereas baselines conflate object existence with usage.}
    \label{fig:qual_sfn}
\end{figure*}

\begin{figure*}[t]
    \centering
    \includegraphics[width=\linewidth]
    {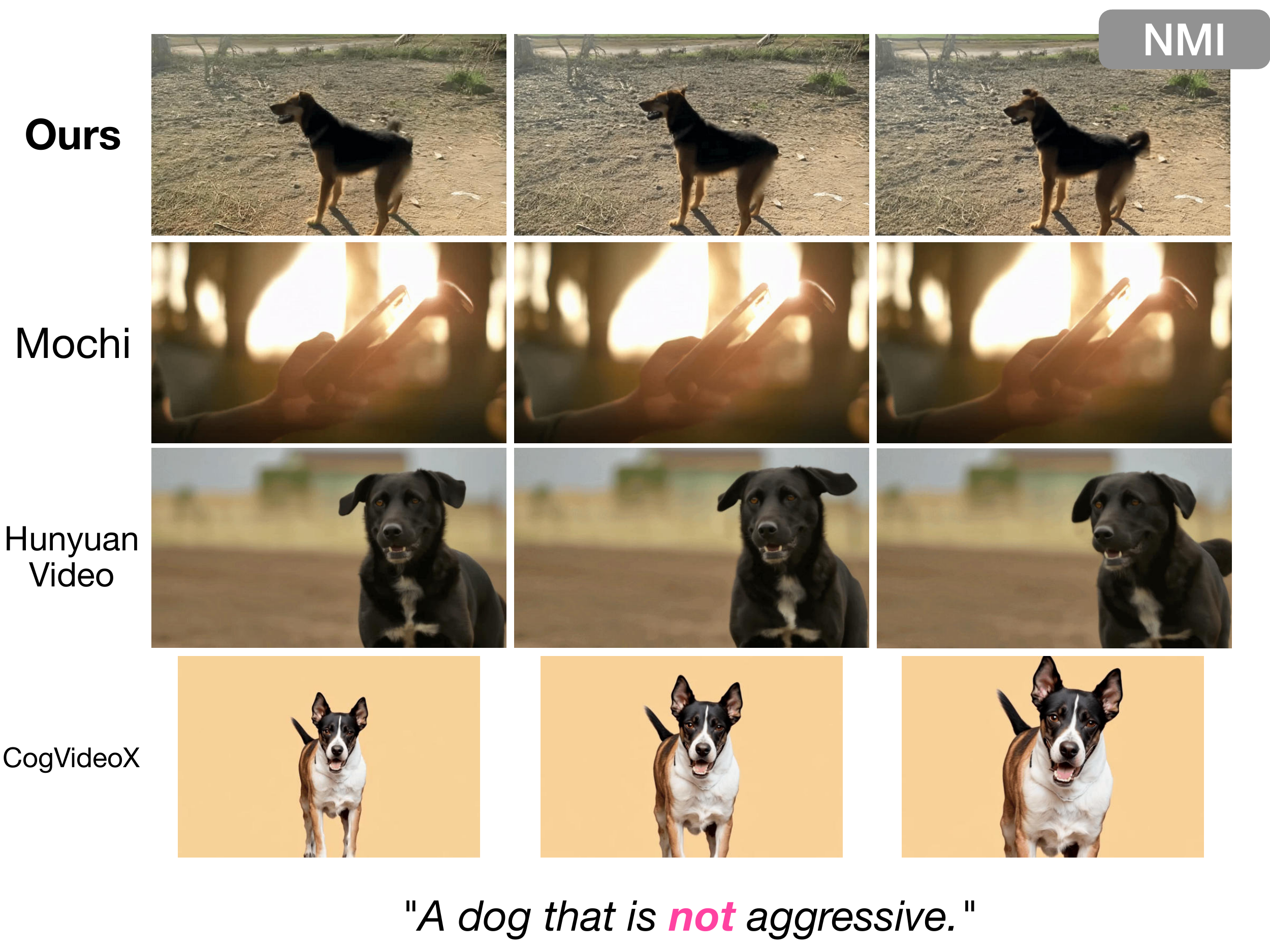}
    \caption{\textbf{Comparison of negation compliance under the NMI (Non-Inversion Mitigation) scenario.} Prompt: ``A dog that is not aggressive.'' Our model modulates behavioral attributes while maintaining subject identity, whereas competing models fail to disentangle appearance from disposition.}
    \label{fig:qual_nmi}
\end{figure*}

\begin{figure*}[t]
    \centering
    \includegraphics[width=\linewidth]
    {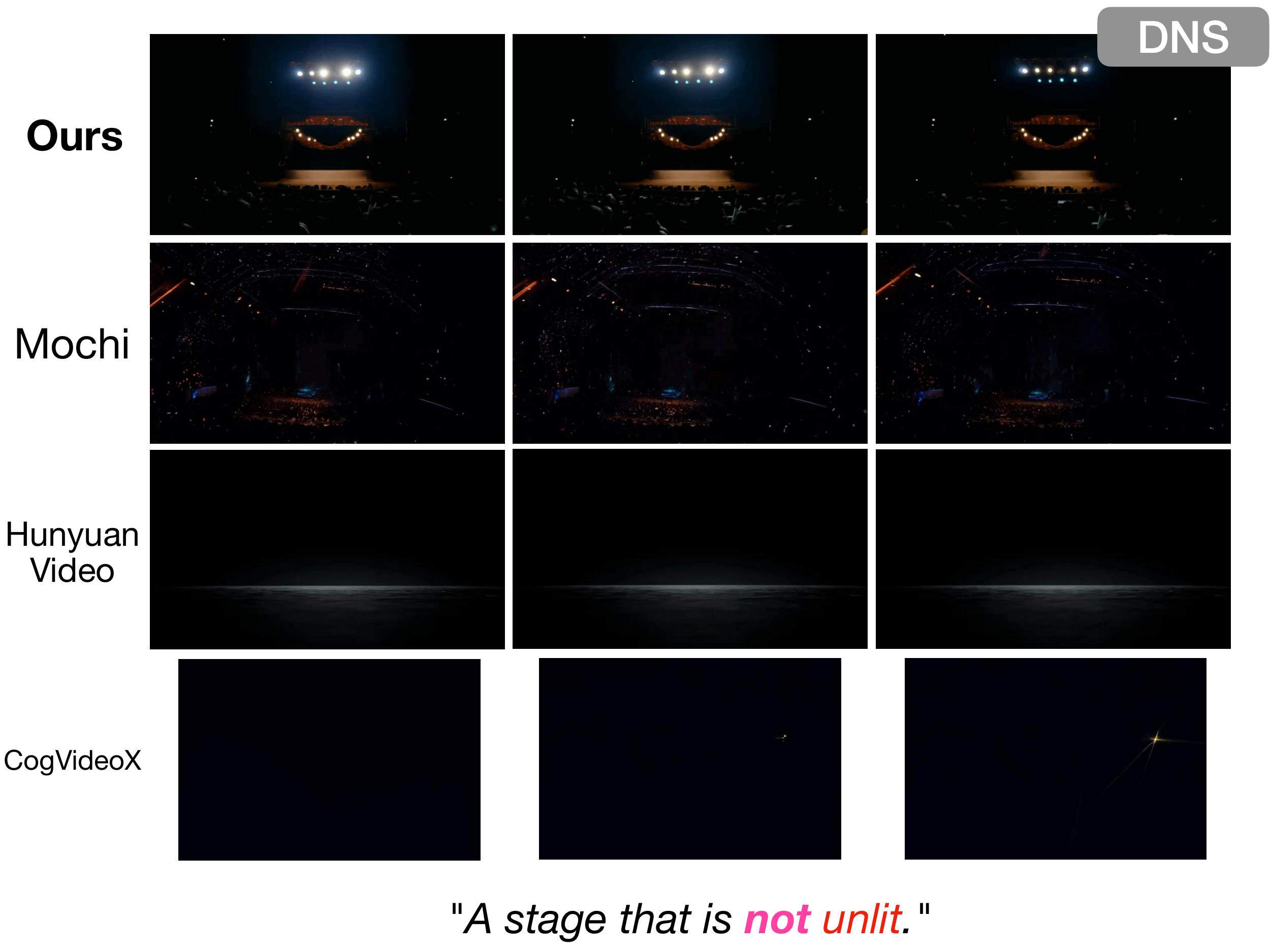}
    \caption{\textbf{Comparison of negation compliance under the DNS (Double Negation Sensitivity) scenario.} Prompt: ``A stage that is not unlit.'' Our model correctly resolves double negation semantics, producing illuminated scenes, while baselines collapse into dark or ambiguous outputs.}
    \label{fig:qual_dns}
\end{figure*}

\begin{figure*}[t]
    \centering
    \includegraphics[width=\linewidth]
    {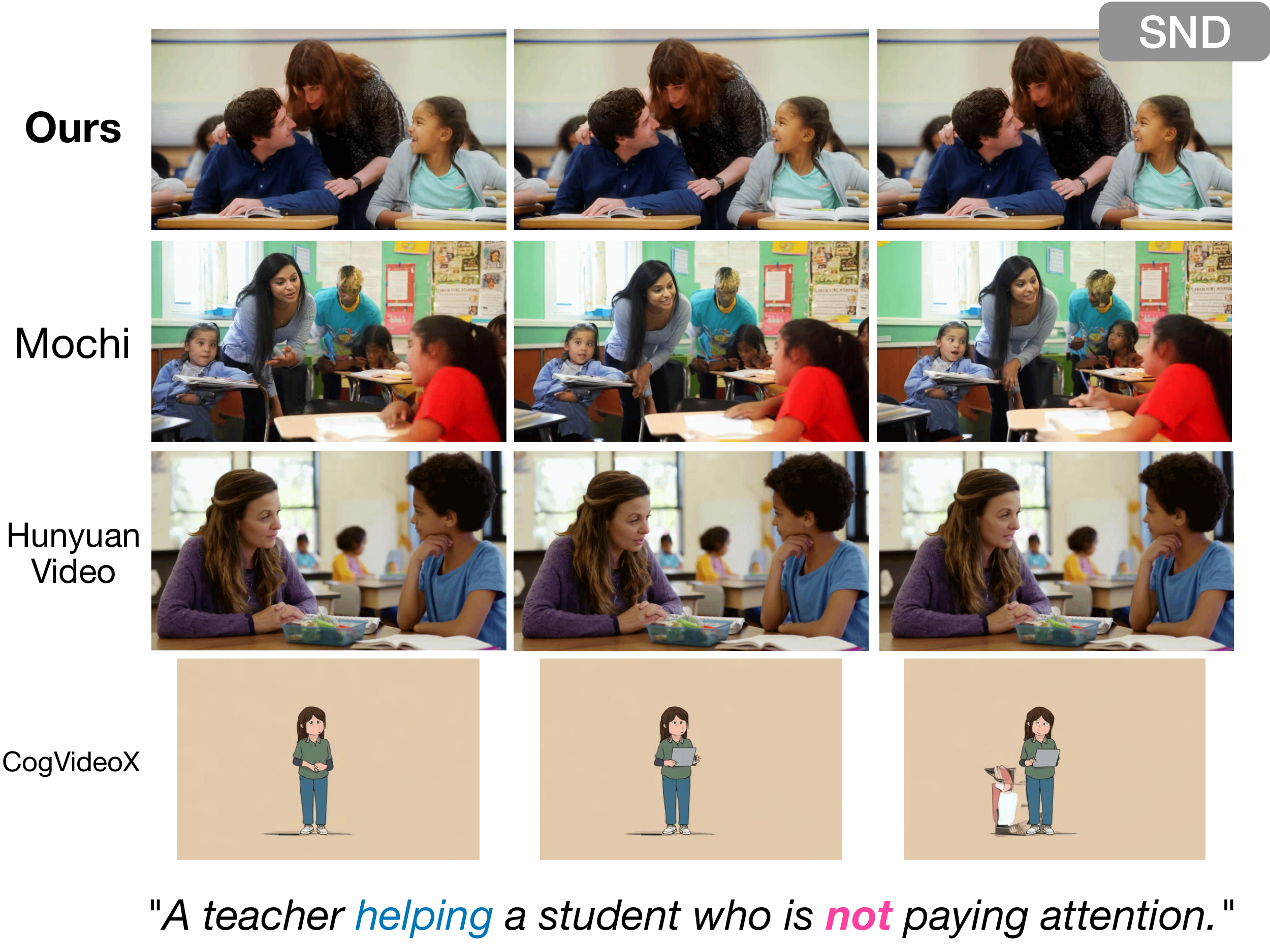}
    \caption{\textbf{Comparison of negation compliance under the SND (Scoped Negation Disambiguation) scenario.} Prompt: ``A teacher helping a student who is not paying attention.'' Our model correctly localizes the negation scope to the student’s behavior, whereas competing methods misinterpret the scope or alter unrelated elements.}
    \label{fig:qual_snd}
\end{figure*}

\begin{figure*}[t]
\centering
\includegraphics[width=\linewidth]{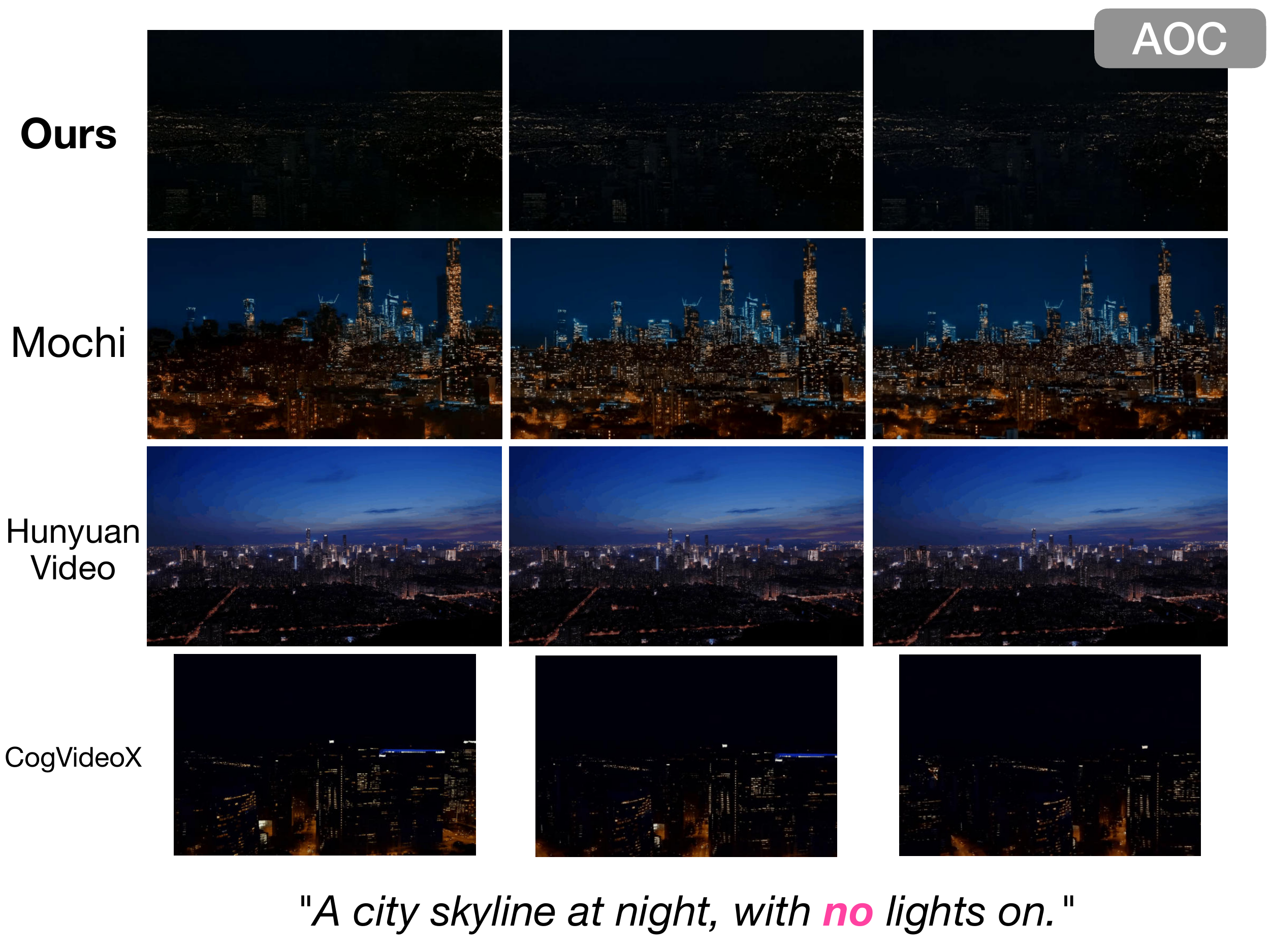}
\caption{\textbf{Qualitative comparison under the AOC (Absence of Object) scenario.} Prompt: ``A city skyline at night, with no lights on.'' Our model correctly suppresses artificial light sources across the skyline while preserving realistic nighttime structure. In contrast, Mochi and HunyuanVideo illuminate buildings despite the negation constraint, while CogVideoX produces unstable or partially lit scenes.}
\label{fig:supp_aoc_1}
\end{figure*}

\begin{figure*}[t]
\centering
\includegraphics[width=\linewidth]{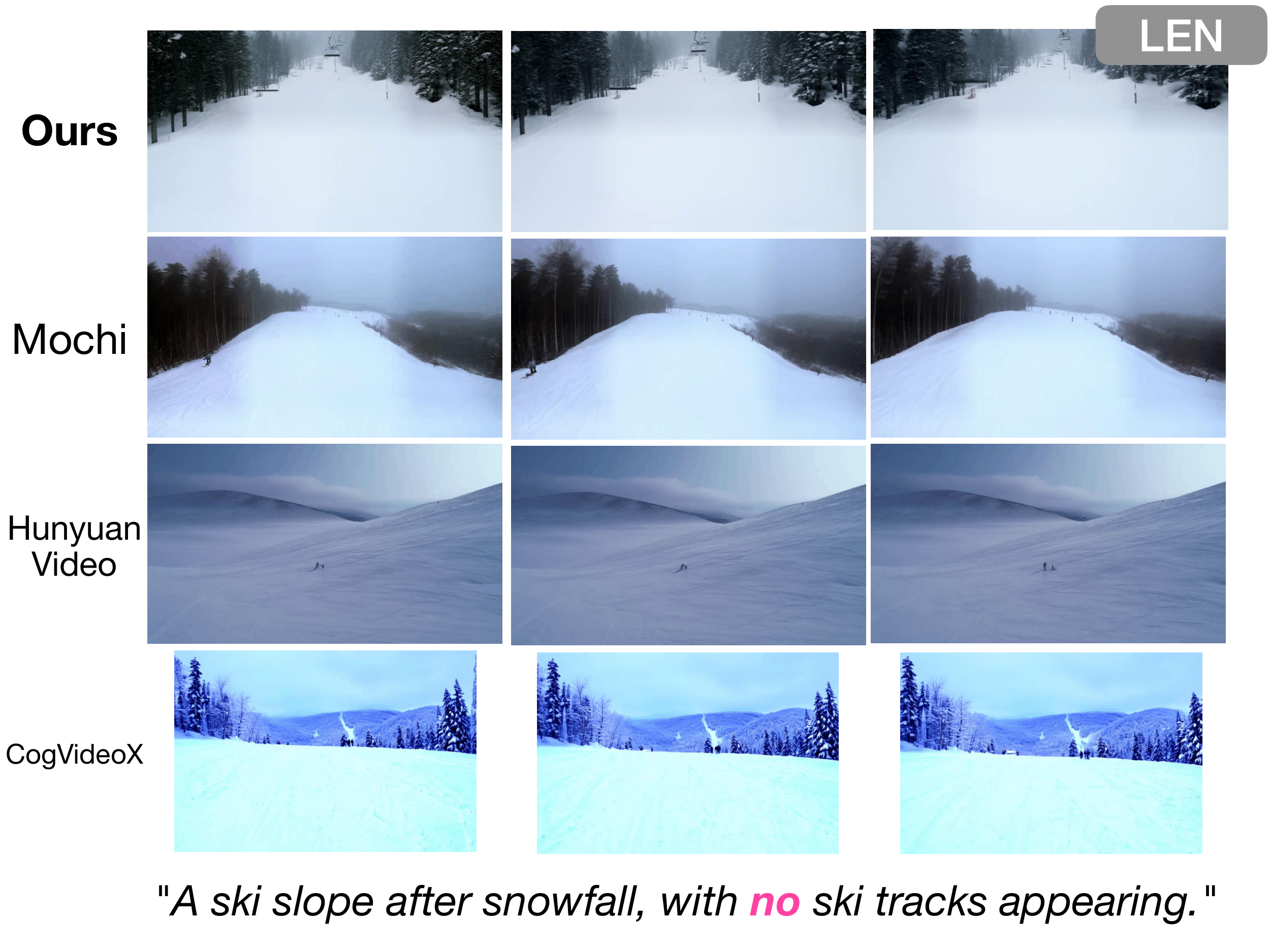}
\caption{\textbf{Qualitative comparison under the LEN (Logical Exclusion of Entities) scenario.} Prompt: ``A ski slope after snowfall, with no ski tracks appearing.'' Our model maintains a pristine snow surface consistent with the negation constraint. Competing models frequently introduce ski tracks or surface disturbances, indicating failure to enforce logical exclusion during generation.}
\label{fig:supp_len_1}
\end{figure*}

\begin{figure*}[t]
\centering
\includegraphics[width=\linewidth]{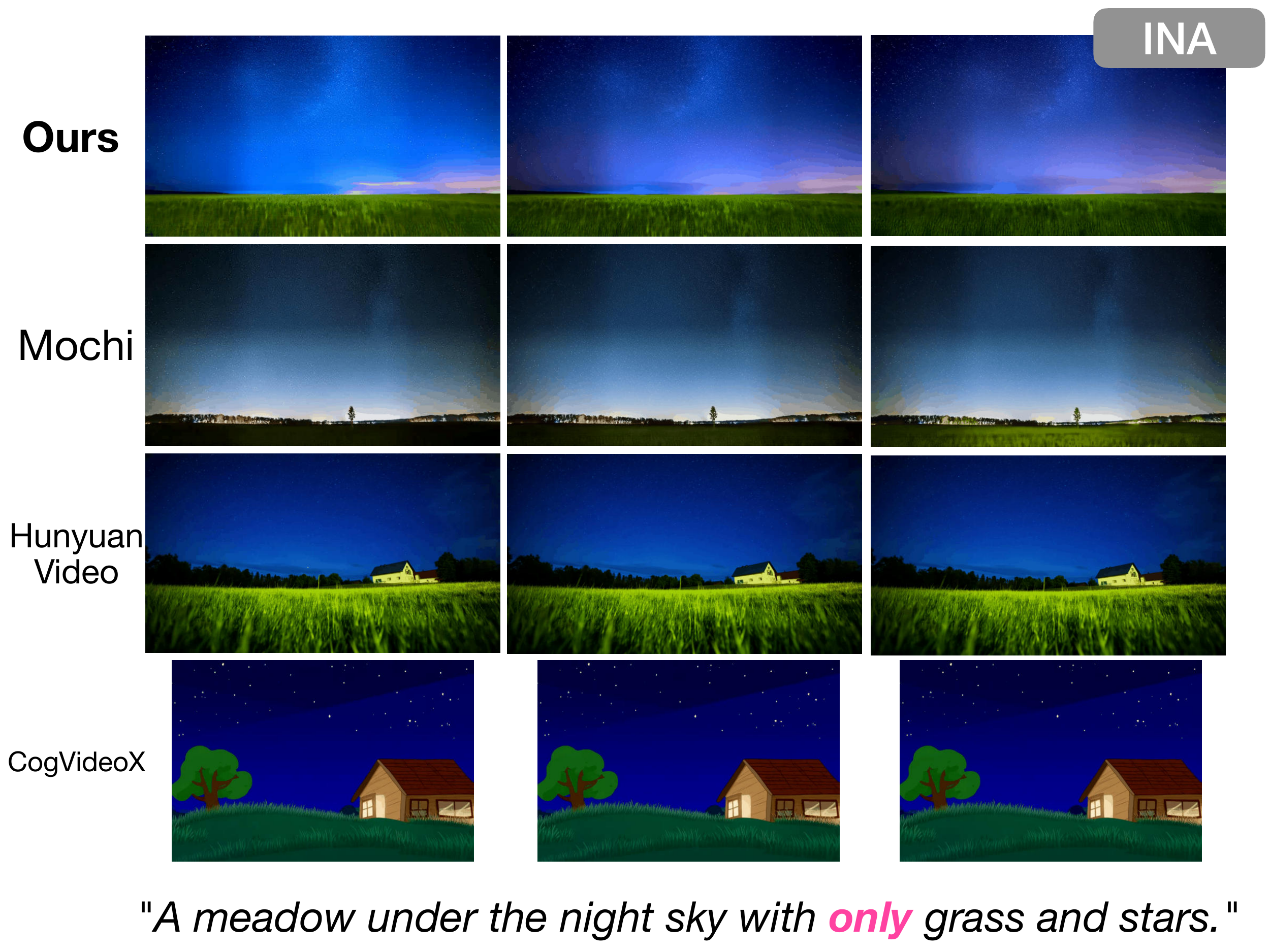}
\caption{\textbf{Qualitative comparison under the INA (Implicit Natural-Only Attribute) scenario.} Prompt: ``A meadow under the night sky with only grass and stars.'' Our model preserves a purely natural scene without introducing artificial structures. Baselines frequently insert buildings or synthetic elements, demonstrating difficulty enforcing implicit natural-only constraints.}
\label{fig:supp_ina_1}
\end{figure*}

\begin{figure*}[t]
\centering
\includegraphics[width=\linewidth]{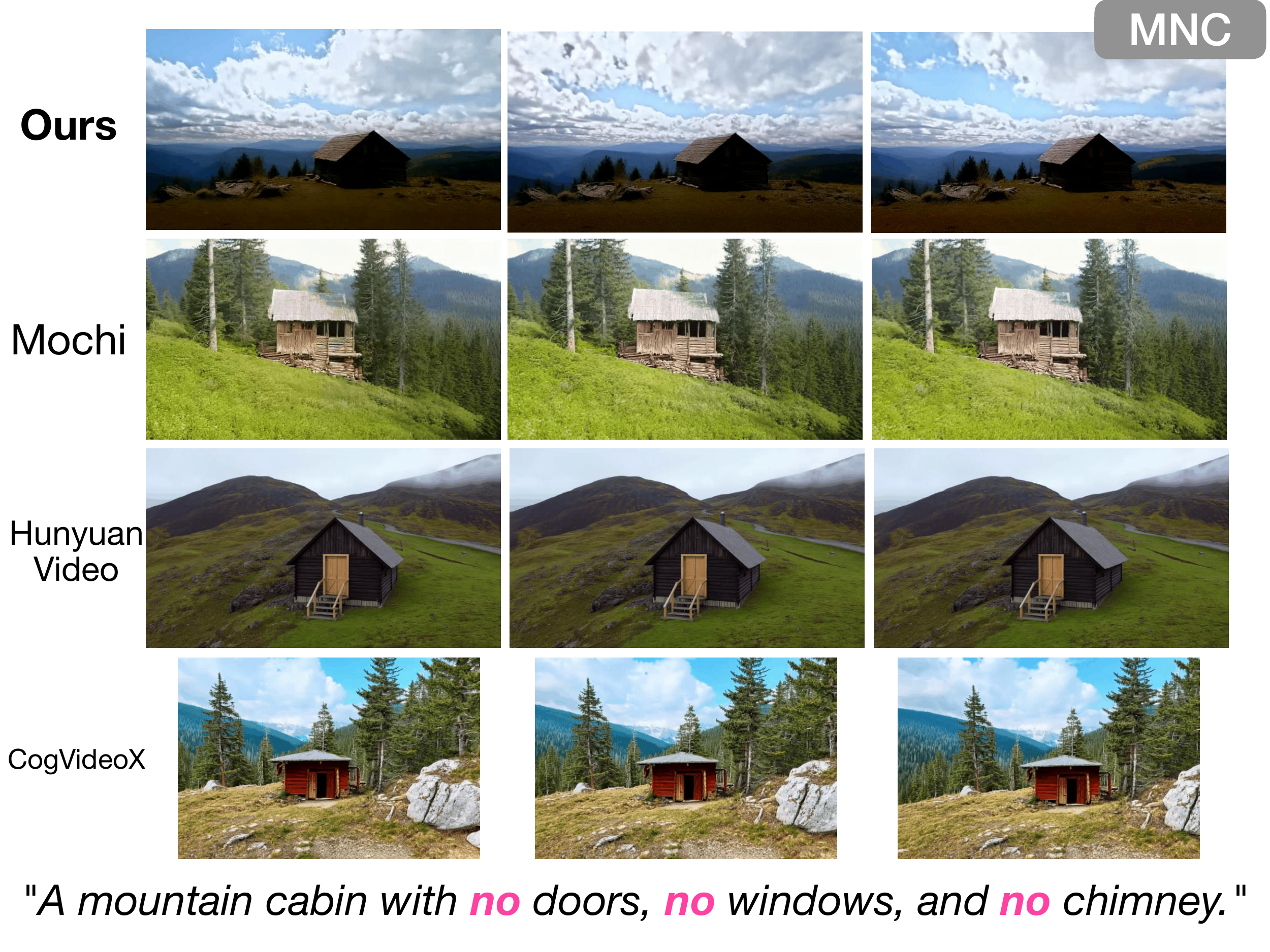}
\caption{\textbf{Qualitative comparison under the MNC (Multi-Negation Composition) scenario.} Prompt: ``A mountain cabin with no doors, no windows, and no chimney.'' Our model simultaneously enforces multiple negation constraints while preserving global structure. Competing models fail to remove all prohibited components, often leaving doors or windows visible.}
\label{fig:supp_mnc_1}
\end{figure*}

\begin{figure*}[t]
\centering
\includegraphics[width=\linewidth]{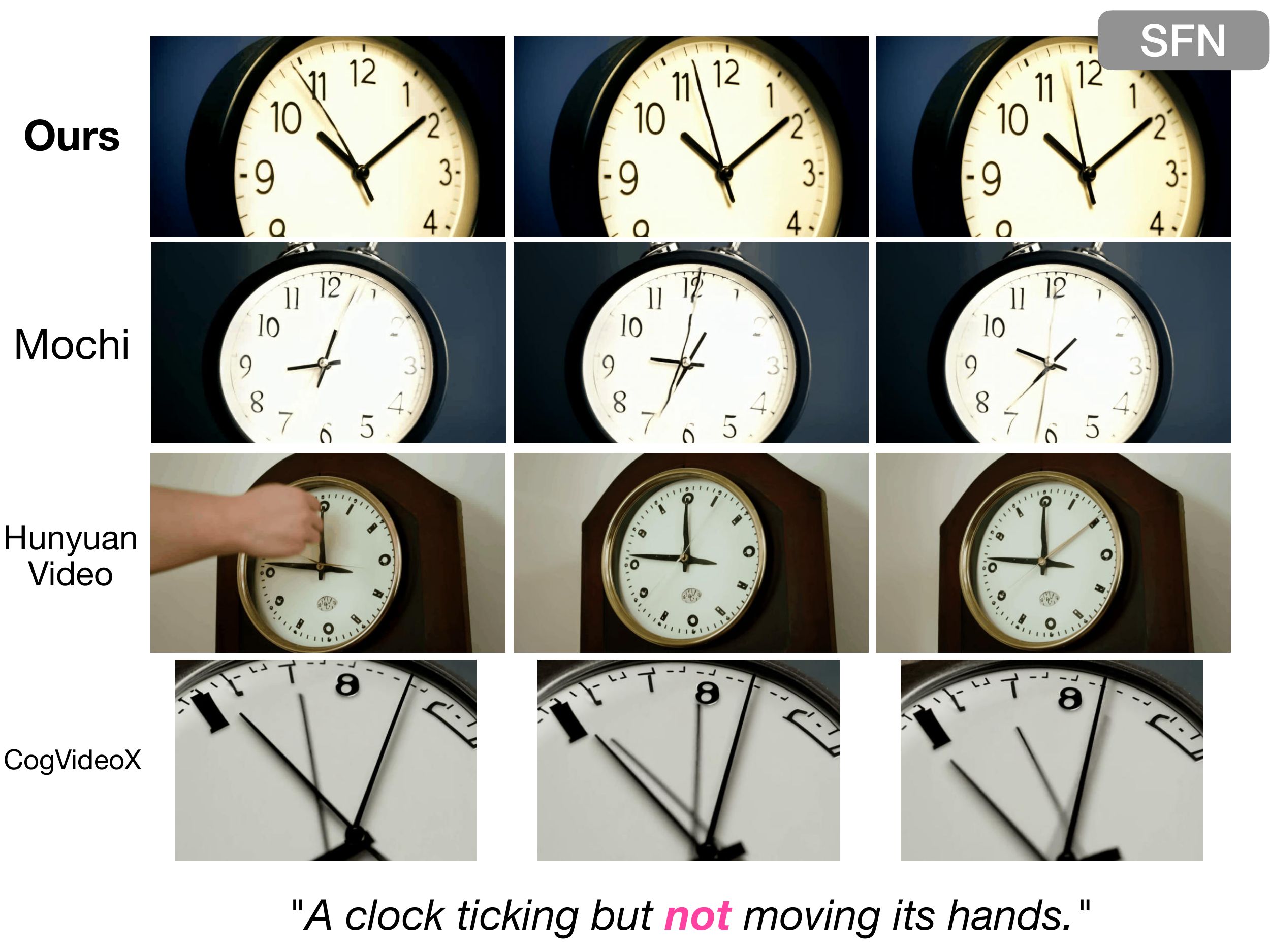}
\caption{\textbf{Qualitative comparison under the SFN (Structural Functional Negation) scenario.} Prompt: ``A clock ticking but not moving its hands.'' Our model captures the semantic distinction between object activity and functional motion. Notably, HunyuanVideo misinterprets the constraint and shows moving a person's hands, while other baselines fail to maintain consistent negation behavior.}
\label{fig:supp_sfn_1}
\end{figure*}

\begin{figure*}[t]
\centering
\includegraphics[width=\linewidth]{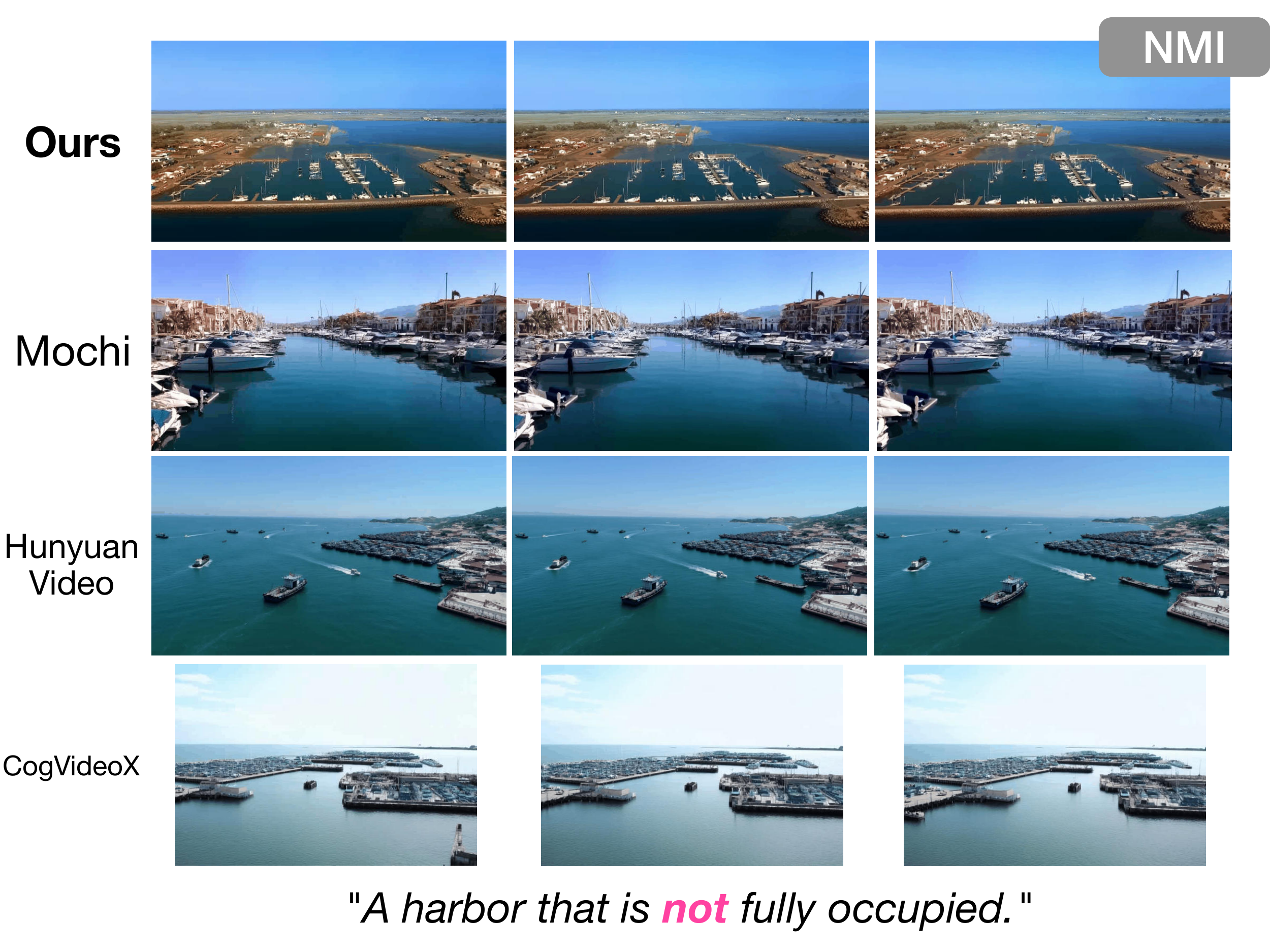}
\caption{\textbf{Qualitative comparison under the NMI (Negation with Moderated Intensity) scenario.} Prompt: ``A harbor that is not fully occupied.''  Our model correctly represents partial occupancy while respecting the negation condition. Baseline methods frequently produce fully crowded scenes, indicating poor control over graded negation semantics.}
\label{fig:supp_nmi_1}
\end{figure*}

\begin{figure*}[t]
\centering
\includegraphics[width=\linewidth]{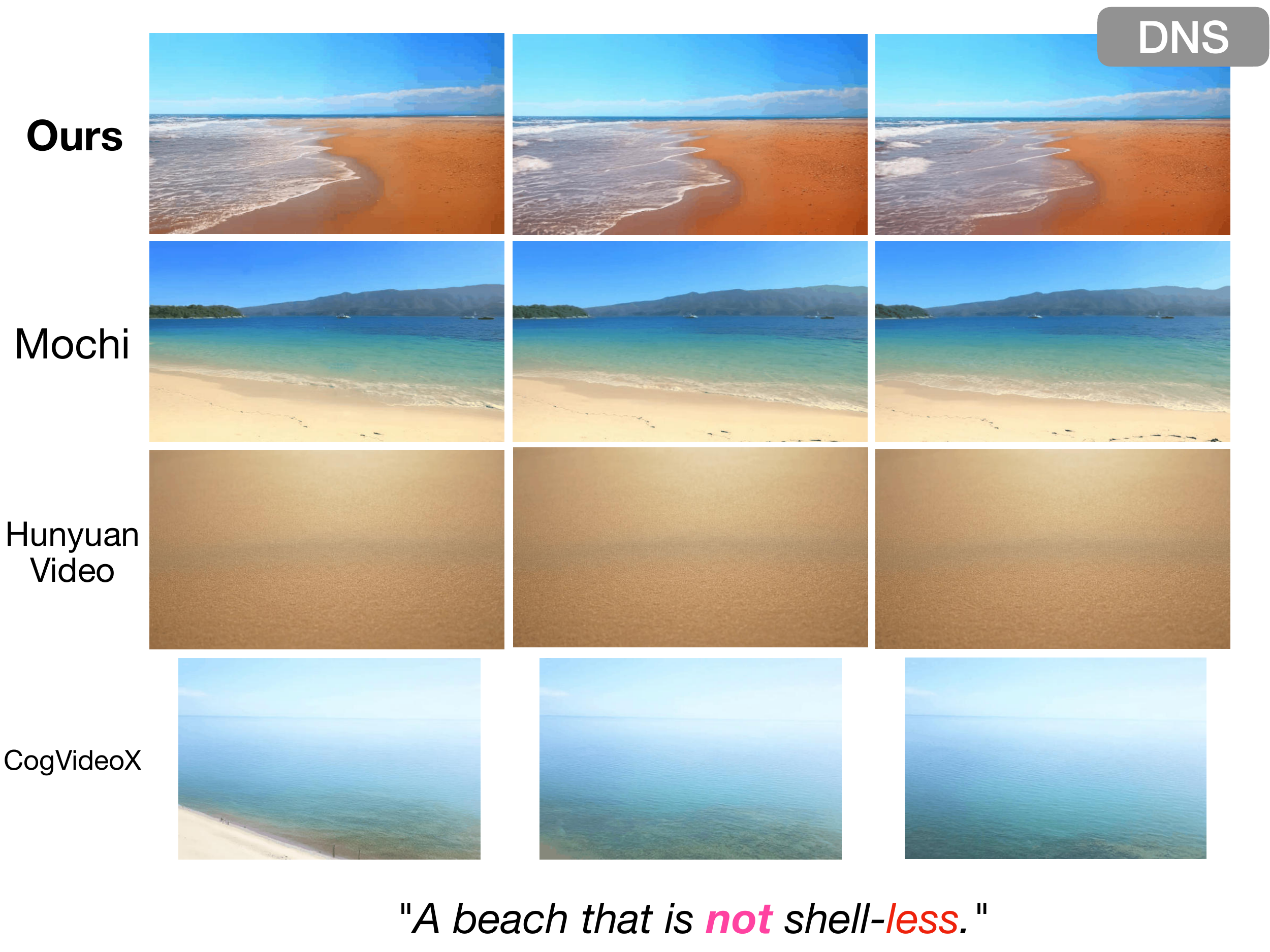}
\caption{\textbf{Qualitative comparison under the DNS (Double Negation Sensitivity) scenario.} Prompt: ``A beach that is not shell-less.'' Our model correctly resolves the double-negation constraint and generates beaches containing shells. Baselines often collapse to trivial empty scenes or fail to capture the intended semantic inversion.}
\label{fig:supp_dns_1}
\end{figure*}

\begin{figure*}[t]
\centering
\includegraphics[width=\linewidth]{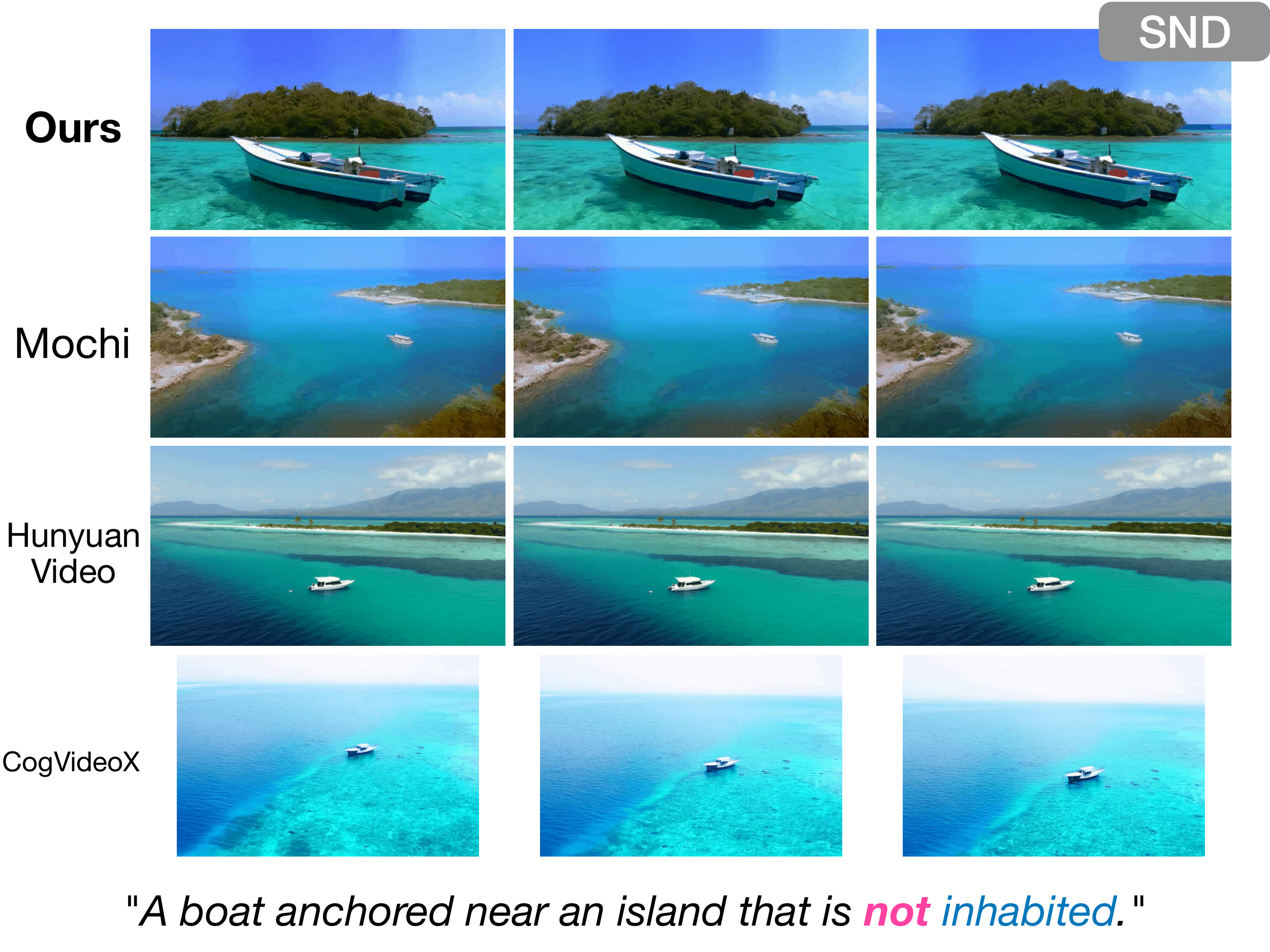}
\caption{\textbf{Qualitative comparison under the SND (Scoped Negation Disambiguation) scenario.} Prompt: ``A boat anchored near an island that is not inhabited.'' Our model correctly applies the negation constraint to the island while preserving the boat. Competing models frequently misinterpret the scope, altering the boat or generating populated islands.}
\label{fig:supp_snd_1}
\end{figure*}

\begin{figure*}[t]
\centering
\includegraphics[width=\linewidth]{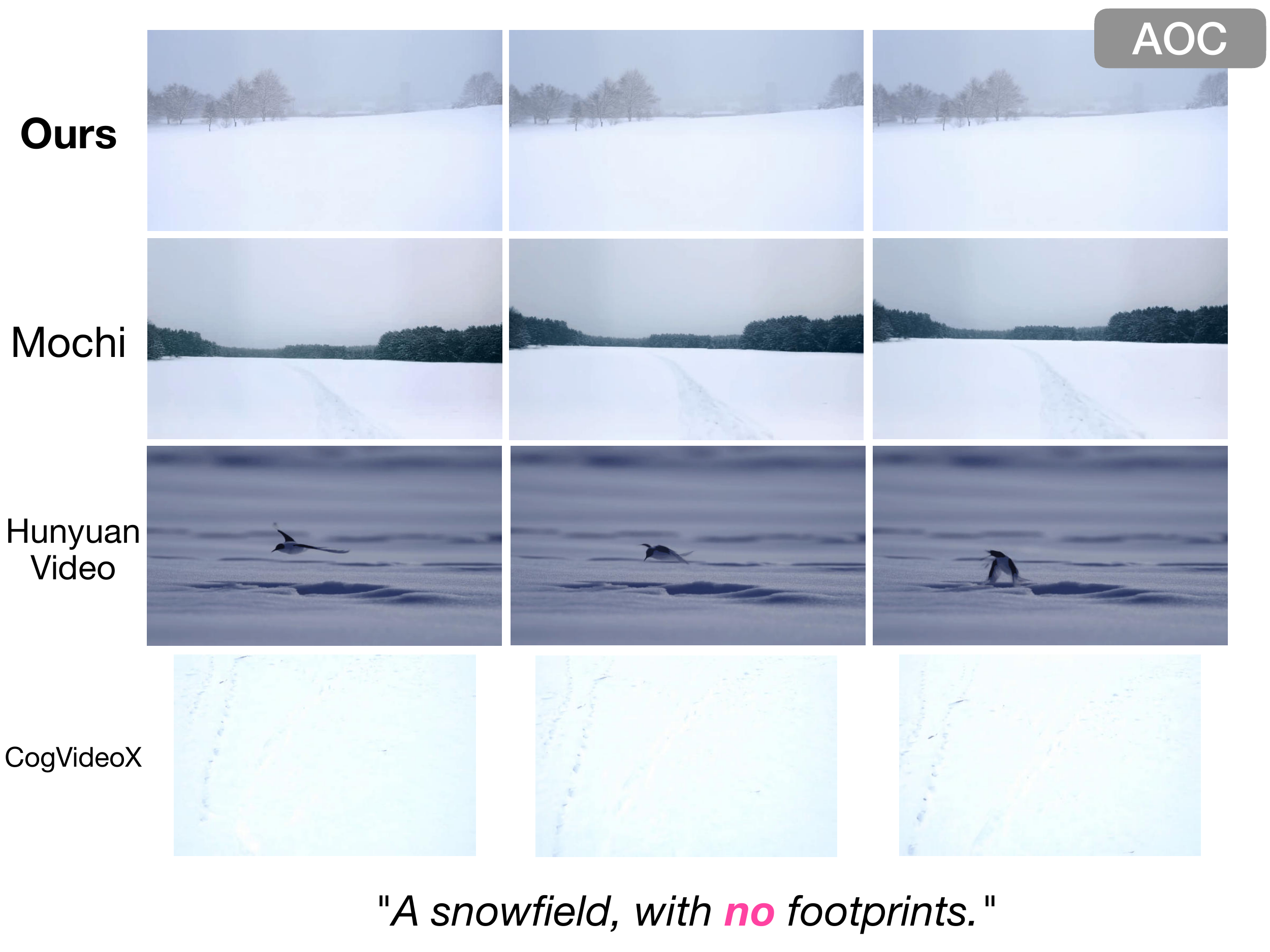}
\caption{\textbf{Qualitative comparison under the AOC scenario.} Prompt: ``A snowfield, with no footprints.''  Our model generates a clean untouched snowfield consistent with the negation constraint. Baseline methods frequently introduce footprints or surface artifacts.}
\label{fig:supp_aoc_2}
\end{figure*}

\begin{figure*}[t]
\centering
\includegraphics[width=\linewidth]{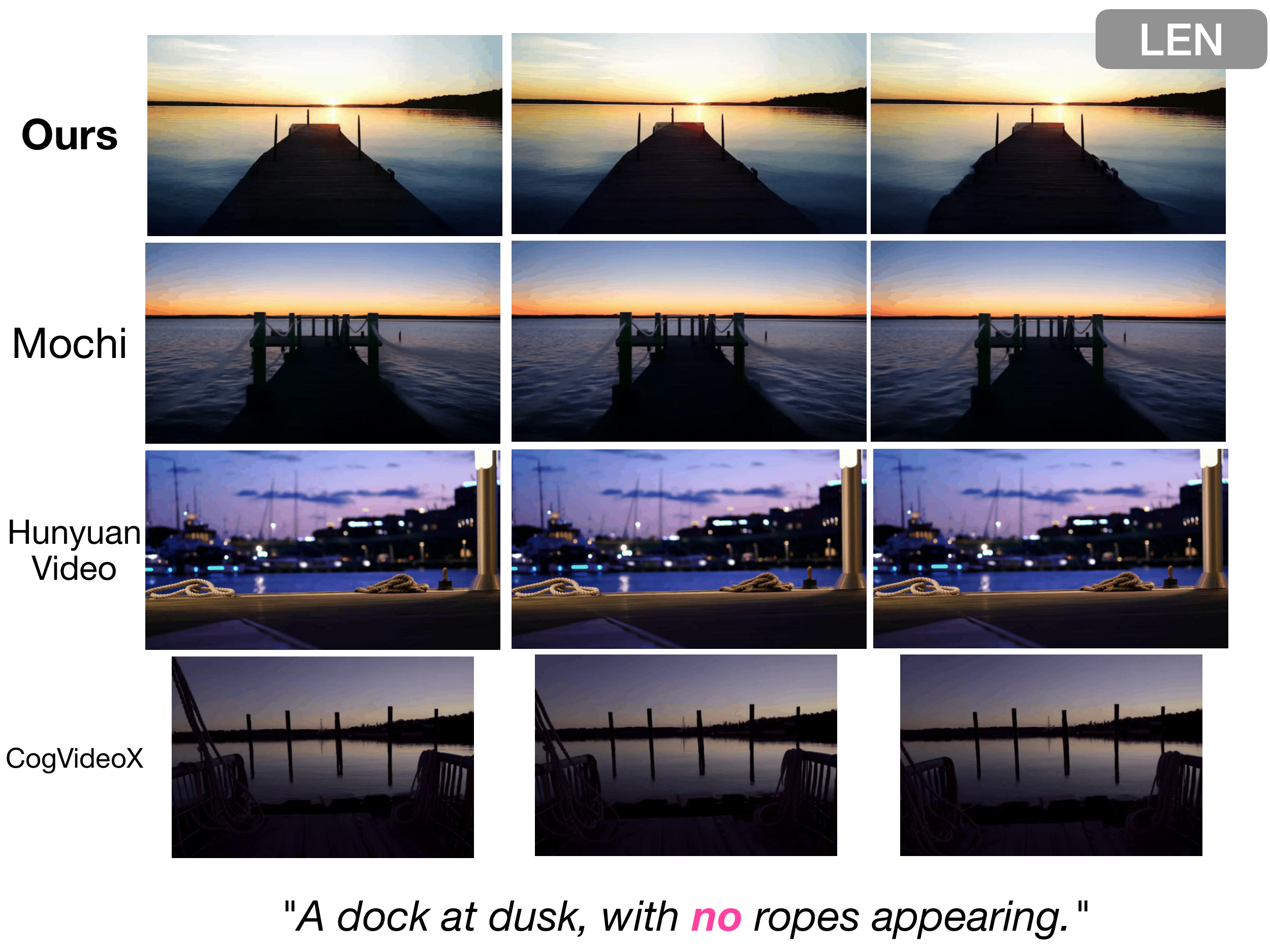}
\caption{\textbf{Qualitative comparison under the LEN scenario.} Prompt: ``A dock at dusk, with no ropes appearing.'' Our model correctly suppresses rope structures while maintaining scene realism. Baselines often introduce ropes or structural elements violating the negation constraint.}
\label{fig:supp_len_2}
\end{figure*}

\begin{figure*}[t]
\centering
\includegraphics[width=\linewidth]{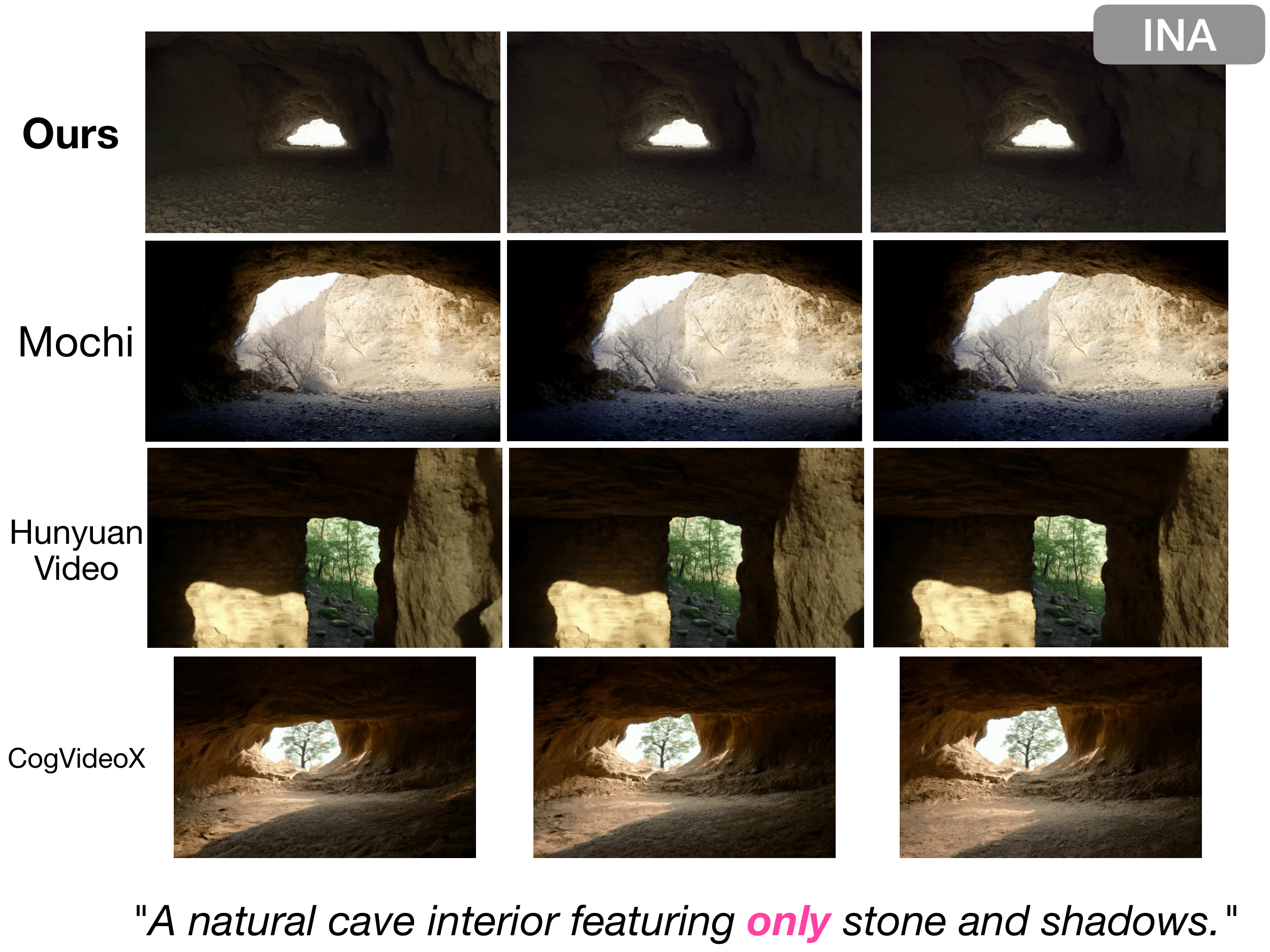}
\caption{\textbf{Qualitative comparison under the INA scenario.} Prompt: ``A natural cave interior featuring only stone and shadows.'' Our model maintains a purely geological environment, while competing models frequently introduce vegetation or artificial lighting elements inconsistent with the prompt.}
\label{fig:supp_ina_2}
\end{figure*}

\begin{figure*}[t]
\centering
\includegraphics[width=\linewidth]{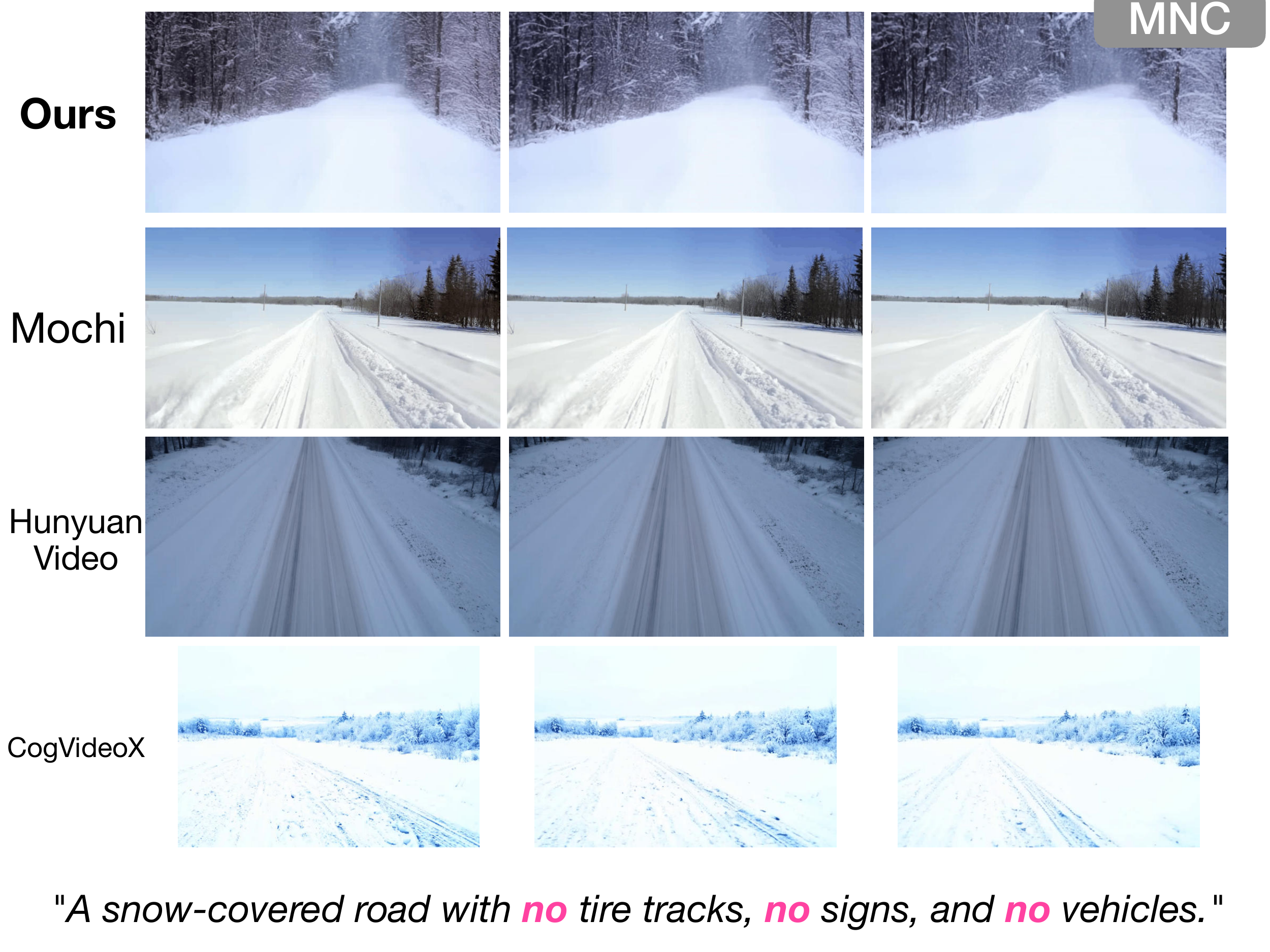}
\caption{\textbf{Qualitative comparison under the MNC scenario.} Prompt: ``A snow-covered road with no tire tracks, no signs, and no vehicles.'' Our model enforces multiple negation constraints simultaneously, producing a clean road. Baseline models commonly violate at least one constraint.}
\label{fig:supp_mnc_2}
\end{figure*}

\begin{figure*}[t]
\centering
\includegraphics[width=\linewidth]{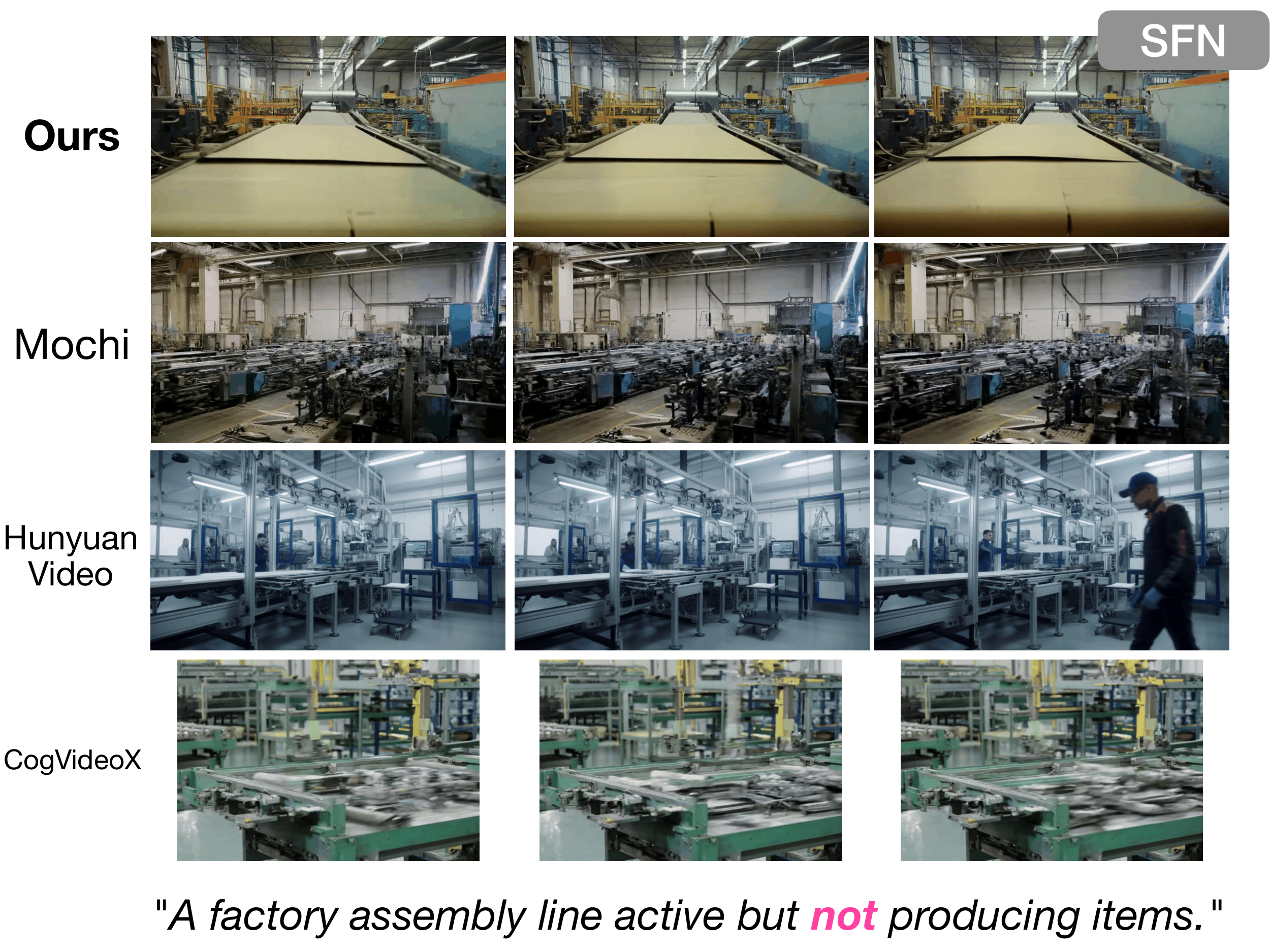}
\caption{\textbf{Qualitative comparison under the SFN scenario.} Prompt: ``A factory assembly line active but not producing items.'' Our model separates mechanical motion from product output, preserving activity without generated items. Baselines frequently produce items or misinterpret the constraint entirely.}
\label{fig:supp_sfn_2}
\end{figure*}

\begin{figure*}[t]
\centering
\includegraphics[width=\linewidth]{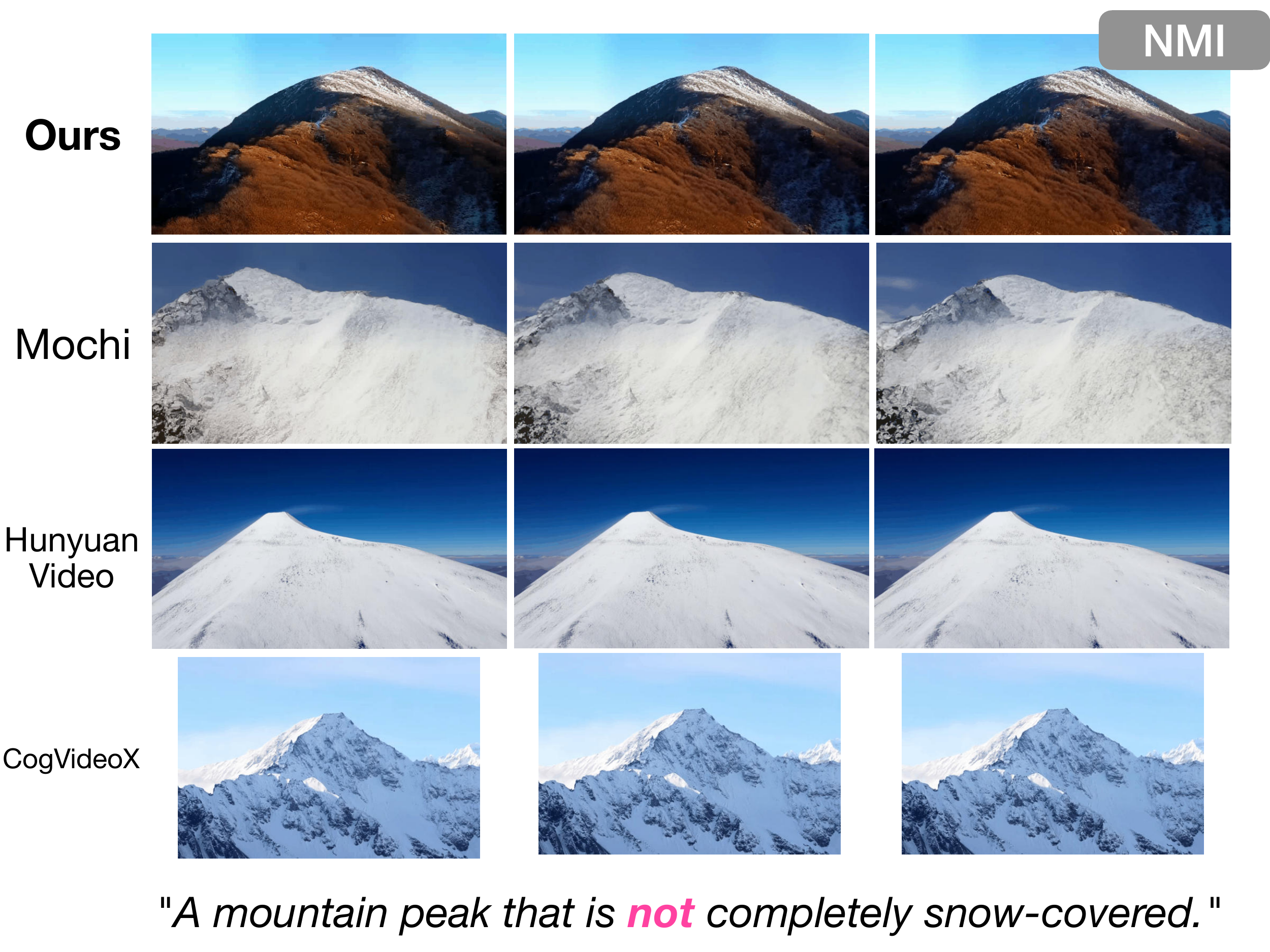}
\caption{\textbf{Qualitative comparison under the NMI scenario.} Prompt: ``A mountain peak that is not completely snow-covered.'' Our model correctly produces partial snow coverage. Baselines often generate fully snow-covered peaks, ignoring the moderated negation constraint.}
\label{fig:supp_nmi_2}
\end{figure*}

\begin{figure*}[t]
\centering
\includegraphics[width=\linewidth]{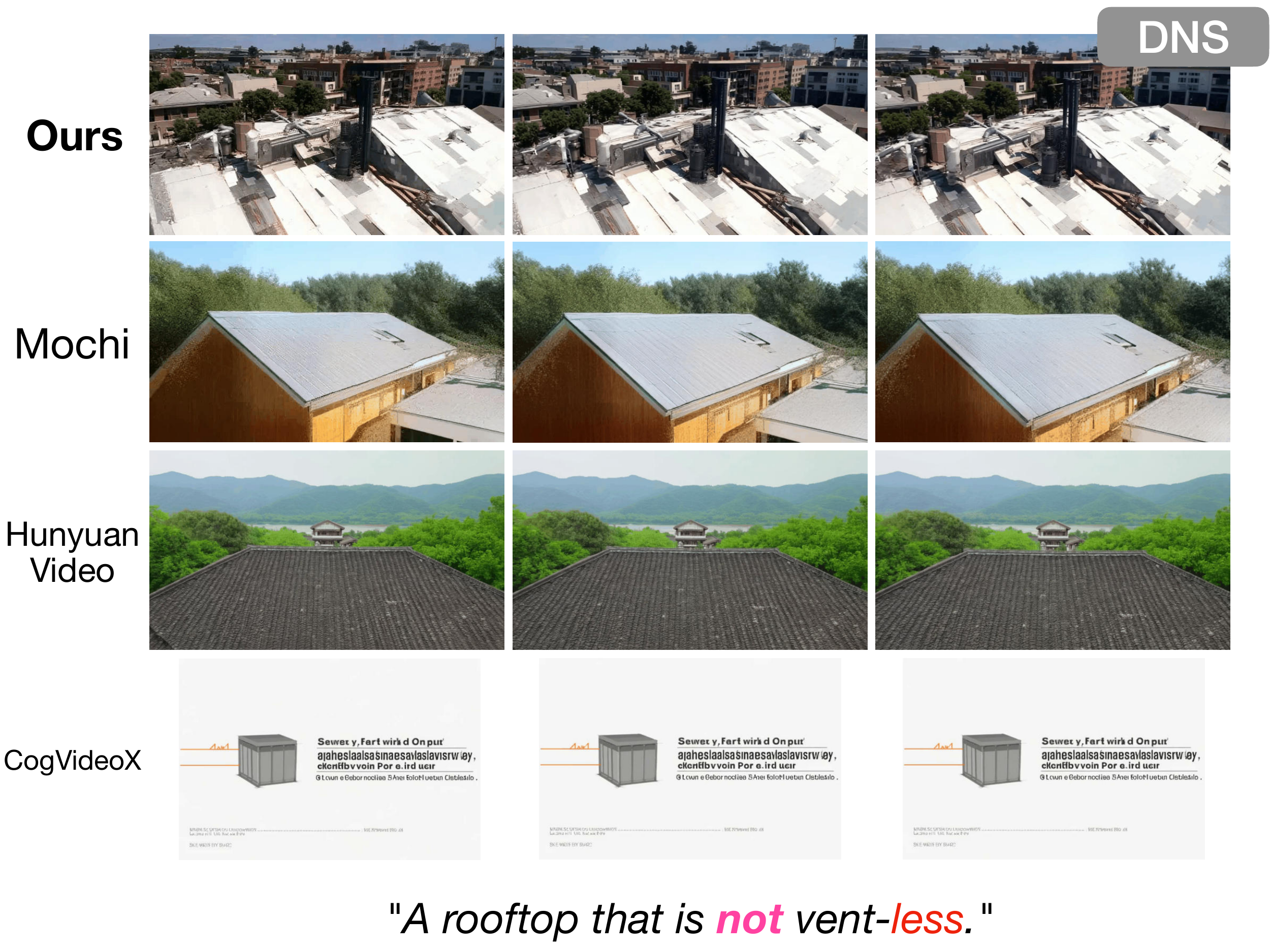}
\caption{\textbf{Qualitative comparison under the DNS scenario.} Prompt: ``A rooftop that is not vent-less.'' Our model correctly resolves double negation and generates rooftops with vents. Baselines frequently ignore the constraint and produce vent-less rooftops.}
\label{fig:supp_dns_2}
\end{figure*}

\begin{figure*}[t]
\centering
\includegraphics[width=\linewidth]{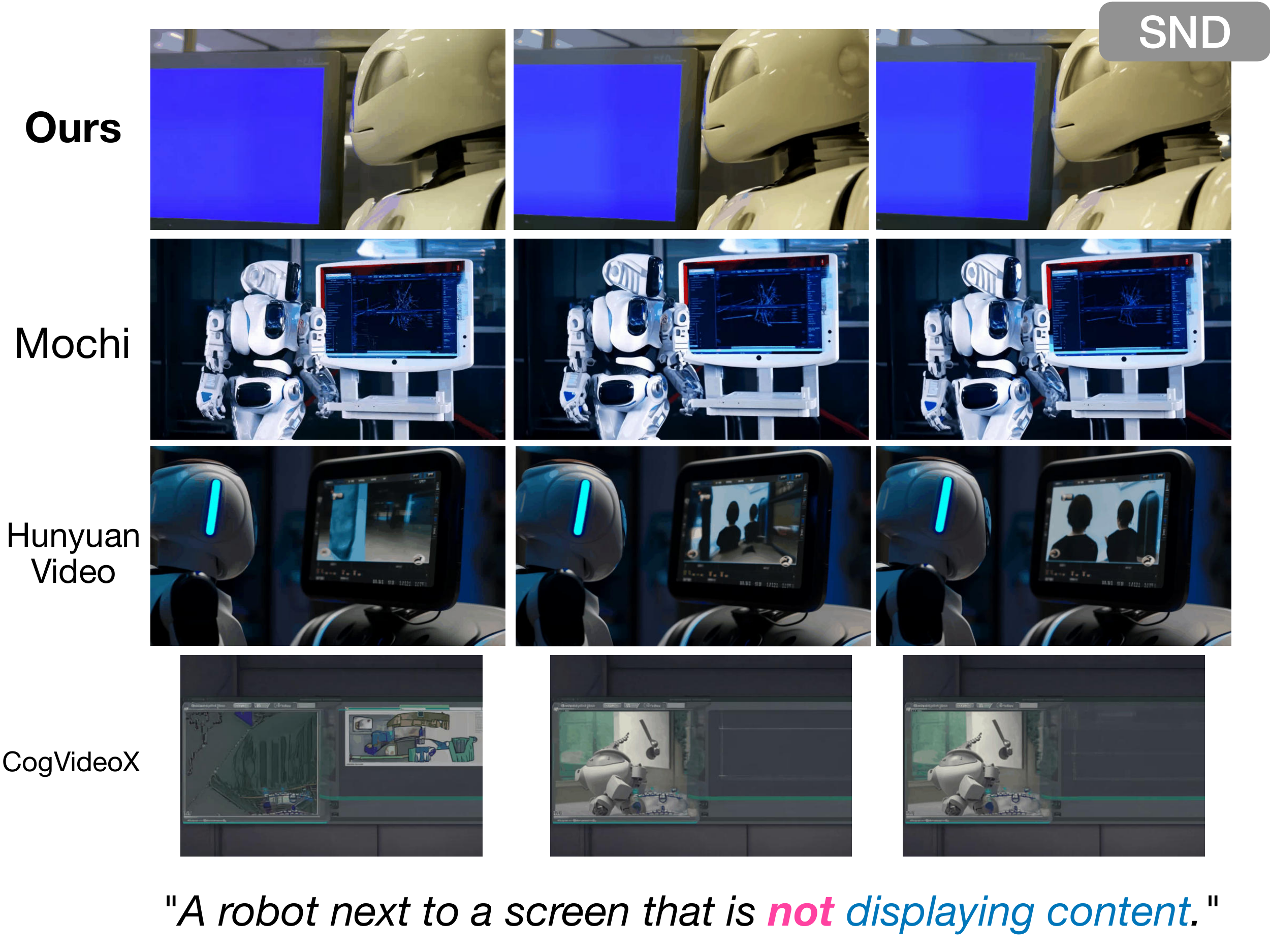}
\caption{\textbf{Qualitative comparison under the SND scenario.} Prompt: ``A robot next to a screen that is not displaying content.'' Our model correctly applies the negation constraint to the screen while maintaining the robot and surrounding environment. Baseline methods often display content on the screen or misapply the negation to unrelated objects.}
\label{fig:supp_snd_2}
\end{figure*}

\textbf{Figures~\ref{fig:detailed_qualitative}, \ref{fig:detailed_qualitative_supp1}, and \ref{fig:detailed_qualitative_supp2}} (Detailed close-up results are shown in Figures~\ref{fig:qual_aoc}--\ref{fig:supp_snd_2}) presents qualitative comparisons across eight negation categories, spanning simple object absence (AOC, MNC), logical exclusion (LEN), implicit constraints (INA), functional negation (SFN), moderated semantic negation (NMI), double negation (DNS), and scoped relational negation (SND).

In straightforward absence settings such as AOC (``no vehicles'') and MNC (``no students, no teacher, no books''), baseline models often partially retain forbidden objects or introduce subtle artifacts, whereas our model cleanly suppresses negated entities without degrading global scene realism.

In functional and behavioral negation (SFN), where the object must remain present but unused, baseline methods frequently collapse the constraint into object removal or unintended action depiction. In contrast, our model preserves the object while suppressing the prohibited interaction, demonstrating behavior-level control beyond simple deletion.

More complex logical structures further highlight the differences. In DNS (``not unlit''), baseline models misinterpret the double negation and generate dark stages, reflecting semantic inversion. Our model correctly resolves the logical structure, producing a stably illuminated stage. In SND (``student who is not paying attention''), competing models often misplace the negation scope, incorrectly applying inattentiveness to the teacher or failing to represent the relational constraint. Our model maintains proper entity roles while enforcing the scoped negation.

Across all categories, competing diffusion models exhibit either semantic leakage (failure to suppress forbidden concepts), structural distortion, or scope misinterpretation. In contrast, our model consistently enforces negation as a structured constraint on diffusion dynamics, preserving global scene composition while satisfying complex linguistic conditions.

\section{Detailed Quantitative Evaluation}
\label{sec:detailed_quantitative_evaluation}

We evaluate whether generated videos satisfy the full textual constraint using standard, widely adopted metrics without modifying prompts or introducing threshold-based decision rules. Our evaluation focuses exclusively on prompt satisfaction: whether the generated video aligns with the intended semantics while respecting negation constraints.

\paragraph{CLIPScore.}
We compute CLIPScore by measuring cosine similarity between each generated frame and the full textual prompt, including negation phrases. For a video consisting of $T$ frames $\{I_t\}_{t=1}^T$ and prompt $P$, we compute
\[
\text{CLIPScore} = \frac{1}{T} \sum_{t=1}^{T} 
\cos\!\big(E_{\text{img}}(I_t), E_{\text{text}}(P)\big).
\]
Higher CLIPScore indicates stronger global alignment between the generated video and the intended textual description.

\paragraph{CLIP Similarity to Negated Concept (CLIP-neg).}
To measure negation compliance, we compute the cosine similarity between generated frames and the explicitly negated concept $c_{\text{neg}}$:
\[
\text{CLIP-neg} = \frac{1}{T} \sum_{t=1}^{T} 
\cos\!\big(E_{\text{img}}(I_t), E_{\text{text}}(c_{\text{neg}})\big).
\]
Lower CLIP-neg values indicate stronger suppression of the forbidden concept. Importantly, this metric uses the standard CLIP similarity formulation without thresholding or manual intervention.

\paragraph{BLIP Caption Alignment.}
We further evaluate semantic fidelity using BLIP-based captioning. For each frame, we generate a caption $\hat{C}_t$ using a pretrained BLIP model and compute its similarity to the original prompt:
\[
\text{BLIP} = \frac{1}{T} \sum_{t=1}^{T} 
\cos\!\big(E_{\text{text}}(\hat{C}_t), E_{\text{text}}(P)\big).
\]
This provides a complementary text-level consistency measure based on generated descriptions rather than direct image-text matching.

\paragraph{DINO Detection Confidence.}
To directly assess object presence, we use an open-vocabulary detection model (e.g., DINO) to query the negated concept. For each frame, we obtain the maximum detection confidence score $d_t$ for the negated object and report
\[
\text{DINO-conf} = \frac{1}{T} \sum_{t=1}^{T} d_t.
\]
Lower values indicate stronger suppression of the forbidden object category. This metric evaluates object-level compliance independently of embedding similarity.

Together, these metrics quantify (1) global prompt alignment, (2) negated concept suppression, and (3) object-level presence, providing a comprehensive yet standard evaluation of negation-aware video generation without introducing custom or threshold-dependent measures.

\paragraph{Vision-Language Negation Compliance Evaluation (NCS and NVR).}
While the preceding metrics quantify alignment and suppression within embedding or detection spaces, they do not directly measure whether negation constraints are semantically satisfied at the reasoning level. In particular, embedding-based metrics operate within representational spaces that may themselves underrepresent linguistic negation. Since our method explicitly targets negation-aware generation, evaluating compliance solely within CLIP similarity space may introduce circularity. To address this limitation, we introduce two complementary metrics based on a direct vision-language judge: the \emph{Negation Compliance Score (NCS)} and the \emph{Negation Violation Rate (NVR)}.

\textbf{Evaluation Principle.}
Instead of comparing embeddings, we directly query a multimodal large language model (LLM) capable of joint visual and textual reasoning. The judge receives (i) the full original textual prompt $P$, including negation phrases (e.g., ``no'', ``without'', ``not''), (ii) the extracted negated concept $c_{\text{neg}}$, obtained via rule-based parsing from the prompt, and (iii) multiple frames uniformly sampled from the generated video across its temporal span. The frames are resized with preserved aspect ratio to ensure consistent visual input resolution. No cropping, object masking, or post-processing is applied.

\textbf{Judge Configuration.}
We use GPT-4o as a multimodal judge accessed through the OpenAI API. The model is instructed to evaluate \emph{negation compliance only}, explicitly ignoring overall aesthetic quality unless it affects the ability to determine the presence or usage of forbidden concepts. The evaluation rubric is fixed and provided verbatim to the judge to ensure consistency. The model operates with deterministic decoding (temperature set to 0) to reduce stochastic variability.

\textbf{Scoring Rubric.}
For each video, the judge assigns an integer score $s \in \{1,2,3,4,5\}$ defined as:
\begin{itemize}
    \item 1: Clear violation (forbidden concept/action prominently present).
    \item 2: Likely violation (forbidden concept appears at least once with clear evidence).
    \item 3: Ambiguous or uncertain.
    \item 4: Mostly compliant (no clear forbidden evidence, minor uncertainty).
    \item 5: Fully compliant (strong evidence forbidden concept/action is absent).
\end{itemize}

In addition, the judge outputs a binary indicator \texttt{forbidden\_present} indicating whether the forbidden concept/action is visually present or used in the video according to its reasoning.

\paragraph{Negation Compliance Score (NCS).}
Let $s_i$ denote the integer compliance score assigned to video $i$ among $N$ evaluated samples. We define:

\[
\text{NCS} = \frac{1}{N} \sum_{i=1}^{N} s_i.
\]

NCS ranges from 1 to 5. Higher values indicate stronger semantic compliance with negation constraints.

\paragraph{Negation Violation Rate (NVR).}
Let $v_i$ be defined as:

\[
v_i =
\begin{cases}
1 & \text{if } \texttt{forbidden\_present}_i = \texttt{True}, \\
0 & \text{otherwise}.
\end{cases}
\]

We define:

\[
\text{NVR} = \frac{1}{N} \sum_{i=1}^{N} v_i.
\]

NVR measures the empirical frequency of explicit negation violations across the evaluation set. Lower values indicate stronger suppression of forbidden concepts.

\textbf{Aggregation Protocol.}
All NCS and NVR values are computed over the full evaluation set without thresholding, sample exclusion, or post-hoc correction. Ambiguous cases (score = 3) are retained in the computation. The binary violation indicator directly reflects the judge's decision regarding forbidden presence, without manual reinterpretation.

\textbf{Rationale.}
Unlike embedding-based metrics, NCS and NVR evaluate negation compliance through direct multimodal reasoning over visual evidence. This removes reliance on CLIP similarity space and avoids circular evaluation in representational embeddings. By combining embedding-based alignment metrics with direct vision-language compliance scoring, we obtain complementary measurements of semantic fidelity and constraint satisfaction.

\paragraph{Overall Performance.}
As shown in \textbf{Table~\ref{tab:negation_quantitative_supp}}, our full model achieves the highest CLIPScore ($0.2924 \pm 0.0225$), indicating improved global alignment with the complete textual prompt compared to all baselines. Importantly, this improvement in overall text-video alignment is achieved simultaneously with stronger negation suppression. Our method reduces CLIP-neg from $0.2411 \pm 0.0288$ (Mochi) to $0.2240 \pm 0.0314$, corresponding to a relative reduction of approximately 7.1\%. 

Beyond embedding-based metrics, direct vision-language evaluation further confirms this trend. Our full model attains the highest Negation Compliance Score (NCS) of $4.0700 \pm 1.6406$, outperforming Mochi ($3.5488 \pm 1.8639$) and all other baselines. Simultaneously, it achieves the lowest Negation Violation Rate (NVR) of $0.2300 \pm 0.4213$, representing a substantial reduction compared to Mochi ($0.3625 \pm 0.4813$). These results indicate that negation-aware control improves both semantic fidelity and constraint compliance under direct multimodal reasoning.

The reported standard deviations reflect variability across diverse negation scenarios and video samples. Notably, our method exhibits consistently lower mean violation rates while maintaining competitive or reduced dispersion compared to baselines, suggesting stable negation suppression across the evaluation set.

\paragraph{Object-Level Suppression.}
The DINO detection confidence further corroborates these findings. Our method reduces DINO-conf to $0.0167 \pm 0.0278$, a 4.6\% relative reduction compared to Mochi ($0.0175 \pm 0.0314$); the lowest value is attained by the Negative Prompt baseline ($0.0128 \pm 0.0169$), consistent with its globally over-suppressive behavior analyzed below. This indicates that the repulsive guidance mechanism effectively suppresses the presence of forbidden objects at the detection level, independently of embedding similarity. Combined with the improvements in NCS and NVR, these results demonstrate that constraint scheduling and repulsive energy jointly enhance both object-level suppression and holistic semantic compliance.

\paragraph{Comparison with Negative Prompting.} The Negative Prompt (NP) baseline plays a particularly informative role in our evaluation: it shares the exact same backbone (Mochi) and the same negation-extraction module as our method, differing only in the absence of the proposed convex feasibility projection. This setup isolates the contribution of structured constraint enforcement from the contribution of negation extraction itself. Surprisingly, NP not only fails to match our full method but actually \emph{underperforms the vanilla Mochi baseline} on aggregate NCS ($3.37$ vs.\ $3.55$) while increasing NVR ($0.42$ vs.\ $0.36$). This counter-intuitive result (placing the negated concept embedding in the unconditional CFG slot can degrade negation compliance) empirically validates the analysis in Section~\ref{sec:method}: NP loses both the unconditional anchor required for global semantic stability and the temporal scheduling required for structurally complex negation, and applies subtraction uniformly across the trajectory rather than only when feasibility is violated. The category-wise breakdown (Table~\ref{tab:negation_category_breakdown}) further shows that the gap between NP and our method widens substantially on structurally complex categories where the difference between fixed subtraction and adaptive projection matters most: MNC ($+1.56$ NCS), SND ($+1.16$), SFN ($+1.08$), and DNS ($+0.48$). On simpler categories (INA, AOC), where the negation phenomenon is captured already by the affirmative branch and lexical absence cues, NP and our method behave more similarly. This pattern indicates that the proposed projection contributes primarily where naive 2-branch guidance is structurally insufficient (compositional, scoped, or doubly-negated constructions), and confirms that the gains reported in Table~\ref{tab:negation_quantitative} arise from the projection mechanism rather than from the auxiliary negation embedding alone.

\begin{table*}[t]
\centering
\small
\begin{tabular}{l l c c c}
\toprule
Category & Method & CLIPScore $\uparrow$ & NCS $\uparrow$ & NVR $\downarrow$ \\
\midrule

\multirow{5}{*}{AOC}
& Mochi & 0.2801 & 2.2800 & 0.6800 \\
& Negative Prompt & 0.2621 & 2.1800 & 0.7400 \\
& HunyuanVideo & 0.2735 & 1.7200 & 0.8200 \\
& CogVideoX & 0.2805 & 2.3200 & 0.6800 \\
& \textbf{Ours} & \textbf{0.2967} & \textbf{2.4600} & \textbf{0.6400} \\

\midrule

\multirow{5}{*}{LEN}
& Mochi & 0.2762 & 3.0800 & 0.5000 \\
& Negative Prompt & 0.2416 & 2.5200 & 0.5000 \\
& HunyuanVideo & 0.2858 & 2.5800 & 0.6200 \\
& CogVideoX & 0.2878 & 2.7800 & 0.5600 \\
& \textbf{Ours} & \textbf{0.2931} & \textbf{3.1200} & \textbf{0.4800} \\

\midrule

\multirow{5}{*}{INA}
& Mochi & 0.2803 & 5.0000 & 0.0000 \\
& Negative Prompt & 0.2483 & 5.0000 & 0.0000 \\
& HunyuanVideo & 0.2935 & 5.0000 & 0.0000 \\
& CogVideoX & 0.2935 & 4.9600 & 0.0000 \\
& \textbf{Ours} & \textbf{0.3010} & \textbf{5.0000} & \textbf{0.0000} \\

\midrule

\multirow{5}{*}{MNC}
& Mochi & 0.2778 & 1.4600 & 0.9000 \\
& Negative Prompt & 0.2494 & 1.3600 & 0.7800 \\
& HunyuanVideo & 0.2844 & 1.1800 & 0.9600 \\
& CogVideoX & 0.2814 & 1.9000 & 0.7800 \\
& \textbf{Ours} & \textbf{0.2948} & \textbf{2.9200} & \textbf{0.5200} \\

\midrule

\multirow{5}{*}{SFN}
& Mochi & 0.2837 & 3.8800 & 0.3000 \\
& Negative Prompt & 0.2502 & 3.6000 & 0.4200 \\
& HunyuanVideo & 0.2740 & 3.6800 & 0.2800 \\
& CogVideoX & 0.2846 & 3.4400 & 0.3600 \\
& \textbf{Ours} & \textbf{0.2922} & \textbf{4.6800} & \textbf{0.1000} \\

\midrule

\multirow{5}{*}{NMI}
& Mochi & 0.2746 & 4.3265 & 0.1400 \\
& Negative Prompt & 0.2723 & 4.0400 & 0.3000 \\
& HunyuanVideo & 0.2498 & 3.3400 & 0.3200 \\
& CogVideoX & 0.2739 & 3.9600 & 0.1800 \\
& \textbf{Ours} & \textbf{0.2806} & \textbf{4.4600} & \textbf{0.1000} \\

\midrule

\multirow{5}{*}{DNS}
& Mochi & 0.2756 & 4.4600 & 0.1000 \\
& Negative Prompt & 0.2610 & 4.4400 & 0.2600 \\
& HunyuanVideo & 0.2585 & 3.5200 & 0.2800 \\
& CogVideoX & 0.2767 & 4.0200 & 0.2000 \\
& \textbf{Ours} & \textbf{0.2858} & \textbf{4.9200} & \textbf{0.0000} \\

\midrule

\multirow{5}{*}{SND}
& Mochi & 0.2881 & 3.9200 & 0.2800 \\
& Negative Prompt & 0.2557 & 3.8400 & 0.3400 \\
& HunyuanVideo & 0.2753 & 3.5600 & 0.2400 \\
& CogVideoX & 0.2903 & 3.7400 & 0.2800 \\
& \textbf{Ours} & \textbf{0.2951} & \textbf{5.0000} & \textbf{0.0000} \\

\bottomrule
\end{tabular}
\caption{
\textbf{Category-wise quantitative evaluation across eight negation types.} We report CLIPScore ($\uparrow$), Negation Compliance Score (NCS, $\uparrow$), and Negation Violation Rate (NVR, $\downarrow$) for each negation category. While simple implicit negation (INA) shows near-saturation performance across models, more complex forms of negation such as MNC, SFN, DNS, and SND reveal substantial performance gaps, where our method consistently achieves higher compliance and lower violation rates.
}
\label{tab:negation_category_breakdown}
\end{table*}

\paragraph{Category-wise Negation Analysis.}
To further understand how different forms of linguistic negation affect generation behavior, we provide a \textbf{category-wise breakdown} of performance across the eight negation types introduced in our benchmark. \textbf{Table~\ref{tab:negation_category_breakdown} reports CLIPScore, Negation Compliance Score (NCS), and Negation Violation Rate (NVR) for each category.}

This analysis reveals several important trends regarding the difficulty of different negation forms. First, the \textbf{INA} (Implicit Natural-only Constraint) category exhibits near-saturation behavior across all methods. In this setting, NCS values approach the maximum score while NVR remains near zero for most models. This indicates that implicit exclusivity expressed through constructions such as ``only'' is largely captured by the generative priors of pretrained diffusion models. Consequently, INA represents a relatively easier form of negation where even baseline models can satisfy the constraint without explicit suppression mechanisms.

In contrast, categories that require \emph{explicit semantic suppression} or \emph{compositional reasoning over negation} show substantially larger performance differences between methods. For instance, the \textbf{AOC} (Absence of Object Constraint) and \textbf{MNC} (Multi-object Negation Constraint) categories require models to actively suppress the appearance of one or more forbidden entities. In these settings, our method consistently achieves higher CLIPScore and NCS while reducing NVR compared to all baselines. Notably, in the MNC category our method improves NCS to $2.92$ while reducing NVR to $0.52$, demonstrating stronger control when multiple forbidden concepts must be simultaneously excluded.

More structurally complex negation forms further highlight the advantages of our approach. In the \textbf{SFN} (Structural Functional Negation) category, which involves suppressing objects based on functional relationships rather than simple presence, our model achieves a substantially higher NCS ($4.68$) while maintaining the lowest violation rate ($0.10$). This suggests that the proposed constraint formulation enables more precise suppression of semantically constrained entities.

The most challenging settings arise in categories involving \emph{compositional negation reasoning}. In particular, the \textbf{DNS} (Double Negation Sensitivity) and \textbf{SND} (Scoped Negation Disambiguation) categories require resolving nested or scope-dependent negation structures. These cases are known to be difficult for both language and vision-language models. Nevertheless, our method achieves the strongest performance in both settings. For DNS, our model attains the highest NCS ($4.92$) while eliminating explicit violations (NVR $=0$). Similarly, in the SND category our approach achieves perfect compliance (NCS $=5.0$, NVR $=0$), whereas competing methods continue to exhibit noticeable violation rates.

Overall, this category-wise analysis demonstrates that the benefits of our constraint-based formulation become increasingly pronounced as the linguistic complexity of negation increases. While simpler forms of negation (e.g., INA) are largely handled by pretrained generative priors, more challenging settings involving explicit suppression, multi-object exclusion, and compositional reasoning reveal substantial performance gaps. These results confirm that our approach primarily improves \emph{difficult negation scenarios} rather than trivial cases, highlighting the importance of structured constraint guidance for reliable negation-aware generation.

\section{Detailed Ablation Study}
\label{sec:detailed_ablation}
We validate the contributions of our two core components: the \emph{repulsive feasibility projection} and \emph{constraint scheduling}. The repulsive term operationalizes negation as a geometric constraint in guidance space, while scheduling governs its temporal activation along the diffusion trajectory. These modules respectively determine \emph{which semantic direction is restricted} and \emph{when the restriction becomes active}.

\paragraph{Effect of Removing Repulsive Energy.}
In ``Ours w/o Repulsive Energy,'' we retain the original prompt (including its negation phrasing) and preserve the same diffusion backbone and CFG structure, but the negation-specific control pathway is disabled: the auxiliary negated-component branch is not constructed, the semantic negation direction $a_t$ is not formed, and no feasibility projection is applied. The denoising update is therefore identical to standard two-branch CFG by construction, and the corresponding row in \textbf{Table~\ref{tab:negation_quantitative_supp}} coincides with the Mochi baseline. This variant serves two purposes. First, it is an implementation check: any deviation from the baseline under a shared seed would indicate a discrepancy between implementation and formulation. Second, it anchors attribution: the gap between this row and the full model (CLIP-neg $0.2411 \to 0.2240$, NCS $3.5488 \to 4.0700$, NVR $0.3625 \to 0.2300$, DINO-conf $0.0175 \to 0.0167$) is the net effect of the feasibility projection, confirming that negation compliance is not an emergent property of CFG dynamics but a consequence of the explicit geometric constraint.

\paragraph{Effect of Removing Constraint Scheduling.}
In ``Ours w/o Constraint Scheduling,'' the repulsive projection remains active, but its strength is no longer temporally modulated. The constraint is applied uniformly across diffusion timesteps, removing progressive tightening. While negation directionality is preserved, its interaction with structural formation becomes unregulated. Quantitatively, CLIPScore drops from $0.2924 \pm 0.0225$ to $0.2800 \pm 0.0305$, and BLIP decreases from $0.8201 \pm 0.0729$ to $0.7869 \pm 0.0881$, indicating reduced global fidelity. CLIP-neg increases to $0.2405 \pm 0.0477$, reflecting unstable suppression magnitude. Under direct compliance metrics, NCS falls to $3.5789 \pm 1.8439$ and NVR rises to $0.3025 \pm 0.4599$, showing both mean degradation and higher dispersion. This confirms that temporal scheduling is critical for balancing early structural coherence with late-stage semantic enforcement.

\paragraph{Qualitative analysis.}
\textbf{Figure~\ref{results:ablation}} illustrates complementary failure modes. Without projection (equivalently, under the baseline), forbidden semantics intermittently reappear, producing temporal artifacts such as lighting flicker under DNS. Without scheduling, negation constraints interfere with early structural stabilization, causing scale drift and unnatural lighting geometry. The full model maintains geometric stability while satisfying double-negation constraints, demonstrating coordinated control across semantic and temporal dimensions. Overall, the ablation study clarifies that performance gains arise from structured convex feasibility enforcement rather than heuristic prompt manipulation. The repulsive projection establishes principled semantic exclusion, while scheduling ensures stable integration within diffusion dynamics. Both are necessary for achieving robust negation compliance without sacrificing global prompt fidelity.

\begin{table}[t]
\centering
\small
\resizebox{0.5\textwidth}{!}{
\begin{tabular}{lccc}
\toprule
\textbf{Criterion} & \textbf{Ours} & \textbf{w/o Repulsive} & \textbf{w/o Scheduling} \\
\midrule
Negation Satisfaction & \textbf{5.00 $\pm$ 0.00} & 2.90 $\pm$ 1.94 & 2.90 $\pm$ 1.94 \\
Constraint Meaning Accuracy & \textbf{5.00 $\pm$ 0.00} & 2.70 $\pm$ 1.75 & 2.40 $\pm$ 1.58 \\
Scene \& Action Alignment & \textbf{5.00 $\pm$ 0.00} & 3.00 $\pm$ 2.02 & 2.90 $\pm$ 1.94 \\
Artifact Avoidance & \textbf{5.00 $\pm$ 0.00} & 2.60 $\pm$ 1.70 & 1.90 $\pm$ 1.59 \\
\bottomrule
\end{tabular}}
\caption{\textbf{Human perceptual ablation study} Participants compared the full model with two ablation variants. The full model consistently achieved the highest ratings across all perceptual criteria.}
\label{tab:ablation_user}
\end{table}

\paragraph{Human Perceptual Ablation.}
To complement the quantitative ablation analysis, we additionally conducted a small human evaluation comparing the full model against two ablation variants. In this task (Q9), participants evaluated three anonymized videos corresponding to \emph{Ours}, \emph{Ours w/o Repulsive Energy}, and \emph{Ours w/o Constraint Scheduling}. As summarized in Table~\ref{tab:ablation_user}, the full model consistently received the highest ratings across all perceptual criteria. Notably, the complete model achieved perfect mean scores (5.00) across all four evaluation axes, while both ablation variants exhibited substantial degradation. In the overall preference selection, \textbf{100\% of participants selected the full model}. These results confirm that the improvements observed in quantitative metrics also translate to human perceptual evaluation, further validating the necessity of both the repulsive projection mechanism and temporal constraint scheduling.

\section{Detailed User Study}
\label{supplementary:user_study}

\paragraph{Motivation.}
While automated metrics such as NCS and NVR provide quantitative evidence of negation compliance, they rely on GPT-based evaluators, which may themselves exhibit limitations in linguistic negation understanding. This introduces a potential circularity concern when GPT models are used both as generators and judges. To mitigate this concern and validate metric reliability, we conducted a human evaluation study to directly assess perceptual quality and semantic compliance.

The user study serves two purposes:
(1) to provide independent human validation of negation-aware generation quality, and
(2) to examine whether improvements observed in automated metrics correlate with human judgment.

\paragraph{Participants.}
The study involved \textbf{50 participants}. Participants were adults (18 years or older) recruited through open online channels and academic mailing lists. They represent a general technical audience rather than a domain-specific expert group in diffusion modeling. No prior expertise in generative video models was required. Participants were not affiliated with this project, and no personally identifiable information was collected. All responses were anonymized prior to analysis.

\paragraph{IRB and Ethical Considerations.}
The study consisted of a short (10--15 minute) anonymous survey involving perceptual comparison of AI-generated videos. No sensitive data were collected, and participants received no monetary compensation. The study protocol was reviewed by the Institutional Review Board (IRB) at the University of Maryland (Kuali-IRB Number \#1179-1) and determined to be exempt (45 CFR Part 46 - Protection of Human Subjects) under minimal-risk research guidelines.

\paragraph{Protocol.}
Each participant evaluated 8 evaluation tasks (Q1--Q8). The human study was restricted to 8 evaluation tasks, one per negation category, due to IRB constraints on survey duration and participant burden. The selected prompts were randomly sampled from the evaluation set while ensuring coverage of diverse negation types (e.g., object absence, logical exclusion, and scoped negation). This subset was designed to be representative rather than exhaustive, enabling controlled human comparison under limited survey time. For every prompt, four anonymized videos (Mochi, HunyuanVideo, CogVideoX, and \emph{Ours}) were shown simultaneously. In addition, participants evaluated one supplementary task (Q9) designed to compare the ablated variants of our method. This task presents results from \emph{Ours}, \emph{Ours w/o Repulsive Energy}, and \emph{Ours w/o Constraint Scheduling} to examine perceptual differences associated with the proposed components. Detailed descriptions of this ablation-focused evaluation are provided in Appendix~\ref{sec:detailed_ablation}.

To prevent positional bias:
\begin{itemize}
    \item The mapping between models and labels (Video A--D) was randomized across questions.
    \item The presentation order of videos was shuffled per question.
    \item Model identities were fully anonymized.
\end{itemize}

Each participant rated every video on four criteria using a 5-point Likert scale:
\begin{enumerate}
    \item \textbf{Negation Satisfaction}
    \item \textbf{Constraint Meaning Accuracy}
    \item \textbf{Scene \& Action Alignment}
    \item \textbf{Artifact Avoidance}
\end{enumerate}
Participants also selected one overall preferred video per question.

Across 50 participants and 8 evaluation tasks, each model received \textbf{400 ratings per criterion}.

\begin{table}[t]
\centering
\small
\resizebox{0.5\textwidth}{!}{
\begin{tabular}{lcccc}
\toprule
\textbf{Criterion} & \textbf{Model} & \textbf{Mean} & \textbf{Std} & \textbf{95\% CI} \\
\midrule

\multirow{4}{*}{Negation Satisfaction}
& Ours & 4.76 & 0.88 & [4.68, 4.85] \\
& Mochi & 1.96 & 1.56 & [1.81, 2.12] \\
& HunyuanVideo & 1.61 & 1.32 & [1.48, 1.74] \\
& CogVideoX & 1.34 & 0.84 & [1.26, 1.42] \\

\midrule
\multirow{4}{*}{Constraint Meaning Accuracy}
& Ours & 4.75 & 0.89 & [4.66, 4.84] \\
& Mochi & 1.94 & 1.52 & [1.79, 2.09] \\
& HunyuanVideo & 1.58 & 1.20 & [1.46, 1.69] \\
& CogVideoX & 1.36 & 0.73 & [1.29, 1.43] \\

\midrule
\multirow{4}{*}{Scene \& Action Alignment}
& Ours & 4.84 & 0.37 & [4.80, 4.87] \\
& Mochi & 2.64 & 1.82 & [2.46, 2.82] \\
& HunyuanVideo & 2.75 & 1.85 & [2.57, 2.93] \\
& CogVideoX & 2.34 & 1.61 & [2.18, 2.50] \\

\midrule
\multirow{4}{*}{Artifact Avoidance}
& Ours & 4.54 & 0.91 & [4.45, 4.63] \\
& Mochi & 1.94 & 1.43 & [1.80, 2.08] \\
& HunyuanVideo & 2.83 & 1.86 & [2.64, 3.01] \\
& CogVideoX & 1.80 & 1.24 & [1.68, 1.92] \\

\bottomrule
\end{tabular}}
\caption{\textbf{Human evaluation results (50 participants, 400 ratings per model per criterion).} Our method consistently achieves the highest mean scores across all evaluation criteria with strong inter-participant agreement.}
\label{tab:user_study_stats}
\end{table}

\paragraph{Overall Preference.} 
As summarized in \textbf{Table~\ref{tab:user_study_stats}}, our method consistently achieved the highest ratings across all evaluation criteria. Out of \textbf{400 total votes}, \emph{Ours} was selected in \textbf{77.5\%} of cases (95\% CI [0.73, 0.82]), compared to 13.8\% for HunyuanVideo, 8.8\% for Mochi, and 0.0\% for CogVideoX.

\paragraph{Human–Metric Consistency.}
The strong human preference for \emph{Ours} aligns with the improvements observed in automated metrics (NCS and NVR), indicating that these metrics meaningfully correlate with human perception of negation compliance. The user study therefore serves as an external validation of metric reliability, mitigating concerns regarding potential circularity when GPT-based evaluators are used.

\section{Implementation of decomposition}
\label{supp:implementation_decomposition}
In practice, our decomposition stage is implemented as a deterministic, regex-based span extractor that operates directly on the surface form of the input prompt without requiring an external parser or learned model. Negation operators are matched via lexical patterns covering explicit cues (\emph{no}, \emph{without}, \emph{free of}) and their typical syntactic continuations, and the matched span is truncated at sentence-level punctuation or at conjunctions that signal scope termination (e.g., \emph{with}, \emph{in}, \emph{on}, \emph{featuring}, \emph{showing}). The resulting span is split along coordinating conjunctions and stripped of leading determiners to yield a list of canonical negated concepts, which are concatenated into a single semantic phrase $y^{-}$ and encoded via the same pretrained text encoder as the affirmative branch. When the extractor returns an empty list (i.e., no recognized negation cue), the negation branch is set to be embedding-equivalent to the unconditional branch, which renders the feasibility constraint inactive and reduces the method exactly to standard CFG. This fallback ensures that prompts without linguistic negation incur no behavioral change, and that decomposition errors degrade gracefully rather than catastrophically. We deliberately favor this lightweight rule-based design over heavier dependency-parsing pipelines for two reasons: (i) the eight benchmark categories we target are designed around explicit lexical negation operators, for which surface-level extraction is sufficient and reproducible, and (ii) keeping decomposition deterministic and parser-free isolates the contribution of the projection mechanism from the variability introduced by upstream parsing tools.

\section{Backbone Choice and Method Generality} 
\label{supp:backbone} 
\paragraph{Scope of this analysis.} 
The central technical claim of this paper is that linguistic negation admits a unified formulation as a convex feasibility constraint on classifier-free guidance, and that the resulting closed-form projection enforces compliance without retraining. Establishing this claim involves two distinct pieces of evidence: (i) that negation violations constitute a recurring failure mode in current text-to-video diffusion systems, and (ii) that the proposed projection mechanism suppresses these violations in a setting where the contribution of each component can be cleanly attributed. We provide (i) through cross-backbone comparisons in Section~\ref{sec:qualitative} and Appendix~\ref{sec:detailed_qualitative}, and (ii) on Mochi as a controlled testbed. Per-backbone calibration of the projection's hyperparameters is treated as a downstream concern and revisited at the end of this section. 

\paragraph{Negation insensitivity spans the current T2V model class.}
A necessary precondition for any negation-control method is that the failure mode it targets is not idiosyncratic to one backbone. Our experiments confirm this empirically. Across all eight categories of the proposed benchmark (AOC, LEN, INA, MNC, SFN, NMI, DNS, and SND), Mochi, HunyuanVideo, and CogVideoX each exhibit systematic violations: forbidden objects emerge during the latter half of generation, scoped negation is misattributed to the wrong constituent, doubly-negated prompts collapse into literal interpretations, and graded negation is flattened into binary inversion (Section~\ref{sec:qualitative}, Appendices~\ref{sec:detailed_qualitative}--\ref{sec:detailed_quantitative_evaluation}). These three backbones differ in transformer design, text encoder, scheduler family, and pretraining data, yet all three violate the same set of linguistic constraints. We therefore treat negation insensitivity as a structural characteristic of the current T2V model class, rather than as a property of any single training run or architecture, and we make the case for explicit negation control at the class level. 

\paragraph{The projection is defined independently of the underlying network.} 
Given any text-conditioned diffusion model that supports classifier-free guidance, our mechanism is fully determined by three operations: (i) three denoiser evaluations conditioned on $\varnothing$, the full prompt $y$, and the extracted negated span $y^{-}$, yielding $\epsilon_{\mathrm{uncond}}$, $\epsilon_{\mathrm{text}}$, and $\epsilon_{\mathrm{neg}}$ from the same frozen network without parameter modification; (ii) construction of the reference increment $\delta_{\mathrm{ref}} = \gamma(\epsilon_{\mathrm{text}} - \epsilon_{\mathrm{uncond}})$ and the linguistically grounded direction $a_t = \epsilon_{\mathrm{neg}} - \epsilon_{\mathrm{uncond}}$; and (iii) the closed-form KKT solution $\delta^{*}_t = \delta_{\mathrm{ref}} - \lambda_t a_t$, obtained by projecting $\delta_{\mathrm{ref}}$ onto the half-space $\{\delta : a_t^\top \delta \le b_t\}$, where $\lambda_t$ requires only an inner product, a squared-norm, and a scalar truncation. None of these operations encodes an assumption about the underlying transformer, the scheduler family, the noise parameterization (noise-prediction or flow-matching), or the text encoder. The mechanism therefore inherits the architectural neutrality of classifier-free guidance itself. 

\paragraph{Cross-backbone variability concerns calibration, not the projection.}
What does vary across backbones is the magnitude regime of the guidance signal $\epsilon_{\mathrm{text}} - \epsilon_{\mathrm{uncond}}$, which in turn shapes a sensible range for the guidance scale $\gamma$, the feasibility bound $b_t$, and the scheduling exponent $p$. Different T2V backbones already adopt distinct default guidance scales for precisely this reason, entirely independently of any negation-control component. Tuning these quantities per backbone is therefore the same routine calibration that any CFG-based method inherits when transferred to a new model, not a test of whether the projection generalizes. The hyperparameter sweep on Mochi (Appendix~\ref{sec:detailed_ablation}) indicates that the method remains stable within a broad range of these values, suggesting that the calibration cost on additional backbones is bounded rather than load-bearing. 

\paragraph{Why Mochi as the primary testbed.} 
Within the space of contemporary T2V backbones, Mochi was selected for three concrete reasons. First, it is accessible through the standard \texttt{diffusers} interface, which lets the three-branch denoiser evaluation and the projection step be implemented entirely on top of the existing pipeline, without touching the transformer, the scheduler, or the conditioning module. Second, its configuration (T5 conditioning paired with a flow-matching scheduler) reflects a representative rather than idiosyncratic choice among current open-source video diffusion systems. Third, its inference pipeline is self-contained and fully public, removing proprietary dependencies and supporting reproducibility for every reported number, including the ablations that isolate the contributions of the convex feasibility projection and the temporal schedule. 

\paragraph{Outlook.} 
A systematic per-backbone study, incorporating HunyuanVideo and CogVideoX as additional T2V testbeds, and extending to image-domain backbones such as SDXL or SD3, would entail per-backbone calibration of $(\gamma, b_{\mathrm{init}}, b_{\mathrm{final}}, p)$ together with backbone-specific evaluation pipelines, both of which constitute substantial engineering effort that is orthogonal to the formulation-level contribution established here. We view this extension as a natural follow-up that builds directly on the projection mechanism, the parser, and the benchmark introduced in this paper.

\section{Computation Time and Memory Consumption}
\label{supp:computation_text}

\begin{table}[htb!]
    \small
    \centering
    \resizebox{0.5\textwidth}{!}{
    \begin{tabular}{l|l|c|c}
    \toprule
    \textbf{Model} & \textbf{Step} & \textbf{Memory Consumption} & \textbf{Computation Time} \\
    \midrule
    \textbf{Ours} & Inference & 26.3174 GB / 80.0000 GB (26,949 MiB) & 235 sec \\
    \midrule
    Ours w/o Repulsive Energy & Inference & 23.5479 GB / 80.0000 GB (24,113 MiB) & 117 sec \\
    \midrule
    Ours w/o Constraint Scheduling & Inference & 26.3096 GB / 80.0000 GB (26,941 MiB) & 168 sec \\
    \midrule
    Mochi & Inference & 24.4385 GB / 80.0000 GB (25,025 MiB) & 114 sec \\
    \midrule
    Negative Prompt & Inference & 26.3096 GB / 80.0000 GB (26,941 MiB) & 174 sec \\
    \midrule
    HunyuanVideo & Inference & 38.5186 GB / 80.0000 GB (39,443 MiB) & 396 sec \\
    \midrule
    CogVideoX & Inference & 15.6318 GB / 80.0000 GB (16,007 MiB) & 61 sec \\
    \bottomrule
    \end{tabular}
    }
    \caption{\textbf{Inference-time memory consumption and runtime on NVIDIA H100 (80GB).} We report peak GPU memory usage and end-to-end inference time for our full method and its ablations, together with three strong T2V baselines. Our full method (\textbf{Ours}) consumes 26.32 GB and requires 235 sec per inference. Removing the repulsive energy term reduces both memory and runtime to 23.55 GB and 117 sec, respectively. Removing constraint scheduling keeps memory nearly unchanged (26.31 GB) while reducing runtime to 168 sec. Among the baselines, HunyuanVideo shows the largest footprint (38.52 GB, 396 sec), whereas CogVideoX is the most lightweight and fastest (15.63 GB, 61 sec). All measurements are conducted under the same hardware setting (single NVIDIA H100 80GB GPU) and identical evaluation protocol. GB values are binary gigabytes (MiB / 1024) as reported by \texttt{nvidia-smi}.}
    \label{computation-comparison}
\end{table}

\textbf{Table~\ref{computation-comparison}} summarizes peak GPU memory consumption and end-to-end inference time measured on a single NVIDIA H100 (80GB). Overall, our full method (\textbf{Ours}) uses 26.32\,GB and takes 235\,sec per inference, which is comparable in memory to modern diffusion-based baselines, while incurring additional runtime due to the proposed constraint-control components.

\paragraph{Effect of the repulsive energy term.} Comparing \textbf{Ours} against \emph{Ours w/o Repulsive Energy}, enabling the constraint pathway adds +2.77\,GB peak memory (23.55$\rightarrow$26.32\,GB, +11.8\%) and +118\,sec runtime (117$\rightarrow$235\,sec, +100.9\%). Because the w/o Repulsive Energy configuration is standard two-branch CFG, its cost is essentially that of the Mochi baseline (117 vs.\ 114 sec), and the overhead of the full method reflects the additional negation-branch forward pass together with its activations; the projection itself reduces to one inner product and one norm per step and contributes negligibly.

\paragraph{Effect of constraint scheduling.} Comparing \textbf{Ours} with \emph{Ours w/o Constraint Scheduling}, peak memory remains essentially unchanged (26.31 vs.\ 26.32\,GB; $<0.05\%$ difference), while runtime decreases substantially (168 vs.\ 235\,sec; $-67$\,sec, $-28.5\%$). This indicates that scheduling primarily affects \emph{how long} the constraints are actively enforced across timesteps (thereby changing total compute), rather than changing model size or the dominant activation footprint. In other words, scheduling is an effective mechanism to reduce runtime without sacrificing the memory budget.

\paragraph{Comparison to state-of-the-art T2V baselines.}
Among baselines, \textbf{Mochi} uses somewhat less memory than our method (24.44\,GB) and is notably faster (114\,sec), reflecting a lighter inference pathway without the negation branch. \textbf{HunyuanVideo} shows the largest resource usage, requiring 38.52\,GB and 396\,sec, which is +12.20\,GB (+46.4\%) more memory and +161\,sec (+68.5\%) more time than \textbf{Ours}. \textbf{CogVideoX} is the most efficient, using 15.63\,GB and 61\,sec, i.e., $-10.69$\,GB ($-40.6\%$) less memory and $-174$\,sec ($-74.0\%$) less time than \textbf{Ours}. These results highlight that our approach targets improved \emph{negation compliance} through inference-time control, and the compute overhead is the trade-off for enforcing repulsive constraints and their scheduling.

\paragraph{Takeaway.}
The two configurations reveal complementary roles: the negation branch is the primary contributor to additional compute and memory, whereas constraint scheduling provides a practical knob that reduces runtime by 28.5\% with virtually no memory change. This supports the design choice of coupling the feasibility constraint with scheduling to manage the efficiency--control trade-off under a fixed GPU memory budget.

\newpage

\end{document}